\documentclass[12pt]{article}
\usepackage{amsmath}
\usepackage{graphicx}
\usepackage{enumerate}
\usepackage{natbib}
\usepackage{graphicx,psfrag,epsf,amsfonts,amscd, amssymb,latexsym,verbatim}
\usepackage[english]{babel}
\usepackage{bm}
\usepackage{color, colortbl}
\usepackage{algorithm}
\usepackage{algpseudocode}
\usepackage{tikz}%draw graphs
\usetikzlibrary{patterns}
\usepackage{tabularx}
\usepackage{adjustbox}
\usepackage{textcomp}
\usepackage[title,toc,titletoc]{appendix}
\usepackage{titlesec}
\usepackage{caption}
\def\bSig\mathbf{\Sigma}

\def\T{{ \mathrm{\scriptscriptstyle T} }} %transpose

\def\T{{ \mathrm{\scriptscriptstyle T} }}

\newcommand{\mathsc}[1]{{\normalfont\textsc{#1}}}

\newcommand{\prob}{\pi} 

\newcommand{\bagoft}{\mathsc{bag}}

\newcommand{\data}{D}
\newcommand{\rv}{R}
\newcommand{\rvt}{R^*}
\newcommand{\sig}{\sigma}
\newcommand{\sigt}{\sigma^*}

\newcommand{\cst}{c}
\newcommand{\bd}{B}

\newcommand{\gp}{G}
\newcommand{\Support}{\mathbb S}
\newcommand{\ntrain}{n_1}
\newcommand{\ntest}{n_2}
\newcommand{\limp}{\rightarrow_p}

\def\T{{ \mathrm{\scriptscriptstyle T} }}
\def\N{\mathcal{N}}

\newcommand{\csum}{T}
\newcommand{\csumt}{T^*}

\newcommand{\rate}{r}

\newcommand{\suset}{\mathbb M_n}
\newcommand{\V}{\mathcal V}
\newcommand{\convex}{\mathcal C^{K_n}}
\newcommand{\floor}[1]{\left\lfloor #1 \right\rfloor}

\graphicspath{{figures/}}

\newtheorem{definition}{Definition}

\newtheorem{theorem}{Theorem}%\newtheorem{theorem}{Theorem}[section]

\newtheorem{corollary}{Corollary}

\newtheorem{condition}{Condition}

\begin{document}

\def\spacingset#1{\renewcommand{\baselinestretch}%
{#1}\small\normalsize} \spacingset{1}

%%%%%%%%%%%%%%%%%%%%%%%%%%%%%%%%%%%%%%%%%%%%%%%%%%%%%%%%%%%%%%%%%%%%%%%%%%%%%%

  \title{\bf %BAGofT for Measuring the Goodness-of-Fit of Classification Procedures
  Is a Classification Procedure Good Enough?-A Goodness-of-Fit Assessment Tool for Classification Learning}
   \author{Jiawei Zhang, Jie Ding, and Yuhong Yang\\
     School of Statistics, University of Minnesota\\
     zhan4362@umn.edu, dingj@umn.edu, and yangx374@umn.edu}
\date{}
  \maketitle

\bigskip
\begin{abstract}
In recent years, many non-traditional classification methods, such as Random Forest, Boosting, and neural network, have been widely used in applications. Their performance is typically measured in terms of classification accuracy. While the classification error rate and the like are important, they do not address a fundamental question: Is the classification method underfitted? To our best knowledge, there is no existing method that can assess the goodness-of-fit of a general classification procedure. Indeed, the lack of a parametric assumption makes it challenging to construct proper tests. To overcome this difficulty, we propose a methodology called BAGofT that splits the data into a training set and a validation set. First, the classification procedure to assess is applied to the training set, which is also used to adaptively find a data grouping that reveals the most severe regions of underfitting. Then, based on this grouping, we calculate a test statistic by comparing the estimated success probabilities and the actual observed responses from the validation set. The data splitting guarantees that the size of the test is controlled under the null hypothesis, and the power of the test goes to one as the sample size increases under the alternative hypothesis. For testing parametric classification models, the BAGofT has a broader scope than the existing methods since it is not restricted to specific parametric models (e.g., logistic regression). Extensive simulation studies show the utility of the BAGofT when assessing general classification procedures and its strengths over some existing methods when testing parametric classification models.

\end{abstract}

\noindent%
{\it Keywords:}  goodness-of-fit test, classification procedure, adaptive partition 

\vfill

\newpage
\tableofcontents
\newpage

\section{Introduction}
\label{sec:intro}

% Goodness-of-fit tests aim to assess the potential discrepancy between the observed data and a fitted model.
% Briefly speaking, the `good enough' means the procedure, once trained on the observed data, can accurately characterize the class label's conditional distribution given the predictors. %In other words, there is no need to try another procedure and seek further improvement.
The development of various classification procedures has been a backbone of the contemporary learning toolbox to solve various data challenges. This work addresses the following fundamental problem in classification learning:
\textit{How to assess whether a classification procedure is good enough, in the sense that it has no systematic defects, as reflected in its 
convergence to the data-generating process, for given data?} 

We highlight that the assessment raised in the above question is fundamentally different from assessing the predictive performance.
In most applications, a classification procedure's performance is often assessed based on its \textit{classification accuracy} on preset validation data or through cross-validation.
Conceptually, the predictive accuracy does not characterize a procedure's deviation from the underlying data generating process per se.
For instance, when the conditional probability function of success given the covariates is simply 0.5, the best possible classifier is a random guess, which provides a low classification accuracy.  

It is critical to address the above question in several emerging learning scenarios where the classification accuracy alone cannot solve the problems.  For example, an increasing number of entities use Machine-Learning-as-a-Service (MLaaS)~\citep{ribeiro2015mlaas} or cooperative learning protocols~\citep{DingAssist} to train a model from paid cloud-computing services. It is economically significant to decide whether the current learning method has a significant discrepancy from the data and needs to be further improved.
Another example concerns the use of `benchmark data' for comparing classification procedures, e.g., those from %\citep{Kaggle,UCI}. 
Kaggle (https://rb.gy/bvepug) or UCI (https://archive.ics.uci.edu/ml/datasets.php).
Based on the validation accuracy as an evaluation metric, the winning procedure selected from many candidate learners may have already been overfitting luckily and deviating from the underlying data generating process. In this case, assessing the deviation of the learning procedures from the data distribution is also very helpful.

% Motivated by the above potential applications, we develop a methodology to address the above question. The developed tools will guide data analysts to understand whether a given procedure, possibly selected from a set of candidates, deviates significantly from the underlying data distribution. We focus on assessing binary-classification procedures that provide estimates of the conditional probability function. %(e.g., the softmax in neural networks).
% The same idea can be generalized to multi-class classification problems.
  
In dealing with a parametric model, the existing literature addresses the above problem from a goodness-of-fit (GOF) test perspective. 
For binary regression, two classical approaches are the Pearson's chi-squares ($\chi^2$) test and the residual deviance test, which group the observations according to distinct covariate values. When the number of observations in each group is small, e.g., there is at least one continuous covariate, the above two tests cannot be applied. Various tests have been developed to address this issue.
These include the tests based on the distribution of the Pearson's $\chi^2$ statistic under sparse data \citep{mccullagh1985asymptotic, osius1992normal, farrington1996assessing}, 
kernel smoothed residuals~\citep{le1991goodness}, the comparison with a generalized model~\citep{stukel1988generalized}, the comparison between an estimator from the control data and an estimator from the joint data in the context of case-control studies~\citep{Bondell}, the Pearson-type statistics calculated from bootstrap samples~\citep{yin2013pearson}, information matrix tests \citep{white1982maximum,orme1988calculation}, the grouping of observations into a finite number of sets \citep{hosmer1980goodness, pigeon1999improved, pulkstenis2002two,xie2008increasing,liu2012omnibus}, and the predictive log-likelihood on validation data~\citep{lu2019assessing}.

However, there are two weaknesses of the existing GOF tests for parametric classification models. 
First, most tests only control the Type I error probability, but without theoretical guarantees on the test power. Second, the existing methods focus on the GOF of specific models, such as logistic regression, and may not be applied to general binary regression models. 

For general classification procedures such as decision trees, neural networks, $k$-nearest neighbors, and support vector machines, to our best knowledge, there is no existing method to assess their GOF.
%To our best knowledge, there is no existing method to assess the goodness-of-fit of general classification procedures such as decision trees, neural networks, $k$-nearest neighbors, and support vector machines. 
We broaden the notion of the GOF test to address the question above for general classification procedures. 

We propose a new methodology named the binary adaptive goodness-of-fit test (BAGofT) for testing the GOF of both parametric classification models and general classification procedures.  The developed tools may guide data analysts to understand whether a given procedure, possibly selected from a set of candidates, deviates significantly from the underlying data distribution. We focus on assessing binary classification procedures that provide estimates of the conditional probability function. %(e.g., the softmax in neural networks).
% The same idea can be generalized to multi-class classification problems.

The BAGofT employs a data splitting technique, which helps the test overcome the difficulties in the general setting where there is no workable saturated model to compare with, as used in Pearson's chi-squares and deviance-based tests. 
On the `training' set, the BAGofT applies an adaptive partition of the covariate space that highlights the potential underfitting of the model or procedure to assess. Then, the BAGofT calculates a Pearson-type test statistic on the remaining `validation' part of the data based on the grouping from the adaptive partition.
%In the first step, the `training' part of data is used to partition the data space into potential underfitting regions under the model to assess. In the second step, the `validation' part uses the regions from the first step to construct a Pearson's test to assess the classification procedure's adequacy. The partitioning in the above first step is made in an adaptive or adversarial manner to enhance the second step's testing power. Meanwhile, the test statistic is based on validation data and grouping (from training data) independent by design. The above allows us to develop asymptotic theories under both the null the alternative to justify its performance. 

For parametric classifications, the BAGofT enjoys theoretical guarantees for its consistency under a broad range of alternative hypotheses, including those concerning misspecified covariates and model structures. 
Its adaptive partition can flexibly expose different kinds of weaknesses from the parametric classification model to test. Importantly, unlike the previous methods,  it allows the number of groups to grow with the sample size when a finer partition is needed. Moreover, the probability of the Type I error is well controlled due to data splitting. 
%Moreover, the adaptive nature of the BAGofT can flexibly expose different kinds of weaknesses from the parametric classification model to test while maintaining the probability of Type I error due to data splitting. 
%The BAGofT can also test the goodness-of-fit of the model to assess relative to all available covariates, including those not in the model. 

For a general classification procedure without a  workable benchmark to compare with, one major challenge is to define the GOF. Unlike parametric models, whose convergence is well understood, general classification procedures can have different convergence rates.  
If we choose the splitting ratio of the BAGofT according to a specific rate, the size of the test can be controlled as long as the procedure to assess converges not slower than the specified rate under the null hypothesis; the BAGofT consistently rejects the hypothesis otherwise. In practice, since the convergence rate of the procedure to assess is unknown, we advocate a method based on the BAGofT with multiple data splitting ratios.  Our experimental results show that this method can faithfully reveal possible moderate or severe deficiency of a classification procedure. 

The outline of the paper is given as follows.
In Section~\ref{bgt}, we provide the background of the problem. In Section~\ref{sec_bag}, we introduce the BAGofT and establish its properties. In Section~\ref{sec_pracguid}, we provide some practical guidelines on implementing the BAGofT.
We present simulation results in Section~\ref{simstu} and real data examples in Section~\ref{realExp}.
We conclude the paper in Section~\ref{Conclu}.  
The proofs and additional numerical results are included in the supplementary material.

%\subsection{Hosmer-Lemeshow  test}\label{HL}
%\input{HL.tex}
%TODO: Change outline 
\section{Problem Formulation} \label{bgt}

% \subsection{Notations}
% We use $\rightarrow_p$ and $\rightarrow_d$ to denote convergence in probability and in distribution, respectively. We use $\chi_k^2$ to denote the chi-squared distribution with $k$ degrees of freedom, $\textrm{Uniform}[a,b]$ to denote the uniform distribution on the interval $[a,b]$, and $\mathcal{N}(\mu,\sigma^2)$ to denote the Gaussian distribution with mean $\mu$ and variance $\sigma^2$.

\subsection{Setup}\label{setting}

Let $Y$ be the binary response variable that takes 0 or 1, and $\bm{X}$ be the vector of $p$ covariates. The support of $\bm{X}$ is  $\Support  \subseteq \mathbb R^{p}$. 
Let $\prob(\cdot)$ be the conditional probability function: %We denote the conditional probability of the response to be 1 given the covariate $\bm{X}$'s value to be $\bm{x}$ by
\begin{equation}
\prob(\bm{x}) = P(Y=1|\bm{X} = \bm{x}),\ \bm{x}\in \Support.
\end{equation}
 The data, denoted by $\data_n$, consist of $n$ i.i.d.\ observations from a population distribution of the pair $(Y,\bm{X})$. The conditional probability function $\prob(\cdot)$ is allowed to change with the sample size. %, as is common in high dimensional regression. 
 We denote the fitted conditional probability function obtained
 from a classification model (or a procedure) on $\data_n$ by $\hat\prob_{\data_n}(\cdot)$. 

\subsection{Testing parametric classification models}\label{Subsec_paramTest}
% \subsection{Parametric classification model}
 
 A parametric classification model assumes that  $\prob(\cdot)=f(\cdot,\bm{\beta})$, where $f$ is known and the unknown parameter $\bm{\beta}$ is in a finite dimensional set $\mathbb B$.  
For example, a generalized linear model assumes that
$f(\bm{x},\bm{\beta}) =  g^{-1}(\bm{x}^\T \bm\beta)$, where $g(\cdot)$ is a link function.
%$f(\bm{x},\bm{\beta}) = \mathbb E(Y|\theta) = d b(\theta)/d\theta = g^{-1}(\bm{x}\beta)$, where the response has a density function $ c(y)\exp(y\theta - b(\theta))$ for some $c(y)>0$, and $g(\cdot)$ is the link function.
The null and alternative hypotheses of the GOF for testing a parametric classification model are defined by
 \begin{align*}
     H_0:\ \prob(\cdot)\in \{f(\cdot,\bm{\beta})\mid \bm{\beta}\in \mathbb B\},\quad H_1:\ \prob(\cdot)\notin \{f(\cdot,\bm{\beta}) \mid \bm{\beta}\in \mathbb B\}.
 \end{align*}
 We refer to the parametric classification model to assess as MTA. 
% \subsection{General classification procedure}
\subsection{Assessing general classification procedures}

Compared with parametric classification models, general classification procedures are not restricted to be in a parametric form. They include any modeling technique that maps the data $\data_{n}$ to a fitted conditional probability function $\hat\prob_{\data_{n}}(\cdot):\Support\rightarrow [0,1]$. 
For a general classification procedure, the convergence rate of $\hat\prob_{\data_n}(\cdot)$ is essential from a theoretical viewpoint.  
Let $r_n$ be the convergence rate of the classification procedure we assess under the null hypothesis.  The null and alternative hypotheses of the GOF test for a general classification procedure to assess (PTA) are 
\begin{align*}
    H_0:&\ 
    \underset{\bm{x}\in \Support}{\sup}|\hat\prob_{\data_{n}}(\bm{x})- \prob(\bm{x})| = O_p(\rate_n),\\
   H_1:&\ \exists\ \suset\subseteq \mathbb S \text{ with } P(\bm{x}\in \suset) \text{ bounded away  from }0\text{ such that }\\ & \inf_{\bm{x}\in \suset}|\hat\prob_{\data_{n}}(\bm{x})- \prob(\bm{x})|/\rate_n \rightarrow_p\infty,
\end{align*}
as $n\rightarrow\infty$, where the set $\suset$ may change with $n$.
So under $H_0$, $\hat\prob_{\data_{n}}(\cdot)$ converges to $\prob(\cdot)$ not slower than $\rate_n$, and under $H_1$, it converges slower (or does not converge) to $\prob(\cdot)$.

%\section{BAGofT for testing parametric classification models}
\section{Binary Adaptive Goodness-of-fit Test (BAGofT)}\label{sec_bag}

\subsection{Test statistic}\label{param}

The BAGofT is a two-stage approach where the first stage explores a data-adaptive grouping and the second stage performs testing based on that grouping. %Two key ingredients involved in the test are 1) a set of groups and 2) estimated probabilities. 
The adaptive grouping consists of the following steps.
\textbf{(1)} Split the data into a training set $\data_{\ntrain}$ with size $\ntrain$ and a validation set $\data_{\ntest}$ with size $\ntest$. 
\textbf{(2)} Apply the MTA or PTA to $\data_{\ntrain}$ and obtain the estimated probabilities for both the  training set and validation set. 
\textbf{(3)} Generate a partition $\{\hat\gp_{\data_{\ntrain},1}, \ldots \hat\gp_{\data_{\ntrain},K_n}\}$ of the support $\Support$. This partition can be obtained by any method that meets the following two requirements. \textbf{i}
Denote the set of responses and covariates in $\data_{\ntest}$ by $\data_{y_e}$ and $\data_{x_e}$, respectively. The partition needs to be independent of $\data_{y_e}$ conditional on $\data_{x_e}$. It means that we may obtain a partition based on the performance of the MTA/PTA on the training set.  We can also use $\data_{x_e}$ to control the group sizes for the partition of $\data_{\ntest}$. 
\textbf{ii} The number of groups $K_n\geq 2$.  Note that $K_n$ can be data-driven and is not required to be uniformly upper bounded. We propose an adaptive partition algorithm, which is elaborated in Section~\ref{gp}.
\textbf{(4)} Group $\data_{\ntest}$ into sets based on the obtained partition. Let $\bm{x}_{e,i}$ ($i=1,\dots,\ntest$) denote the covariates observations from the validation set. For $i = 1,\ldots,\ntest$, the $i$th observation in the validation set is said to belong to group $k$, if $ \bm{x}_{e,i} \in \hat\gp_{\data_{\ntrain},k}$.

 For the testing stage, let  $\rv_i = y_{e,i} - \hat \prob_{\data_{\ntrain}}(\bm{x}_{e,i})$, $
 \sig_i^2 =  \hat\prob_{\data_{\ntrain}}(\bm{x}_{e,i})\left\{1 -  \hat\prob_{\data_{\ntrain}}(\bm{x}_{e,i})\right\},$ where $i = 1,\dots, \ntest$ and $y_{e,i}$ is the response observation from the validation set, and
$$
 \csum  =  \sum^{K_n}_{k=1}\left(\frac{\sum_{ \{i:\ \bm{x}_{e,i} \in \hat\gp_{\data_{\ntrain},k}\}  } \rv_i}{\sqrt{\sum_{\{i: \ \bm{x}_{e,i} \in \hat\gp_{\data_{\ntrain},k}\}}\sig_i^2}}\right)^2.
$$
 We define the following \textit{p}-value statistic based on the CDF of the chi-squared distribution with degrees of freedom $K_n$:
   \begin{equation}\label{eq_bagoft}
%\bagoft = P\left(\chi^2_{K_n} > \csum \right).
\bagoft = 1- P(\chi^2_{K_n} \leq \csum|\csum, K_n).
 \end{equation}
%   \begin{equation}\label{eq_bagoft}
% %\bagoft = P\left(\chi^2_{K_n} > \csum \right).
% \bagoft = 1- F_{\chi^2}(\csum, K_n),
%  \end{equation}
%  where $F_{\chi^2}(z,k)$ denotes the CDF of $\chi^2_k$ evaluated at $z$.
%where $\sum_{ \{\bm{x}_{e,i} \in \hat\gp_{\data_{\ntrain},k}\} }$ stands for taking the sum with respect to the $i$'s such that $\bm{x}_{e,i} \in \hat\gp_{\data_{\ntrain},k}$. 
%The \textit{p}-value statistic $\bagoft$ can be treated as a \textit{p}-value of the statistic $\csum$. 
We reject $H_0$ when  $\bagoft$ is less than the specified significance level, since 
 $\bagoft$ tends to be small when the discrepancy between $\hat\prob_{\data_{\ntrain}}(\cdot)$ and $\prob(\cdot)$ as quantified by $\csum$ is large.  
 
Compared with the Hosmer-Lemeshow test and other relevant methods, the proposed method allows desirable features such as pre-screening candidate grouping methods (we do not need the Bonferroni correction when considering different groupings), incorporating prior or practical knowledge that is potentially adversarial to the MTA or PTA (e.g., the BAGofT partition can be based on some potentially important variables not in the MTA or PTA), and providing interpretations on the data regions where the MTA or PTA is likely to fail.
The above flexibility often leads to a significantly improved statistical power (elaborated in Section~\ref{gp}). 
It is worth noting that the BAGofT exhibits a tradeoff in data splitting. On the one hand, sufficient validation data used to perform tests can enhance power due to a more reliable assessment of the deviation. On the other hand, more training data enables a better estimation of $\prob(\cdot)$ and the selection of an adversarial grouping that increases power. We will develop theoretical analyses and experimental studies to guide the use of an appropriate splitting ratio.

\subsection{Theory for testing parametric classification models}\label{sub_thparam}

We first establish a theorem that the BAGofT \textit{p}-value statistic converges in distribution to the standard uniform distribution under $H_0$, which asymptotically guarantees the size of the test.
We need the following technical conditions. 

For positive sequences $a_n$ and $b_n$, we write $a_n = \omega(b_n)$ if $a_n/b_n \rightarrow \infty$ as $n\rightarrow \infty$.
\begin{condition}[Sufficient number of observations in each group]\label{gsize}
There exists a positive sequence $\{\underline m_n\}$ such that
$   \min_{k = 1,\ldots,K_n} \sum_{i = 1}^{\ntest}I\{\bm{x}_{e,i} \in \hat\gp_{\data_{\ntrain},k}\}\geq \underline m_n\ a.s.,$
and 
$
\underline m_n = \omega(\ntest^{5/7})
$
 as $n\rightarrow\infty$.

\end{condition}

\begin{condition}[Bounded true probabilities]\label{bddp}
%The support for all continuous covariates $\Support_1$ is compact. Since we assume conditional probability $\prob(\bm{x})$ is continuous with respect to $\bm{x}_c$ given  $\bm{x}_d$, there exists a positive constant $\cst_1<\frac{1}{2}$ such that $\cst_1\leq \prob(\bm{x}_{e,i}) \leq 1-\cst_1$ for all $i = 1,\ldots \ntest$. 
%NOTE: I removed the first part because we also need to assume each discrete alphabet size is bounded by a fixed size and the number of covariates do not depend on n. To avoid complexity, we can simply assume this:
There exists a positive constant $0<\cst_1<1/2$ %$\cst_1<\frac{1}{2}$ 
such that $\cst_1\leq \prob(\bm{x}) \leq 1-\cst_1$ for all $\bm{x}\in\mathbb S$. 
%{\color{red} The minimum number of observations in the groups 
%\[n_{min}  =  \sum_{i = 1}^{\ntest}I\{\bm{x}_{e,i} \in \hat\gp_{\data_{\ntrain},k}\}  \]}
\end{condition}

\begin{condition}[Parametric rate of convergence  under $\textrm{H}_0$]\label{paraC}
Under $H_0$,  $ \underset{\bm{x}\in \Support}{\sup}|\hat\prob_{\data_{n}}(\bm{x})- \prob(\bm{x})| = O_p\left(1/\sqrt{n}\right)\text{ as }n\rightarrow\infty.$
\end{condition}
Condition~\ref{gsize} is mild and can be guaranteed by merging small-sized groups on $\data_{\ntest}$. %(e.g., requiring the minimum number of observations in each group to be at least $\sqrt{\ntest}$).
%An example of such grouping can be found in Section~\ref{gp}.
Condition~\ref{bddp} is a technical requirement so that the Pearson residuals in the theoretical derivations would be bounded, which is satisfied, e.g., under the GLM framework with compact parameter and covariates spaces, and it can be relaxed if more assumptions are made on the tail of the covariate distributions. 
%For brevity, we do not pursue these technical extensions in the paper.
Condition~\ref{paraC} holds for a typical parametric model and a compact set $\Support$. %It will be extended to a general convergence rate in Section~\ref{paraC}.

Throughout the paper, we let $U$ denote the standard uniform distribution.
\begin{theorem}[Convergence of $\bagoft$ for parametric models under $\textrm{H}_0$]\label{wca_p}
Assume that Conditions~\ref{gsize}-\ref{paraC} hold.
Under $H_0$, if $\ntrain,\ntest\rightarrow \infty$ 	and  $\ntest = o(n_1^{3/5})$ as $n\rightarrow \infty$, we have
$
\bagoft \rightarrow_d U.
$
 \end{theorem}

 Accordingly, if we reject the MTA when the BAGofT \textit{p}-value statistic is less than 0.05, we  obtain the asymptotic size 0.05.
%  A possible choice of $\ntest$, for example, is $n_2 = 5\sqrt{n}$.
%  Here, factor 5 is not essential. We used it in the experiments to guarantee a reasonable $n_2$ when $n$ is small.
 %%with size 50 when $n = 100$ in our simulation studies. 
The requirement of $n_1$ and $n_2$ in the above theorem indicates that the number of observations for estimating the parameters and forming groups ($\ntrain$) needs to be much larger than the number for performing tests ($\ntest$). 
Otherwise, the deviation introduced by a random fluctuation due to a small training size (instead of true misspecification)  may be picked up by the BAGofT. It is interesting to note that this data splitting ratio direction is opposite to that for the consistent selection of the best classification procedure via cross-validation (\citealp*{yang2006comparing}; see also \citealp*{yu2014modified}),
%Intuitively, this is because we need the error of the model to assess approximating the true probabilities to be much smaller than the randomness of the statistic.
although other splitting ratios in between have been recommended
for the purpose of tuning parameter  or model selection \citep{bondell2010joint,lei2020cross}. 

%{\color{red}Other works related to the data-splitting include \cite{bondell2010joint} (the method applies $k$-fold cross-validation to select its tuning parameter), \cite{yu2014modified} (the proposed method works when $\ntest/\ntrain\rightarrow\infty$), and \cite{lei2020cross} (its cross-validation method is proved to work in an arbitrary fixed splitting ratio).}

Next, we establish the theorem that shows the BAGofT asymptotically rejects an underfitted model under $H_1$.
\begin{condition}[Convergence  under $\textrm{H}_1$]\label{prpn}
There exists a function $ \prob_a: \Support \rightarrow [0,1]$, which is  not in $\{f(\cdot,\bm{\beta}) \mid \bm{\beta}\in \mathbb B\}$ and allowed to change with $n$, such that   under $H_1$, 
\begin{equation}\label{eq_H1_convergence}
  \underset{ \bm{x}\in \Support}{\text{sup}}|\hat\prob_{\data_{n}}(\bm{x})- \prob_a(\bm{x})|\rightarrow_p 0\ \text{as $n\rightarrow\infty$}.  
\end{equation}
Moreover,
there exists a  constant $0<\cst_2<1/2$ %$\cst_1<\frac{1}{2}$ 
such that $\cst_2\leq \prob_a(\bm{x}) \leq 1-\cst_2$ for $\bm{x}\in \mathbb S$.
\end{condition}

\begin{condition}[Identifiable difference under $\textrm{H}_1$]\label{moddiff}
Under $H_1$, with probability going to one, there exists  $\suset\subseteq\Support$, which may depend on $\data_{\ntrain}$,  such that
\begin{align}
\operatorname*{ess~\inf}_{\bm{x}\in\suset} ( \prob(\bm{x})- \prob_a(\bm{x}) )&\geq \cst, \quad\text{or} \label{eq_larC}\\
\operatorname*{ess~\sup}_{\bm{x}\in\suset} ( \prob(\bm{x})- \prob_a(\bm{x}) )&\leq -\cst,\label{eq_lesC}\
\end{align}
for a positive constant $\cst<1$. We also require that there exists a positive constant $\cst_0<\cst$ such that there is at least one group indexed by $k^*$ with
\begin{equation}\label{eq_groupRatio}
    P(\hat n_{2,k^*}^{\suset}/\hat n_{2,k^*}>(1 + \cst_0)/(1+c))\rightarrow 1,
\end{equation}
as $n\rightarrow\infty$, where $\hat n_{2,k} = \sum_{i = 1}^{\ntest} I{\{\bm{x}_{e,i} \in \hat\gp_{\data_{\ntrain},k}\}}$ denotes the number of  validation observations in the $k$th group, and $\hat n_{2,k}^{\suset} = \sum_{i = 1}^{\ntest} I{\{\bm{x}_{e,i} \in \hat\gp_{\data_{\ntrain},k}\cap \suset\}}$ denotes the number of  validation observations in both the $k$th group and the set $\suset$. 
\end{condition}

%{\color{red}It can be seen that the setting under Conditions~\ref{prpn} and \ref{moddiff} includes the $H_1$ in Section~\ref{Subsec_paramTest} since when the model is not well-specified, the limit $\prob_a(\bm{x})$ will be different from $\prob(\bm{x})$. }
Condition~\ref{prpn} requires the convergence of the model under the alternative.  We can obtain \eqref{eq_H1_convergence} under the  regularity conditions for the convergence of misspecified maximum likelihood estimators   (\cite{white1982maximum}; specifically for GLM, \cite{fahrmexr1990maximum}) . Condition~\ref{moddiff} guarantees that under $H_1$, the deviation between the true model and the fitted model can be captured by the adaptive partition. 
 In particular, \eqref{eq_groupRatio} requires the adaptive selection of a set that contains sufficiently many observations that deviate in the same direction. This is an intuitive and mild condition. The required proportion of observations satisfying \eqref{eq_larC} or \eqref{eq_lesC} is lowed bounded by $1/(1+c)$, which gets smaller when the bias $c$ is larger. We will provide a practical algorithm in Section~\ref{gp} to adaptively search for the most revealing partition according to the Pearson residual (which measures the discrepancy between $\prob_a(\cdot)$ and $\prob(\cdot)$). %which has excellent performance in simulation studies 
Further discussions on how that algorithm meets Condition~\ref{moddiff} are included in the supplement.
Theoretical properties of the algorithm (including the case for assessing general classification procedures) can be found in our discussion section.

%TODO: check the word "Consistency". better to use "asymptotic power"?
\begin{theorem}[Consistency of $\bagoft$ for parametric models under $\textrm{H}_1$]\label{pgo_p}
Suppose that Conditions~\ref{gsize}, \ref{bddp}, \ref{prpn}, and \ref{moddiff} hold. Under $H_1$, if the training and validation sizes satisfy $\ntrain, \ntest\rightarrow \infty$   as $n\rightarrow \infty$, we have
$
\bagoft \rightarrow_p 0,
$
which implies the consistency of the test. 
\end{theorem}
 %It gives desirable performance in simulation studies in Section~\ref{simstu}.

%  \begin{corollary}[Simultaneous size control and consistency for parametric models]
% Assume that Conditions \ref{gsize}-\ref{moddiff} hold, and the data splitting ratio satisfies both $\ntest\rightarrow \infty$ and $\ntest/n_1^{3/4} \rightarrow 0$ as  $n\rightarrow \infty$.
% Then $\bagoft \rightarrow_d U$ under the null hypothesis. If  there is at least one group with $\hat n_{2,k}^{\suset}/\hat n_{2,k}\rightarrow_p 1$ as $n\rightarrow \infty$,  then $\bagoft \rightarrow_p 0$ under the alternative hypothesis.
%  \end{corollary}

In applications, we do not know whether $H_0$ or $H_1$ holds. If we take $\ntrain$ and $\ntest$ such that  $\ntrain,\ntest\rightarrow \infty$ and $\ntest = o(n_1^{3/5})$ as  $n\rightarrow \infty$, under the conditions for Theorems~\ref{wca_p} and \ref{pgo_p} respectively, the BAGofT achieves the desired asymptotic Type I error control and  consistency in power.

\subsection{Theory for assessing general classification procedures}\label{proc}

% \subsection{Theorems for testing general classification procedures}

In this section, we establish properties of the BAGofT for general classification procedures.

\begin{condition}[Convergence at a general rate under $\textrm{H}_0$]\label{prp}
Under $H_0$,  $ \underset{\bm{x}\in \Support}{\sup}|\hat\prob_{\data_{n}}(\bm{x})- \prob(\bm{x})| = O_p\left(\rate_n\right)\text{ as }n\rightarrow\infty,$ with $\rate_n \rightarrow 0$ as $n\rightarrow\infty$.
\end{condition}
%Condition~\ref{prp} generalizes the parametric rate required in Condition~\ref{paraC}.

\begin{theorem}[Convergence of $\bagoft$ under $\textrm{H}_0$ for classification procedures]\label{wca_np}
Under $H_0$, given Conditions~\ref{gsize},\ref{bddp}, and \ref{prp}, if $\ntest\rightarrow \infty$ 	and  $\ntest = o(\rate_{\ntrain}^{-6/5})$ as $n\rightarrow \infty$, we have
$
\bagoft \rightarrow_d U.
$
 \end{theorem}

\begin{condition}[Existence of an identifiable slow converging set under $\textrm{H}_1$]\label{dset}
Under $H_1$, with probability going to one, there exists  $\suset\subseteq\Support$, which may depend on $\data_{\ntrain}$,  such that $\operatorname*{ess~\inf}_{\bm{x}\in\suset} ( \hat\prob_{\data_{\ntrain}}(\bm{x})- \prob(\bm{x}) )\geq 0$ or $\operatorname*{ess~\sup}_{\bm{x}\in\suset} ( \hat\prob_{\data_{\ntrain}}(\bm{x})- \prob(\bm{x}) )\leq 0$, and $\underset{\bm{x}\in \suset}{\inf} | \hat\prob_{\data_{\ntrain}}(\bm{x})- \prob(\bm{x}) |/\rate_{\ntrain}^{(a)} \geq\zeta \text{ almost surely}$,
for a positive constant $\zeta$ and a positive sequence $\rate_{\ntrain}^{(a)}\rightarrow 0$ as $n\rightarrow\infty$. We also require that there is at least one group indexed by $k^*$ with
\begin{equation}\label{eq_groupRatio2}
   \frac{\hat n_{2,k^*} - \hat n_{2,k^*}^{\suset}}{\hat n_{2,k^*}\rate_{n_1}^{(a)}}\limp 0,
\end{equation} as $n\rightarrow \infty$, where $\hat n_{2,k}$ and $\hat n_{2,k^*}^{\suset}$ are defined in  Condition~\ref{moddiff}.% We also require that there is at least one group indexed by $k^*$ that satisfies $\hat n_{2,k^*}^{\suset}/\hat n_{2,k^*}\rightarrow_p 1$  and $(\hat n_{2,k^*}-\hat n_{2,k^*}^{\suset})/(\hat n_{2,k^*}\rate_{\ntrain}^{(a)})\rightarrow_p 0$ as $n\rightarrow \infty$.
\end{condition}

%The above rate of $\rate_{\ntrain}^{(a)}$ will pose a constraint on the splitting ratio in Theorem~\ref{pgo_np}. % for asymptotic consistency.
% We still use the notation $\suset$ since it is a generalization of the set in Condition~\ref{moddiff}. 
%We further let $\hat n_{2,k}^{\suset} = \sum_{i = 1}^{\ntest} I{\{\bm{x}_{e,i} \in \hat\gp_{\data_{\ntrain},k}\cap \suset\}}$ denote the number of observations in both the $k$th group and the slow converging set $\suset$.
\begin{condition}[Bounded predicted probability]\label{bddprob}
There exists a constant $0<\cst_3<1/2$ %$\cst_1<\frac{1}{2}$ 
such that $\cst_3\leq \hat\prob_{\data_{n}}(\bm{x}) \leq 1-\cst_3$ almost surely for $\bm{x}\in\mathbb S$ and for all $n$.  
\end{condition}

Condition~\ref{dset} requires the existence of an identifiable region where $\hat\prob_{\data_{n}}(\cdot)$ from the PTA converges slowly (or not at all) to the data generating $\prob(\cdot)$ as $n \rightarrow \infty$. Further discussions on the theoretical guarantee to identify an $\suset$ in Condition~\ref{dset} are included in the supplement.
% Condition~\ref{bddprob} is for technical convenience.  %the proof by directly requiring the predicted probabilities from the model to assess to be bounded away from 0 and 1. 
% It can be replaced with the uniform convergence of $ \hat\prob_{\data_{n}}(\cdot)$ to a function $\prob_a(\cdot)$ in conjunction with the same boundedness constraint on that limiting function.

For positive sequences $a_n$ and $b_n$, we write $a_n = \Omega(b_n)$ if there exists $C>0$, such that $a_n/b_n \geq C$.
\begin{theorem}[Consistency of $\bagoft$ under $\textrm{H}_1$ for classification procedures]\label{pgo_np}
Under the alternative, assume that Conditions~\ref{wca_p}, \ref{bddp}, \ref{dset}, and \ref{bddprob} hold,  $\ntest\rightarrow \infty$, and 
$\ntest = \Omega((\rate_{\ntrain}^{(a)})^{-6})$ as $n \rightarrow \infty$.
 Then, we have
$
\bagoft \rightarrow_p 0
$
as $n\rightarrow \infty$.
%In case there exists a constant $\overline K>0$ such that $P(K_n<\overline K)\rightarrow1$, we can relax the requirement $1/\rate_{\ntrain}^{(a)} = O(\ntest^{1/4})$ to $1/\rate_{\ntrain}^{(a)} = O(\ntest^{3/4})$.
\end{theorem}
The theorem shows that the BAGofT can flag a slow converging classification procedure when there is sufficient validation data.
% Additionally, if the number of groups $K_n$ chosen by the adaptive partition is upper bounded by a fixed constant, we can relax the 
% requirement $\ntest^{5/16} \rate_{\ntrain}^{(a)}\rightarrow \infty$ to $\ntest^{3/8} \rate_{\ntrain}^{(a)}\rightarrow \infty$. In this way, the test can be asymptotically powerful for a $\rate_{\ntrain}^{(a)}$ that converges faster to $0$, which means that we are able to identify a smaller deviation.   
Theorems \ref{wca_np} and \ref{pgo_np} imply the following corollary. 
%  \begin{corollary}[Obtaining both size control and consistency for learning procedures]\label{lpsc}
%  Let $\xi_n = \rate_{\ntrain}^{(a)}/\rate_{\ntrain}$. Assume that $\xi_n = \omega(\rate_{\ntrain}^{-4/5})$, and 
%  Conditions~\ref{wca_p},  \ref{bddp}, \ref{prp}, \ref{dset}, and \ref{bddprob} hold. If we take $\ntest = \Omega\bigl(({\rate_{\ntrain}^{(a)}})^{-6}\bigr)$ and $\ntest = o(\rate_{\ntrain}^{-6/5})$, we have $\bagoft \rightarrow_d U$ as $n\rightarrow \infty$ under $H_0$ and the asymptotic consistency of the BAGofT under $H_1$.
%  \end{corollary}
%  \begin{corollary}[Obtaining both size control and consistency for learning procedures]\label{lpsc}
% Assume that $\rate_{\ntrain}^{(a)} = \omega(\rate_{\ntrain}^{1/5})$, and 
%  Conditions~\ref{wca_p},  \ref{bddp}, \ref{prp}, \ref{dset}, and \ref{bddprob} hold. If we take $\ntest = \Omega\bigl(({\rate_{\ntrain}^{(a)}})^{-6}\bigr)$ and $\ntest = o(\rate_{\ntrain}^{-6/5})$, we have $\bagoft \rightarrow_d U$ as $n\rightarrow \infty$ under $H_0$ and the asymptotic consistency of the BAGofT under $H_1$.
%  \end{corollary}
 \begin{corollary}[Obtaining both size control and consistency for learning procedures]\label{lpsc}
Assume that $\rate_{\ntrain}^{(a)} = \Omega(\rate_{\ntrain}^{\cst^*})$ as $n\rightarrow\infty$ with $0<\cst^*<1/5$  and 
 Conditions~\ref{wca_p},  \ref{bddp}, \ref{prp}, \ref{dset}, and \ref{bddprob} hold, respectively. If we take $\ntest = \Omega\bigl(\rate_{\ntrain}^{-6\cst^*}\bigr)$ and $\ntest = o(\rate_{\ntrain}^{-6/5})$  as $n\rightarrow\infty$, we have $\bagoft \rightarrow_d U$ as $n\rightarrow \infty$ under $H_0$ and the asymptotic consistency of the BAGofT under $H_1$.
 \end{corollary}
 
%The requirement $\ntest = \Omega\bigl(({\rate_{\ntrain}^{(a)}})^{-6}\bigr)$ in the  corollary implies that for a given $\rate_{\ntrain}$, a larger feasible $\ntest$ will allow the detection of  a wider range of $\rate_{\ntrain}^{(a)}$.

% Corollary~\ref{lpsc} implies the following.
% First, if the PTA converges very slowly under $H_1$, namely $\rate_{\ntrain}^{(a)}$ is large, it is more flexible to choose $\ntest$.
% Second, the convergence of the PTA under $H_0$ needs to be significantly faster than that under $H_1$, namely $\xi_n = \omega(\rate_{\ntrain}^{-4/5})$, for the BAGofT to reject under $H_1$. 
% %guarantees the existence of an eligible $\ntest$. %such that $\ntest = \Omega((\rate_{\ntrain}\xi_n)^{-6})$ and $\ntest = o(\rate_{\ntrain}^{-6/5})$.
% Also, if we view $\rate_{\ntrain}$ as fixed, a larger feasible $\ntest$ will allow the detection of  a wider range of $\rate_{\ntrain}^{(a)}$.
%This corollary indicates that with a larger $\ntest$, the BAGofT can identify a smaller deviation between $\rate_{\ntrain}$ and $\rate_{\ntrain}^{(a)}$, while controlling the Type I error. Moreover, when the PTA does not converge, i.e., $\xi_n = \Omega(\rate_{\ntrain}^{-1})$, we can drop the requirement $\ntest = \Omega((\rate_{\ntrain}\xi_n)^{-6})$, so each $\ntest$ such that $\ntest\rightarrow\infty$ and $\ntest = o(\rate_{\ntrain}^{-6/5})$ works.

%%%%%
% Note: the rates are taken square roots since the original paper considers the rate of the squared L2 norm difference.
For example, suppose that the PTA is a neural network-based method and the number of covariates $p>46$. Also suppose under $H_0$, $\prob(\cdot)$ admits a neural network representation, and under $H_1$, $ \prob(\cdot)$ is in the Besov class with the smoothness parameter $\alpha=2$ (details about the two classes of functions can be found in \citealp*{yang1999minimax}).
According to \cite{yang1999minimax}, typically we have $\rate_{n}=O((n/\log n)^{-(p+1)/(4p+2)})$ and $\rate_{n}^{(a)} = n^{-2/(4 + p)}$. Then, $\rate_{n}=O(n^{-1/4})$ and $\rate_{n}^{(a)}=\Omega(n^{-1/25})$, so $\rate_{n}^{(a)} = \Omega(\rate_{n}^{4/25})$. 
If we set  $\ntest$, e.g., of the order $\ntrain^{24(p+1)/(25(4p+2))}$, given the other required conditions for Corollary~\ref{lpsc}, 
the BAGofT asymptotically controls the Type I error probability under $H_0$ and rejects $H_0$ with probability going to one under $H_1$.

% Let $N(C)$ (neural network class) be the closure in $L_2[0,1]^d$ of the set of all functions  $g:[0,1]^d\rightarrow [0,1]$ of the form
% $$g(\bm{x}) = c_0 + \sum_i c_i s(\nu_i\cdot \bm{x} + b_i),$$
% where $|c_0| + \sum_i |c_i|\leq C$, $|\nu_i| = 1$ and $s(\cdot)$ is the step function with 
% $$s(t) = 
% \begin{cases}
% 0,\text{ for } t<0\\
% 1,\text{ for } t\geq0.
% \end{cases}$$
% Let $B^{\alpha}_{\sigma,q}(C)$ be the Besov class \citep{triebeltheory}. 

\section{Practical Guidelines for Implementing the BAGofT}\label{sec_pracguid}
%We emphasize that unlike previous results in the literature, 
Unlike previous methods in the literature, our approach allows the number of groups to be adaptively chosen, and it may grow when finer partitions are needed to pinpoint the poorly fitted regions.
We recommend setting the largest allowed number of groups to $K_{max} =  \sqrt{\ntest} $ as a default choice, where $\lfloor a \rfloor$ denotes the largest integer less than or equal to $a$.  We suggest $\ntest = 5\sqrt{n}$ for the training-validation splitting for testing parametric models, which can guarantee enough validation size when $n \geq 100$.  In this way, $K_{max}\rightarrow\infty$ as $n\rightarrow\infty$, despite that the selected $K_n$ may be small.
%Details about the adaptive partition that applies the above setting is given in Section~\ref{gp}.
Our experimental results in Section~\ref{simstu} and the supplementary material show the desirable performance of the default choices under both $H_0$ and $H_1$.

\subsection{Splitting ratios and interpretations in assessing general classification procedures}\label{subsec_splitratio}
This subsection includes more details on how to assess the GOF of classification procedures.
In practice, the convergence rate $\rate_n$ under $H_0$  may not be known. Moreover, when the sample size is finite, the convergence rate $\rate_{\ntrain}$ provides limited insight on selecting a suitable splitting ratio.
For practical implementations, we advocate considering three splitting ratios where the training set takes $90\%$, $75\%$, and $50\%$ of the observations, respectively.
%and the validation set takes the remaining part, respectively. %We refer to these ratios by $90\%$, $75\%$, and $50\%$, respectively.
%{\color{blue}Note that the above three ratios are inspired by the theory, and designed to obtain easy interpretations in practice.}
The four typical results are given as follows.

\textbf{Pattern 1:} The BAGofT fails to reject $H_0$ under all the three splitting ratios. The conclusion is that the classification procedure converges quite fast to the underlying conditional probability function, and there is little concern about the lack of fit.

\textbf{Pattern 2:} The BAGofT rejects  $H_0$ only at  $50\%$ training. The conclusion is that the classification procedure converges moderately fast, and the procedure fits the data well.

\textbf{Pattern 3:} The BAGofT rejects $H_0$ at both $50\%$ and $75\%$ and fails to reject at $90\%$. The conclusion is that the classification procedure converges slowly, but the current sample size is most likely enough for the procedure to fit the data properly.

\textbf{Pattern 4:} The BAGofT rejects $H_0$ under all the three splitting ratios. The conclusion is that the classification procedure fails to capture the nature of the data generating process.

A caveat is that there may exist ``boundary'' cases where the $90\%$ training set is still insufficient for the PTA to work well, but the $10\%$ validation set is not enough to identify the weakness of the PTA. In such cases, the failure of rejection may not necessarily be reliable. In general, the BAGofT may have a low power when there is not enough validation data. When the $10\%$ validation set is perceived possibly too small, one possible solution is adding a splitting ratio, e.g., 80\%, in order to offer more information. Also, note that if $\hat \prob(\cdot)$ from the PTA is very sensitive to the sample size and data perturbation, we may fail to observe the gradual change of the rejection results as listed in \textbf{Patterns}~\textbf{1-4}. Since unstable procedures are not really reliable anyway, we recommend applying proper stabilization methods to improve the procedure fit first.
% In this case, the sample size may not be large enough to support the procedure's complexity, and it may not work well stably with the limited data.

% In that case, we suggest the following conclusion.
% \begin{description}
% \item[Pattern 5] The rejection result does not belong to \textbf{Pattern 1} - \textbf{Pattern 4}. The outcome from different sample sizes is not in a sensible order. 
% \end{description}
% %In this case, the performance of the general classification procedure highly depends on the sample size and may even get worse performance when given more training data. 
% To enhance the stability of $\hat \prob(\cdot)$, we can apply the same general classification procedure multiple times and obtain the averaged predictions.

\subsection{Adaptive partition for the BAGofT}\label{gp}

The asymptotic theory of the BAGofT from the earlier section requires a grouping scheme based on the training set that asymptotically reveals at least one region with $\hat\prob_{\data_{n}}(\cdot)$ converging slowly or not converging to $\prob(\cdot)$. In this section, we introduce an adaptive grouping algorithm that may efficiently discover such a region.

The idea of the adaptive grouping is that instead of applying one prescribed partition, we select a partition from a set of partitions based on the training data $\data_{\ntrain}$. 
%The selected partition is independent of the test data $\data_{\ntrain}$.
According to Theorems~\ref{wca_p} and~\ref{wca_np}, while protecting the size of the test, we have much flexibility to adaptively select a grouping rule (including the number of groups $K_n$) as long as it is independent of $\data_{y_e}$ conditional on $\data_{x_e}$. 
Meanwhile, with the adaptive grouping, the power under $H_1$ is expected to be high.
 
One way to find a partition to exploit the regions of model misspecification is to fit the deviations (e.g., Pearson residuals) using a nonparametric regression method and choose a partition based on the fitted values. Then, we group the observations with large positive deviations and those with large negative deviations into separate groups to calculate the statistic $\csum$ for the BAGofT and consequently avoid their cancellation.

In particular, we develop a Random Forest-based adaptive partition scheme as the default choice in our R package `BAGofT.' It shows excellent performance in our simulation studies. The procedure of the scheme is outlined as follows.
%In the next section, we will introduce a Random Forest-based adaptive grouping scheme. We will show its excellent performance in our simulation studies. We have developed this method into our open-source \textit{R} package `BAGofT.' %available from \url{https://github.com/JZHANG4362/BAGofT} and currently under the inspection of CRAN.
%\subsection{A practical scheme: Random Forest type adaptive partition}\label{parAlg}
On the training set, we first apply the MTA or PTA. We then fit a Random Forest on the training set Pearson residuals and obtain fitted values $\hat q_i^{(1)}$, $i = 1,\ldots,\ntrain$. 
%Let $\lfloor a \rfloor$ denote the largest integer no larger than $a$.
For different numbers of groups $K = 1,\ldots, K_{max}$, where $K_{max}>2$, we partition $[0,1]$  into  intervals $\bigl\{\gp^{(K)}_1,\ldots  \gp^{(K)}_K \bigr\}$ by the  $K$-quantiles of $\{\hat q_i^{(1)}\}_{i=1}^{\ntrain}$, and calculate the statistic 
  \begin{equation}
  \label{bq}
\mathcal B_K = \sum^K_{k=1}\left(\frac{\sum_{ \{i:\ \hat q_i^{(1)} \in \gp^{(K)}_k\}  } (y_{t,i} - \hat \prob_{\data_{\ntrain}}(\bm{x}_{t,i}) ) }{\sqrt{\sum_{\{i:\ \hat q_i^{(1)} \in \gp^{(K)}_k \}}   \hat \prob_{\data_{\ntrain}}(\bm{x}_{t,i})\left(1 -  \hat \prob_{\data_{\ntrain}}(\bm{x}_{t,i})\right)  }} \right)^2
 \end{equation}
 using the training set. We choose the partition $\bigl\{\gp^{(K_n)}_1,\ldots  \gp^{(K_n)}_{K_n} \bigr\}$ where $K_n$ is the $K\in 2,\ldots,K_{max}$ that maximizes $\mathcal B_{K} - \mathcal B_{K-1}$.
 The pseudocode is summarized in  Algorithm~\ref{algo_partition}.
 
 Next, we obtain the Random Forest prediction on the validation set $\hat q_i^{(2)}$ ($i = 1,\ldots,\ntest$). We calculate 
 \begin{equation*}
 \csum  =  \sum^{K_n}_{k=1}\left(\frac{\sum_{ \{i:\ \hat q_i^{(2)} \in \gp^{(K_n)}_k\}  } \rv_i}{\sqrt{\sum_{\{i: \ \hat q_i^{(2)} \in \gp^{(K_n)}_k\}}\sig_i^2}}\right)^2,	
 \end{equation*}
 and then, the \textit{p}-value statistic $\bagoft$ in \eqref{eq_bagoft}. 
 
%\State\Return $\bagoft$.

 Note that we may use a set of covariates different from those in the MTA or PTA when applying the Random Forest learning. 
For example, we can apply a variable screening to drop some covariates before fitting the classification model or procedure to obtain a parsimonious model or stabilize the fitting algorithm. In this case, our Random Forest-based adaptive partition may consider all the available covariates to check the GOF. This algorithm also provides some insights on possible misspecifications via the Random Forest variable importance. Since the Random Forest is fitted on the Pearson residual of the MTA or PTA, \textit{variables with larger importance are more likely to be associated with the misspecifications}.  
%A caveat is that some learning methods may return fitted probabilities close to 0 or 1, causing issues to calculate the Pearson residual. If that occurs, we may replace the Pearson residual with the raw residual. 
More details and related simulations for this algorithm are included in the supplement.

% Simulations in Section~\ref{ldsim} and Section~\ref{thmodel} in our supplementary material show that a pre-selection
% of size 5 from 500 covariates may significantly improve the power of the BAGofT.

  \begin{algorithm}
\caption{\label{algo_partition}
A default choice of BAGofT adaptive partition}
\begin{algorithmic}[1]
\Procedure{Partition}{$\data_{\ntrain}, K_{max}, parVar$}  
\Comment{{\footnotesize $parVar$ is the set of variables to construct the partition, which can be different from those in the MTA or PTA (see Section~\ref{gp}).}}     
\State Fit the MTA or PTA on the set $\data_{\ntrain}$ and calculate the Pearson residual.
\State Fit a Random Forest on the Pearson residual with respect to the partition variables $parVar$ and obtain the fitted value on the training set $\{\hat q_i^{(1)}\}_{i=1}^{\ntrain}$ %and predicted value on the validation set $\{\hat q_i^{(2)}\}_{i=1}^{\ntest}$.
 \For{$K$ in $1,\ldots, K_{max}$}
 \State Partition $[0,1]$ by $K$-quantiles of $\{\hat q_i^{(1)}\}_{i=1}^{\ntrain}$ into  $\bigl\{\gp^{(K)}_1,\ldots  \gp^{(K)}_K \bigr\}$.
 \State Calculate $\mathcal B_K $ in Equation~\eqref{bq}.  
 \EndFor
 \State $K_n \leftarrow \arg\max_{K=2,\ldots,K_{max}} (\mathcal B_{K} - \mathcal B_{K-1})$.
\State\Return
$\bigl\{\gp^{(K_n)}_1,\ldots  \gp^{(K_n)}_{K_n} \bigr\}$.
% \Comment{{\footnotesize Select the number of groups $K_n$ based on the increment.}}
% \State Calculate the test statistic $\bagoft$ in Equation~\eqref{eq_bagoft} based on $\{\hat q_i^{(2)}\}_{i=1}^{\ntest}$ and the selected partition interval $\bigl\{\gp^{(K_n)}_1,\ldots  \gp^{(K_n)}_{K_n} \bigr\}$.
%\State\Return $\bagoft$.
\EndProcedure
\end{algorithmic}
\end{algorithm}

In high dimensional settings with many covariates, we have found that a pre-selection used to reduce the number of covariates for the adaptive grouping can help the test performance and save computing cost. We rank the covariates by the distance correlation  \citep{szekely2007measuring} that measures the dependence relation between the Pearson residual and the covariates, and keep the top ones. More details can be found in the supplement.

\subsection{Combing results from multiple splittings}\label{praci}

% CONTEXT
Recall that our test is based on splitting the original data into training and validation sets. 
Due to the randomness of data splitting, we may obtain different test results from the same data.
To alleviate this randomness, we can randomly split the data multiple times and appropriately combine the test result from each splitting. %Our motivation is to reduce the test result variance.

We propose the following procedure.
First, we randomly split the data into training and validation sets multiple times and calculate the \textit{p}-value statistic defined in~\eqref{eq_bagoft};
Second, we calculate the sample mean of the \textit{p}-value statistic values.
Other ways to combine results from multiple splittings include taking the sample median or minimum of the \textit{p}-value statistic values. It is challenging to derive the theoretical distribution of the statistics from the combined results. Thus, we evaluate the obtained statistic using the bootstrap \textit{p}-values. 

The bootstrap \textit{p}-value is based on parametric bootstrapping. First, we fit the model using all the data and obtain the fitted probabilities. Second, we generate some bootstrap datasets from the Bernoulli distributions with those fitted conditional probabilities. Third, we calculate the \textit{p}-value statistic on each of the bootstrap datasets, so these \textit{p}-value statistics correspond to the case where the MTA or PTA is `correct.' 
Fourth, we compare the \textit{p}-value statistic from the original data with those from the bootstrap datasets and calculate the bootstrap \textit{p}-value. 

%  In practice, if the model or procedure takes a lot of time to fit or requires manual tuning, this approach with multiple splitting may be undesirable. So we recommend applying the BAGofT based on a single splitting. If the data comes with a validation set (for calculating the classification accuracy), it is desirable for us to apply the same splitting directly. In this way, the test result and the classification accuracy are based on the same validation set, facilitating the interpretations.

 %can be more related compared to the results from different splitting ratios.

\section{Experimental Studies}\label{simstu}

In the following subsections, we present simulation results to demonstrate the performance of the BAGofT in various settings.

In Section~\ref{classSet}, we check the performance of the BAGofT in parametric settings and compare it with some existing methods, including the recently proposed Generalized Residual Prediction (GRP) test  \citep{jankova2019goodness}.  The GRP  calculates a test statistic by pivoting the Pearson residuals from the MTA. 
It has a different focus compared with the BAGofT. First, the GRP test works for generalized linear models (GLM). In contrast, the BAGofT tests general classification models, e.g., linear discriminant models and naive Bayes models that do not belong to GLM. Secondly, the GRP test focuses on the cases where the link function of the generalized linear model is correctly specified. The BAGofT can have power against a general deviation of the MTA from the truth. Additionally, when covariates outside the MTA are considered (as mentioned in Section~\ref{gp}), the GRP test requires the true model to have the linear effects of these covariates only; the BAGofT can test on other misspecifications, including missing quadratic effects and interactions of the missed covariates. For comparison, we only choose simulation settings that work for both the GRP and BAGofT in this part.  A discussion about the required conditions for the BAGofT in the experimental settings is included in the supplement.

In Section~\ref{procSim_high}, we demonstrate the application of the BAGofT to assess general classification procedures, where we are not aware of any method to compare with. 
%The outcomes from classification procedures with different convergence speeds illustrate the suggested interpretations from Section~\ref{subsec_splitratio}. 

%%%%%%%%%%%%%%%%%%%%%%%%%%%%%%%%%%%%%%%%%%%%%%%
% code can be found in Rcodes137_adaptive_K_std,
% Rcodes138_rejection_plots, Rcodes92_adaptive_K_std_others
% Rcodes118_adaptive_K_std_extra (10 covariates, abandoned)
%%%%%%%%%%%%%%%%%%%%%%%%%%%%%%%%%%%%%%%%%%%%%%%
\subsection{Testing on parametric models}\label{classSet}
We choose some commonly studied parametric settings that are similar to those in \cite{pulkstenis2002two,yin2013pearson,canary2017comparison}. % There are no external covariates in these settings, so the available covariates to apply the BAGofT are the same as those used to generate the response data.
 
%Three settings are as follows.
\textbf{Setting 1.} The response is generated from 
$P(y = 1|x_1,x_2,x_3) = 1/(1 + \exp(-(\beta_1x_1 + \beta_2x_2 + \beta_3x_3))),$
where $x_1$, $x_2$, and $x_3$ are independently generated from $\textrm{Uniform}[-3,3]$,     $\N(0,1)$, and $\chi^2_4$,  respectively. We test the correctly specified model (named Model \textit{A}) and the model that misses~$x_3$ (named Model \textit{B}).

\textbf{Setting 2.} The response is generated from 
$P(y = 1|x_1,x_2) = 1/(1 + \exp(-(\beta_1x_1 + \beta_2x_2 + \beta_3x_1x_2 ))),$
where $x_1$ and  $x_2$ are independently generated from $\textrm{Uniform}[-3,3]$.  We test the correctly specified Model \textit{A} and Model \textit{B} that misses the interaction term.

\textbf{Setting 3.} The response is generated from 
$P(y = 1|x_1,x_2,x_3) = 1/(1 + \exp(-(\beta_1x_1 + \beta_2x_2 + \beta_3x_3  + \beta_4 x_1^2 ))),$
where $x_1$, $x_2$, and $x_3$ are independently generated from $\textrm{Uniform}[-3,3]$,     $\N(0,1)$, and $\chi^2_2$, respectively. We test the correctly specified Model \textit{A} and Model \textit{B} that misses the quadratic term.

For Model \textit{A}, we check the null distribution of the BAGofT statistic. For Model $B$, we compare the  power of the BAGofT with the Hosmer-Lemeshow test \citep{hosmer1980goodness},  le Cessie-van Houwelingen (CH) test \citep{le1991goodness}, and GRP test \citep{jankova2019goodness}.  These three tests are fitted by %{\em hoslem.test()} from
packages {\em ResourceSelection} \citep{ResourceSelection}, {\em rms} \citep{rms}, and {\em GRPtests} \citep{GRP}, respectively, with their default values.  The BAGofT applies 40 data splittings, with all the available covariates considered for the adaptive partition, namely $(x_1,x_2, x_3)$, $(x_1,x_2)$, and $(x_1,x_2,x_3)$ in Settings 1-3, respectively.

To avoid cherry-picking, we independently generate the coefficients from normal distributions with unit standard deviation. Coefficients $\beta_3$ in Setting 1 and Setting 2, and $\beta_4$ in Setting 3 are generated with mean $1$ and others are generated with mean $0$. To reflect different degrees of deviation of the MTA from the data generating distribution when testing Model \textit{B}, we consider an additional setting with standard deviation $0.5$  for those coefficients generated with mean $1$. The other coefficients remain the same as before. The considered sample sizes are 100, 200, and 800, and the testing process in each setting is independently replicated 100 times.

\begin{figure}[!ht]
  \caption{The Q-Q plot of  the BAGofT bootstrap \textit{p}-values from Model \textit{A} versus $\textrm{Uniform}[0,1]$
   distribution in Setting 1. The \textit{x}-axis and \textit{y}-axis correspond to the  theoretical quantiles and observed sample quantiles, respectively. The red straight line corresponds to the perfect match between the theoretical and observed sample quantiles. }\label{qplot}
  \centering
   \includegraphics[width=1\textwidth]{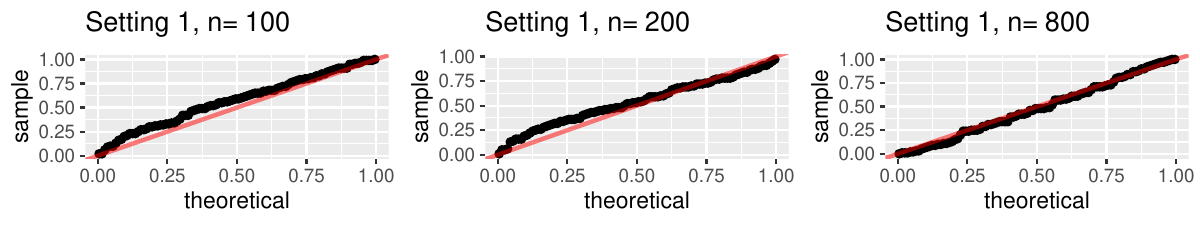}
\end{figure}

\begin{figure}[!ht]
    \caption{The rejection rates of tests  for Model \textit{B} in Settings 1-3. We take standard deviation $\gamma = 1$ or $0.5$ for $\beta_3$ in Setting 1, Setting 2, and $\beta_4$ in Setting 3, respectively. A smaller $\gamma$ makes it harder to reject.  The BAGofT is compared with the HL,   CH, and GRP tests. The significance level is 0.05. }\label{power} 
  \centering
    \includegraphics[width=1\textwidth]{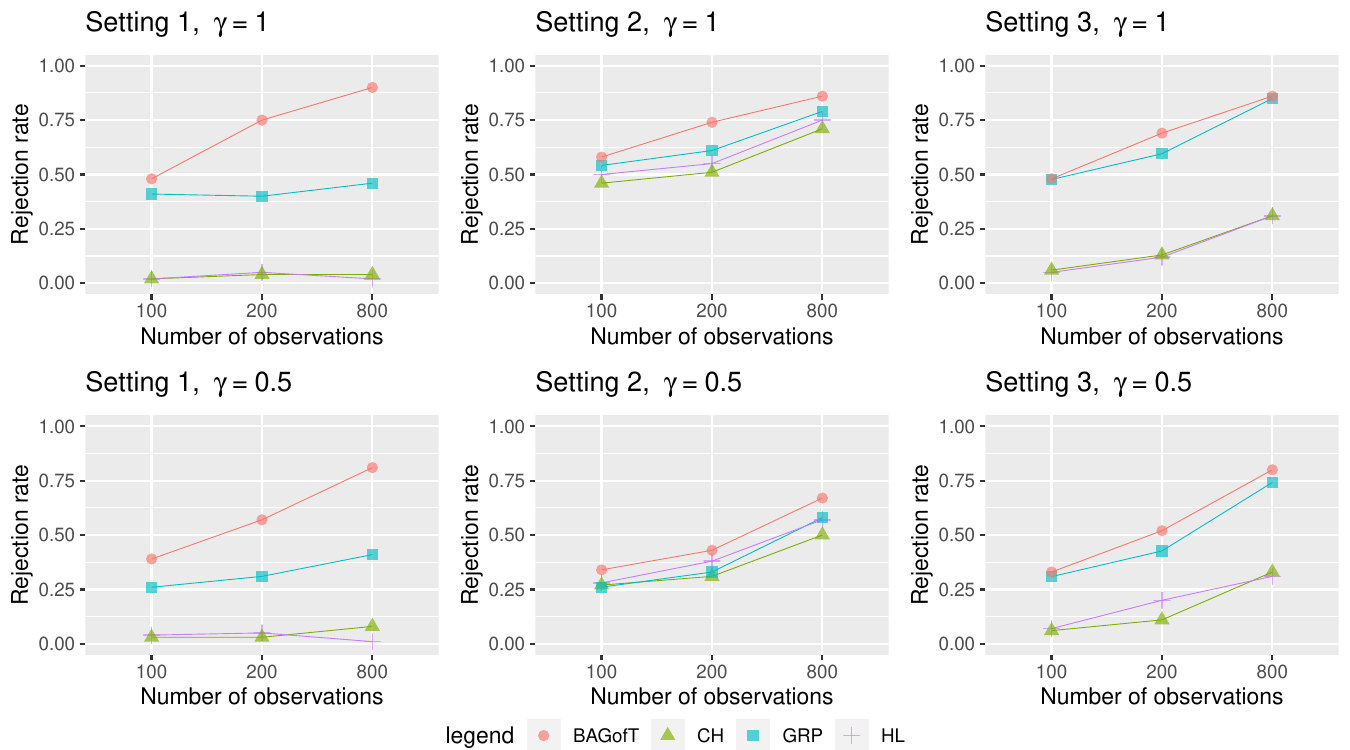}
\end{figure}

The BAGofT results with the three ways to combine multiple splitting results in Section~\ref{praci} (namely, those based on mean, median, and minimum, respectively) are very close. We thus only present those based on the mean. For Model \textit{A}, the Q-Q plots of the BAGofT \textit{p}-value statistic against $\textrm{Uniform}[0,1]$ in Setting 1  are shown in Figure~\ref{qplot}.  We observe that in general, the statistic has a good approximation to $\textrm{Uniform}[0,1]$ under $H_0$.  When the sample size is small, the simulated Type I error tends to be less than nominal.     The results of the other settings are included in the supplementary material, and they show similar results. For Model \textit{B}, the rejection rates of the BAGofT compared with the other tests at the significance level of 0.05 are shown in Figure~\ref{power}.  Due to the random generation of the coefficients, a small portion of the datasets is unbalanced. It caused computation errors for the CH and GRP tests. We dropped these cases when computing the rejection rates.
From the results in Figure~\ref{power}, the BAGofT (in circles) has the best performance in all of the cases. The GRP test (in squares) gets close to the BAGofT in Settings 2 and 3.

We also study the relationship between the number of splittings and the variation of the BAGofT \textit{p}-value statistic. Recall that the purpose of multiple splitting is to obtain a test statistic with smaller variation. The results show that 10 to 20 splittings are usually good enough to get stable results. Additionally, we check the covariates with the largest variable importance (from the Random Forest fitted on the Pearson residuals) in Settings 1 and 3 when the models are misspecified (Model $B$). Recall that the covariates with large variable importance tend to be the major source of misspecification. 
Most of the times in our simulation, the missing variable $x_3$ in Setting 1 has the largest variable importance; $x_1$ in Setting 2, whose quadratic effect is missing, has the largest variable importance.  
Additional experimental details on the variations of the statistics and the variable importance are included in the supplementary material.

%Rcodes148_procedures
\subsection{Assessing classification learning procedures}\label{procSim_high}

In this subsection, we demonstrate the application of the BAGofT to assess classification procedures. We focus on a high dimensional setting with 1000 covariates and a sample size of 500. A low dimensional study is included in the supplement. The response is generated by the Bernoulli distributions with the following settings.
\begin{align*}
 \textbf{Setting 1:}\ \quad P(y = 1|x_1,\ldots, x_{1000}) =& 1/(1 + \exp(-(-6 + 3 \cdot I\{-2 < x_1 < 2\} +\\
 &0.5  (x_2 + x_3 + x_4 + x_5) ))). \\
    \textbf{Setting 2:}\ \quad 
    P(y = 1|x_1,\ldots, x_{1000}) =& 1/(1 + \exp(-(0.5 x_1 + 0.3 x_2 + 0.1 x_3 + 
                0.1 x_4 + 0.1 x_5) )).
\end{align*}
The covariates $x_1,\ldots, x_{1000}$ are independently generated from $\textrm{Uniform}[-5,5]$. The PTAs are  the logistic regression with LASSO penalty, Random Forest, and XGBoost \citep{chen2016xgboost}. 

\begin{figure}[!ht]
  \centering
  \includegraphics[width=1\textwidth]{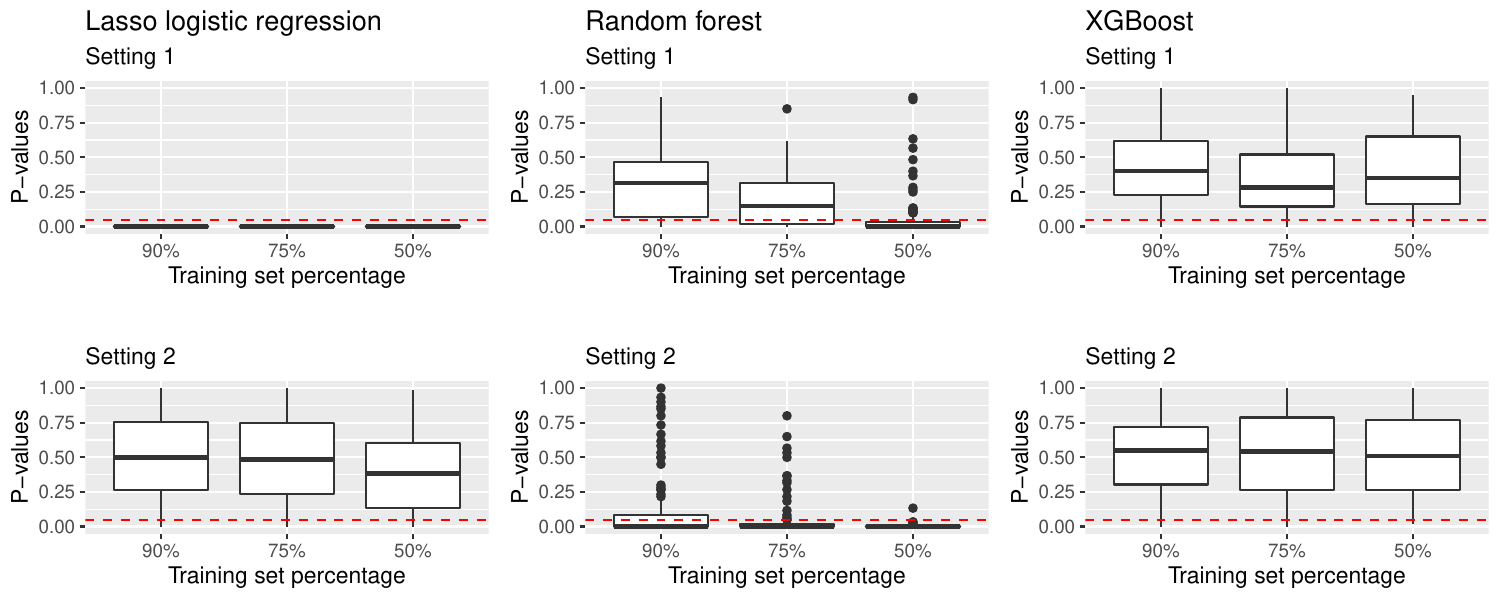}
  \caption{The BAGofT \textit{p}-value box plots  in the high dimensional settings. The red dashed lines correspond to the 0.05 significance level. }\label{fig_hdprocedure}
\end{figure}

We first randomly generate the sample data and apply the BAGofT with the three splitting ratios to the PTAs. We apply 20 data splittings, and the adaptive partition is based on all the available covariates $x_1,\ldots,x_{1000}$. The Random Forest is fitted by the package \textit{randomForest} \citep{randomForest} with maximum nodes 10. The XGBoost is fitted by the package \textit{xgboost} \citep{xgboost} with 25 iterations. The above process is performed with 100 replications, and the results are summarized in Figure~\ref{fig_hdprocedure}.

The result of the LASSO logistic regression in Setting 1 belongs to \textbf{Pattern 4} since the LASSO logistic fails to capture the nonlinearity in the data-generating model. 
For Setting 2, it belongs to \textbf{Pattern 1} (converging quite fast). 
The Random Forest has moderate fast or slow convergence speed (\textbf{Pattern 2} or \textbf{Pattern 3}) in Setting 1. It has a slow convergence speed (\textbf{Pattern 3}) or fails to capture the nature of the data generating process (\textbf{Pattern 4}) in Setting 2. The Random Forest's overall slow convergence is because its single trees are fitted on some small subsets of the available covariates. As a result, it tends to miss important signals in the sparse setting. The XGBoost converges quite fast (\textbf{Pattern 1}) in both settings. 

\section{Real Data Example}\label{realExp}
In the following three subsections, we demonstrate the application of the BAGofT by real-world data examples. 
In Section~\ref{sub_MRNA}, we test a parametric classification model and compare the BAGofT with other methodologies. In Section~\ref{sub_MNIST}, we present a graphical illustration on how the adaptive partition brings an insight on which variables may be responsible for the deficiency of the procedure.  In Section~\ref{sub_COVID}, we apply the BAGofT to assess three classification procedures. We take 20 data splittings and pre-selection size 5 (see Section~\ref{gp}) for the BAGofT throughout this section. The significance level is 0.05.

% Rcodes113_realdata Comparison3.R
\subsection{Tesing parametric classification models: Micro-RNA data}\label{sub_MRNA}
We consider the study of \citet{shigemizu2019risk}, where the data  is available from the
Gene Expression Omnibus (GEO) database with accession number GSE120584. 
They fitted logistic regressions on micro-RNA data to predict several dementias. Our study focuses on the model that predicts whether a subject has Alzheimer's disease (AD) or not. 
% The data for the AD model are obtained by combining the data of AD and normal controls. 
The data contain $n = 1309$ observations.
\citet{shigemizu2019risk} selected 78 micro-RNA and computed 10 principal components from the data to fit the prediction model for AD.
We first consider a subset model using the first 7 principal components  as the covariates.
\begin{equation}\label{AD1}
\textbf{Model 1: }\quad\log\left(\frac{p}{1-p}\right) = \beta_0 + \beta_1 \textrm{PC}_1 +  \ldots +  \beta_7 \textrm{PC}_7.
\end{equation}
The available covariates for the BAGofT are the first 20 principal components $\textrm{PC}_1, \ldots, \textrm{PC}_{20}$. 
 The bootstrap \textit{p}-value of the BAGofT is 0. The averaged (Random Forest) variable importance shows that $\textrm{PC}_9$ has the largest importance value and is likely to be the major reason for the underfitting.

Next, we add $\textrm{PC}_9$ to the model and consider:
\begin{equation}\label{AD2}
\textbf{Model 2: }\quad\log\left(\frac{p}{1-p}\right) = \beta_0 + \beta_1 \textrm{PC}_1 +  \cdots+ \beta_{7} \textrm{PC}_{7} + \beta_{9} \textrm{PC}_{9}.
\end{equation}
The \textit{p}-value from the BAGofT is $0.21$.  So  this model cannot be rejected at the significance level of $0.05$. 

To compare the performance of the BAGofT with other GOF tests, we also consider the HL, CH, and GRP tests.  The results are shown in Table~\ref{adpv}. In contrast with the BAGofT, the other tests fail to reject the simpler model, reflecting their lack of power in this case.
\begin{table}[!ht]
\caption{\textit{P}-values for models from Equations~\eqref{AD1} and \eqref{AD2}. 
%$GRP_g$ denotes the GRP group test with a alternative model that contains $\textrm{PC}_1-\textrm{PC}_{20}$. $BAG^*$ denotes the BAGofT using only the covariates in the model to assess to apply the adaptive grouping.
}\label{adpv}
\centering
{\small
\begin{tabular}{ lllll }
Test & HL & CH  &GRP  &$\textrm{BAG}$    \\\hline
Model 1&0.42 & 0.26 & 0.17  &0.00 \\
Model 2& 0.17 & 0.23 & 0.23 &0.15  
\end{tabular}
}
\end{table}
% \begin{table}[!ht]
% \caption{\textit{P}-values for models from Equations~\eqref{AD1} and \eqref{AD2}. 
% %$GRP_g$ denotes the GRP group test with a alternative model that contains $\textrm{PC}_1-\textrm{PC}_{20}$. $BAG^*$ denotes the BAGofT using only the covariates in the model to assess to apply the adaptive grouping.
% }\label{adpv}
% \centering
% \begin{tabular}{ llllll }
% Test & HL & CH  &GRP &GRP-g &$\textrm{BAG}^*$    \\\hline
% Model 1&0.42 & 0.26 & 0.17 &0.13  & 0.00\\
% Model 2& 0.17 & 0.23 & 0.23&0.18 &   0.47
% \end{tabular}
% \end{table} 

\subsection{Testing classification procedures: Fashion MNIST data}\label{sub_MNIST}
%Rcodes149_mnist

\begin{figure}[!ht]
  \caption{An example of trouser and dress images from the Fashion MNIST data.}\label{trousers_dress}
  \centering
    \includegraphics[width=0.6\textwidth]{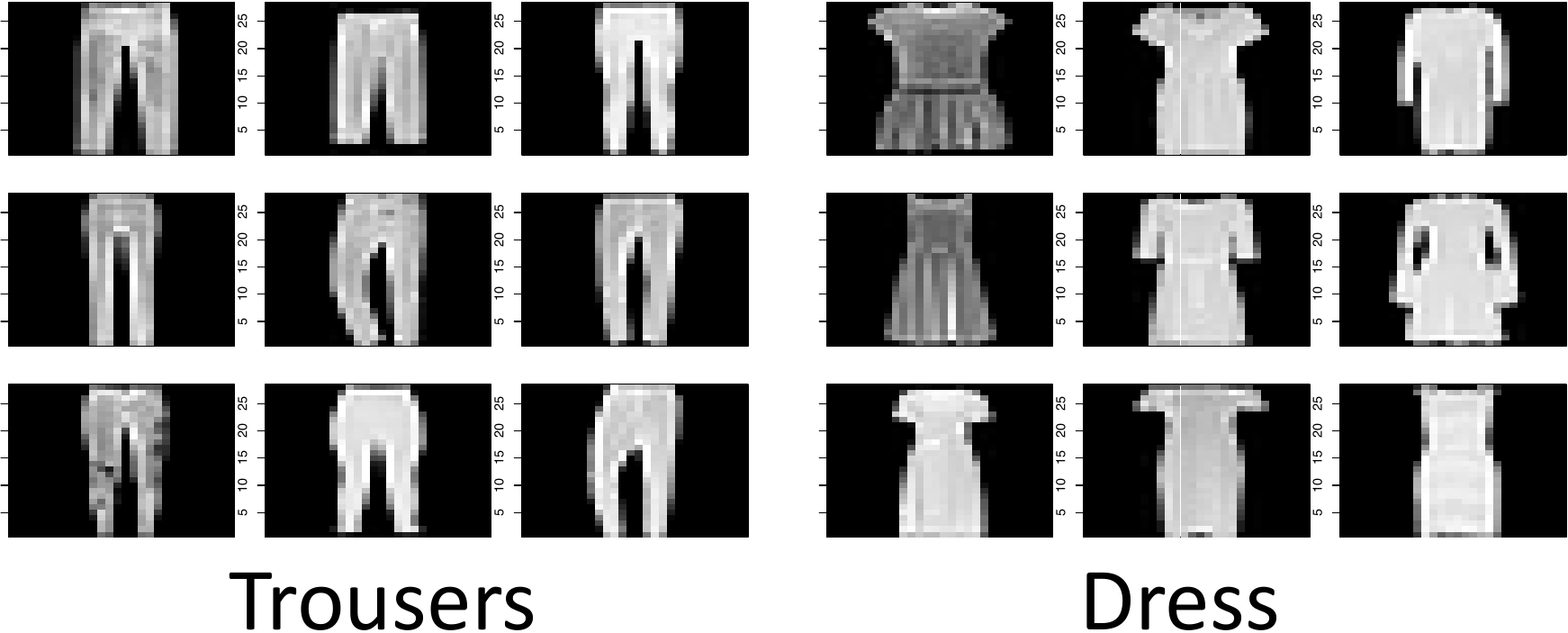}
\end{figure}

We consider the Fashion MNIST data \citep{xiao2017fashion}, which contain images of different clothes with a pixel size of $28\times 28$. We take the first 500 images of trousers and the first 500 images of blouses with a total sample size of 1000. An example snapshot of these images is shown in Figure~\ref{trousers_dress}. The PTA is a feed-forward neural network with one hidden layer and one neuron.  %The activation function we take is \textit{ReLu}~\citep{nair2010rectified}. 

\begin{figure}[!ht]
  \caption{Variable importance of the neural network fitted to the Fashion MNIST data. Covariates with higher variable importance are marked by brighter color. The neural network still has room for a major improvement with those highlighted covariates. }\label{trousers_dress_comp}
  \centering
    \includegraphics[width=0.3\textwidth]{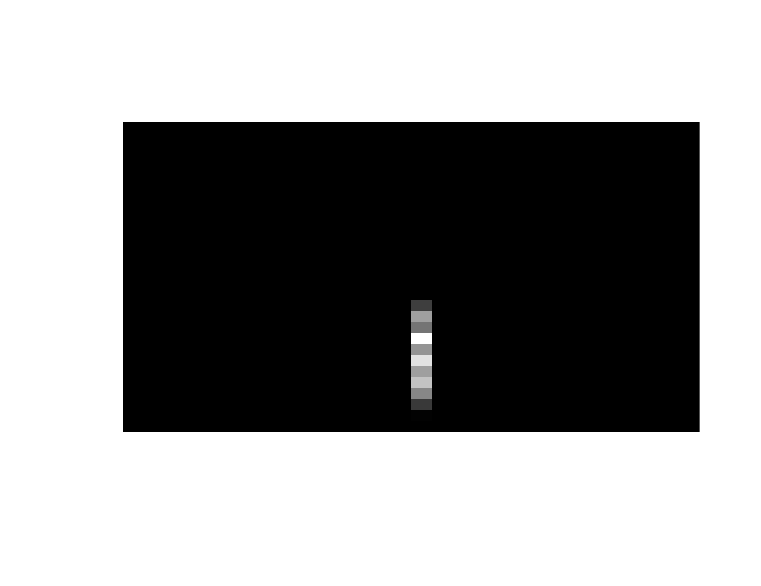}
\end{figure}

The BAGofT has a bootstrap \textit{p}-value 0 in each of the three splitting ratios. It indicates that the neural network fails to capture at least one major aspect from the data (\textbf{Pattern 4}).  
To interpret the testing results, we plot the (Random Forest) variable importance of the $28\times 28$ covariates from the BAGofT (with $90\%$ data for training) in Figure~\ref{trousers_dress_comp}.  
As is remarked in Section~\ref{gp}, the covariates with high variable importance are likely to be the major reason for the underfitting.
It can be interpreted from Figure~\ref{trousers_dress_comp} that the space between the two legs of the trousers is where the PTA underfits. This is indeed the major difference between the two kinds of clothes.  

\subsection{Testing classification procedures: COVID-19 CT scans}\label{sub_COVID}
% Rcodes150 ne5, ne7, XG
Coronavirus disease 2019 (COVID-19) has had a massive impact on the world. We consider the data in the study from \cite{he2020sample}, which is available at https://github.com/UCSD-AI4H/COVID-CT. The training and test sets contain a total of 339 positive cases and 289 negative cases.

Our study considers assessing classification procedures fitted on the 1000 features generated from the pre-trained deep learning model MobileNetV2 \citep{sandler2018mobilenetv2}. The images are resized into $224\times 224$ RGB pixels before entering MobileNetV2. The PTAs are two one-layer neural networks and two XGBoost classifiers. The two neural networks 
%(with the \textit{ReLu}  activation) 
consist of 1 and 7 neurons, respectively. The two XGboost classifiers consist of 10 and 500 base learners, respectively. The  details of the PTAs are included in the supplement.

The \textit{p}-values are summarized in Table~\ref{covid}. It can be seen that both the neural network with 1 neuron and XGBoost with 10 base learners are too restrictive to capture the nature of the data (\textbf{Pattern 4}). Both the neural network with 7 neurons and  XGBoost with 500 base learners belong to \textbf{Pattern 1}, and thus handle the data quite well.  We also calculate the prediction accuracies of the PTAs by taking 0.5 as the threshold and averaging the accuracies over 100 replications under the three splitting ratios (namely 90\%, 75\%, 50\%). The result shows that the models not rejected by the BAGofT have accuracies uniformly better than those that are rejected. Note that when assessed by prediction accuracies, the neural network with 1 neuron is only slightly worse than the one with 7 neurons. Nevertheless, the BAGofT is able to indicate that the difference in accuracies comes from a systematic defect of the 1-neuron network.
\begin{table}[!ht]
\caption{\textit{P}-values and prediction accuracies from classification procedures fitted on the COVID-19 data \citep{he2020sample}. NNET-1, NNET-7, XG-10, and XG-500 denote the neural network with 1 neuron, the neural network with 7 neurons, the XGBoost with 10 base learners, and the XGBoost with 500 base learners,  respectively.}\label{covid}
\centering
{\small
\begin{tabular}{ lllllll }
&\multicolumn{3}{c}{P-values}&\multicolumn{3}{c}{Accuracy}
\\\hline
Splitting ratio & $90\%$ & $75\%$  &$50\%$    & 90\% & 75\% & 50\% \\
\hline
NNET-1&0.03 & 0.00& 0.00 & 0.71 & 0.70 & 0.69 \\ 
NNET-7 & 0.62 & 1.00 & 0.32& 0.72 & 0.71 & 0.70 \\
XG-10 &0.00 &0.00 &0.00& 0.65 & 0.65 & 0.64\\
XG-500 &1.00& 0.25& 0.60& 0.72 & 0.72 & 0.70 \\ 
\end{tabular}
}
\end{table}

\section{Conclusion and Discussion}\label{Conclu}

We have developed a new methodology called the BAGofT to assess the GOF of classification learners. One major novelty is that, unlike the previous methodologies in the literature, it can assess general classification procedures, which is more challenging and has a more extensive application scope than testing parametric models.  We have shown both theoretically and experimentally that the BAGofT can effectively reveal different performance patterns of the PTA. 
%It allows researchers to explore the goodness-of-fit of classification models or procedures. 
Another novelty is the adaptive grouping, which can flexibly expose the MTA or PTA's weaknesses and make the developed tool highly powerful.
The adaptive grouping may also be used to interpret which covariates are possibly associated with the underfitting.  
In the context of assessing parametric models, numerical results have demonstrated the significant advantages of the BAGofT compared with some existing tests, including the popular Hosmer-Lemeshow test.

It is worth emphasizing that the BAGofT has a different usage compared with the assessment tools centered on the classification accuracy. Instead of directly measuring the prediction performance of an MTA/PTA, the BAGofT checks whether it has a detectable systematic issue that leads to slow or non-convergence for the observed data. %Two example scenarios of using the BAGofT are listed as follows.
In one application, the BAGofT can be used by scientists to justify the postulated parametric models and consequently interpret the results on the data-generating mechanism. 
In another application, data analysts may use the BAGofT to check for systemic defects and make critical business decisions on whether to put more effort on improving an existing MTA/PTA. For many medical and financial applications, it may be valuable to pursue even the smallest improvement of existing methods when we know that they are defective.  On the other hand, for other applications where the accuracy at a certain level is fully acceptable, there is no need to perform the BAGofT or other GOF test as long as the accuracy of the MTA/PTA is high enough.
% For high-tech companies that pursue even the smallest improvement of existing methods, the BAGofT can help them check systemic defects and reveal the potential for improvement.

One remaining challenge for the BAGofT is the identification of an overfitted MTA/PTA. When an MTA/PTA is substantially overfitted, the adaptive partition may fail to discover the deviation using the training set because the Pearson residuals may look clean. Nevertheless, in the case of severe overfitting, the chi-squared statistic calculated on the validation set may be able to capture the enlarged variance, and thus the BAGofT may still reject the MTA/PTA. An interesting future direction is to effectively identify large variances from an overfitted MTA/PTA. Another future direction is to extend the BAGofT to the classification problems with $d>2$ classes. A possible way is to define the statistic $T$ by $\sum^{K_n}_{k=1}\bm{R_k}^\T \bm{V_k}^{-1}\bm{R_k}$ where $\bm{R_k}$ is the sum of differences between the observed response vectors and estimated probabilities from the $k$th group, and $\bm{V_k}$ is the estimated covariance matrix for that group. It can be verified by the multivariate Berry-Esseen Theorem that $\bagoft = 1- P(\chi^2_{K_n\cdot(d - 1)} \leq \csum|\csum, K_n)$ has an asymptotic standard uniform distribution under $H_0$.  Nevertheless, a large $d$ brings in computational challenges for the adaptive partition. 
 
 The \textit{R} package
`BAGofT' and codes to reproduce the results in Sections~\ref{simstu} and \ref{realExp} are available at https://github.com/JZHANG4362/BAGofT.

%he performance of the test compared to , which is  Simulation studies suggest that our test could be a good companion to  Hosmer-Lemeshow $\hat C$ test since it is more powerful in some cases. The performance of our test for other binary regression 
%models such as Probit regression, complementary log-log regression and comparison to other goodness-of-fit statistics could be interesting topics in the future. 
\begin{appendices}

\section{Organization of the supplementary document}

This supplementary document is organized as follows. 
In Section~\ref{supp_proofs}, we prove all the technical results in the main paper. 
In Section~\ref{supp_identify}, we justify the proposed algorithm by showing that under some reasonable conditions, the sets generated from the $K$-quantiles of the fitted Pearson residuals satisfy Condition 7. In Section~\ref{sec_diss_Cond_parametricExperiment}, we discuss the required conditions for the properties of BAGofT in the specific numerical studies in Section~5.1 from the main text.
In Section~\ref{supp_qq}, we present the Q-Q plots under the null hypothesis in testing parametric models.
In Section~\ref{supp_graph}, we develop visualizations to illustrate the efficacy of BAGofT in generating adaptive partitions in comparison with the HL test.
In Section~\ref{supp_preselection}, we experimentally compare the BAGofT with and without covariates pre-selection.
In Section~\ref{supp_grp_comparison}, we numerically evaluate the BAGofT in assessing high dimensional parametric classification models and compare it with the GRP test, the state-of-the-art approach to measuring the GOF of high dimensional generalized linear models.
In Section~\ref{supp_lowdim}, we demonstrate the application of the BAGofT to assessing low-dimensional classification learning procedures.
In Section~\ref{supp_splits}, we investigate the variation of test statistics against the number of splittings.
Section~\ref{supp_varimport} demonstrates the variable importance of covariates and how they can be used to identify the source of underfitting.
Section~\ref{supp_covid} provides more experimental details on the COVID-19 CT scans data example.

\section{Proof of the main theorems} \label{supp_proofs}
Since the results for learning procedures are more general than those for parametric models in many aspects, we first give the proofs of Theorems~3 and~4, and then prove Theorems~1 and~2.

\subsection{Proof of Theorem~3 (Convergence of $\bagoft$ under $\textrm{H}_0$ for classification procedures)} %\ref{up1}}

%\begin{proof}
We need to prove that for each $u\in (0,1)$,
\begin{equation}\label{eq_goal}
    P(P(\chi^2_{K_n} \leq \csum|\csum, K_n) \leq u) \rightarrow u.
\end{equation}
as $n\rightarrow\infty$. 
Let \begin{align*}
\csumt &=\sum^{K_n}_{k=1}\left(\frac{\sum_{\{i:\  \bm{x}_{e,i} \in \hat\gp_{\data_{\ntrain},k}\} } \rvt_i}{\sqrt{\sum_{\{i:\  \bm{x}_{e,i} \in \hat\gp_{\data_{\ntrain},k}\} }{\sigt_i}^2}} \right)^2,\\
\end{align*}
where
\begin{align*}
\rvt_i &= y_{e,i} - \prob(\bm{x}_{e,i}), \\
 {\sigt_i}^2 &=  \prob(\bm{x}_{e,i})\left\{1 -  \prob(\bm{x}_{e,i})\right\}.
\end{align*}
We claim that it suffices to show that
\begin{align}
&P(P(\chi^2_{K_n} \leq \csumt|\csumt, K_n) \leq u) \rightarrow u, \forall\ u\in(0,1), \label{proof1_part1}\\
&|P(\chi^2_{K_n}\leq\csum\mid \csum, K_n) - P(\chi^2_{K_n}\leq\csumt\mid \csumt, K_n)|\limp 0,\label{proof1_part3}
\end{align}
% \begin{align}
% \underset{\mathfrak{z}\in \mathbb R^+}{\text{sup}} | P(\csumt < \mathfrak{z}\mid K_n) - P(\chi^2_{K_n}< \mathfrak{z}\mid K_n) | &\limp 0, \label{proof1_part1}\\
% |P(\chi^2_{K_n}<\csum\mid \csum, K_n) - P(\chi^2_{K_n}<\csumt\mid \csumt, K_n)|&\limp 0,\label{proof1_part3}
% \end{align}
as $n\rightarrow \infty$.
 The explanation is as follows. 
Let $E_\epsilon$ be the event of 
$$|P(\chi^2_{K_n}\leq\csum\mid \csum, K_n) - P(\chi^2_{K_n}\leq\csumt\mid \csumt, K_n)| < \epsilon,$$
where $0<\epsilon < 1$.
By \eqref{proof1_part3},  there exists $N_\epsilon$ such that for all $n>N_\epsilon$,
\begin{equation}\label{eq_Ep}
  P(E_\epsilon)>1 - \epsilon.  
\end{equation}
Let 
\begin{align*}
    p_1 &= P(\chi^2_{K_n} \leq \csum|\csum, K_n),\\
    p_2 &= P(\chi^2_{K_n} \leq \csumt|\csumt, K_n).
\end{align*}
For each $u\in (0,1)$, we take $\epsilon<\min(u,1-u)$. We then have
\begin{align*}
    \{p_1\leq u\}&\subseteq E_\epsilon^c \cup (\{p_2\leq u + \epsilon\}\cap E_\epsilon)\\
    &\subseteq  E_\epsilon^c \cup \{p_2\leq u + \epsilon\}.
\end{align*}
According to \eqref{eq_Ep}, when $n>N_\epsilon$,
\begin{align*}
    P(p_1\leq u)&\leq P(E_\epsilon^c) + P(p_2\leq u + \epsilon)\\
    &\leq P(p_2\leq u + \epsilon) + \epsilon.
\end{align*}
Similarly, it follows from
$$\{p_2\leq u -\epsilon\}\subseteq E_\epsilon^c\cup (\{p_1\leq u\}\cap E_\epsilon),$$
that for $n>N_\epsilon$,
$$ P(p_1\leq u)\geq P(p_2\leq u-\epsilon) - \epsilon.$$
% Then
% $$P(p_2 \leq u + \epsilon\mid E_\epsilon) \geq P(p_1 \leq u \mid E_\epsilon)\geq P(p_2 \leq u - \epsilon\mid E_\epsilon),$$
% almost surely.
% In addition with \eqref{eq_Ep},  we have when $n>N_\epsilon$, 
% \begin{align*}
%     P(p_1 \leq u) &= P(p_1 \leq u \mid E_\epsilon)\cdot P(E_\epsilon)
% + P(p_1 \leq u \mid E_\epsilon^c)\cdot P(E_\epsilon^c)\\
%     &\leq P(p_1 \leq u \mid E_\epsilon) + P(E_\epsilon^c)\\
%     &< P(p_2 \leq u + \epsilon\mid E_\epsilon)  + \epsilon\\
%     &= \frac{ P(p_2 \leq u + \epsilon) - P(p_2 \leq u + \epsilon\mid E_\epsilon^c)\cdot P(E_\epsilon^c)}{P(E_\epsilon)} + \epsilon\\
%     &< P(p_2 \leq u + \epsilon)/(1-\epsilon) + \epsilon,
% \end{align*}
% and 
% \begin{align*}
% P(p_1 \leq u) &\geq  P(p_1 \leq u \mid E_\epsilon)\cdot P(E_\epsilon)\\
% &\geq P(p_2 \leq u - \epsilon) - P(p_2 \leq u - \epsilon\mid E_\epsilon^c)\cdot P(E_\epsilon^c)\\
% &\geq  P(p_2 \leq u - \epsilon) -P(E_\epsilon^c)\\
% &> P(p_2 \leq u - \epsilon)-\epsilon.
% \end{align*}
The above inequalities, in conjunction with \eqref{proof1_part1}, imply the desired \eqref{eq_goal}.

%Since the probability of having at least one observation in each group goes to 1 as $n\rightarrow \infty$, we may consider this setting with at least one observation in each group without losing generality. Then we can define random variable

To prove (\ref{proof1_part1}), we first show that it suffices to prove that %,\ K_n\in [2,\ntest/\underline m_n]
\begin{equation}\label{eq_sup_csumt}
    \underset{x^*\in \mathbb R^+}{\text{sup}} | P(\csumt \leq x^*\mid K_n) - P(\chi^2_{K_n}\leq x^*\mid K_n)| \limp 0,
\end{equation}
as $n\rightarrow\infty$.
Let  $F_{K_n}^{-1}(x)$ for $0<x<1$ be the inverse CDF of $\chi^2_{K_n}$  conditional on $K_n$ with
$$P(\chi^2_{K_n} \leq F_{K_n}^{-1}(x)\mid K_n)= x,\ \forall x\in (0,1),$$
and 
$$F_{K_n}^{-1}(P(\chi^2_{K_n} \leq x'\mid K_n)) = x',\ \forall x'\in \mathbb R^+,$$
almost surely.
For each $u\in (0,1)$, we have
\begin{align*}
    P(P(\chi^2_{K_n} \leq \csumt\mid\csumt, K_n) \leq u)
    &=P(F_{K_n}^{-1}(P(\chi^2_{K_n} \leq \csumt\mid\csumt, K_n)) \leq F_{K_n}^{-1}(u))\\
    &=E(P(\csumt \leq F_{K_n}^{-1}(u)\mid K_n) ).
\end{align*}
By \eqref{eq_sup_csumt} and the dominated convergence theorem applied to $$P(\csumt \leq F_{K_n}^{-1}(u)\mid K_n)-P(\chi^2_{K_n} \leq F_{K_n}^{-1}(u)\mid K_n),$$ 
we have
$$
E(P(\csumt \leq F_{K_n}^{-1}(u)\mid K_n) )\rightarrow E(P(\chi^2_{K_n} \leq F_{K_n}^{-1}(u)\mid K_n)) = u,
$$
as $n\rightarrow\infty,$  Thus, we have shown that it suffices to prove \eqref{eq_sup_csumt} in order to obtain \eqref{proof1_part1}.

%So next we prove \eqref{eq_sup_csumt}.
Define random vectors 
$\V_i = (\V_{i1},\dots,\V_{i K_n})^T$ $(i=1,\dots,\ntest)$,
where 
$$
\V_{ik}=
\begin{cases}
	\rvt_i\big/\sqrt{\sum_{\{j:\ \bm{x}_{e,j} \in \hat\gp_{\data_{\ntrain},k}\}  }{\sigt_j}^2} &\text{ if } i\in \hat\gp_{\data_{\ntrain},k},\\
	0&\text{ otherwise},
\end{cases}
$$
and $\V = \sum_{i=1}^{\ntest} \V_i$.
Let $D_{\ntrain}$ denote the training set data and $\data_{x_e}$ denote the covariate part of the evaluation data $\data_{\ntest}$.
Since  conditioning on  $\data_{x_e}$ and $\data_{\ntrain}$, $y_{e,i}$ ($i = 1,\dots, \ntest$) are independent and  the partition  $\{\hat\gp_{\data_{\ntrain},k}\}^{K_n}_{k=1}$ is fixed, $\V_{i}$ ($i=1,\dots,\ntest$) are independent. 
%Define random variables
%\begin{equation}\label{eq_Bk}
%\bd_k = 
%\frac{\sum_{\{i:\ \bm{x}_{e,i} \in \hat\gp_{\data_{\ntrain},k}\} } \rvt_i}{\sqrt{\sum_{\{i:\ \bm{x}_{e,i} \in \hat\gp_{\data_{\ntrain},k}\}  }{\sigt_i}^2}} ,\ k = 1,\dots,K_n,
%\end{equation}
% It can be seen that the covariance matrix of $\V$ conditiontional on $\data_{x_e}$ and $\data_{\ntrain}$ is  the identity matrix with dimension $K_n\times K_n$. 
 We also have $ E\left(\V_i|\data_{\ntrain},\data_{x_e}\right) = \bm{0}$, $Cov\left(\V|D_{\ntrain},D_{x_e}\right)  =  \bm{I}$, and 
\begin{align*}
E\biggl(\biggl|\sum_{k=1}^{K_n}\V_{ik}^2\biggr|^{3/2}\mid D_{\ntrain},D_{x_e}\biggr)
=&\ E\biggl( |\rvt_i|^3\big/\bigg(\sum_{\{j:\ \bm{x}_{e,j} \in \hat\gp_{\data_{\ntrain},k_i}\}  }{\sigt_j}^2\bigg)^{3/2}\mid D_{\ntrain},D_{x_e}\biggr)\\
= &\ \frac{\prob(\bm{x}_{e,i})\left\{ 1 -  \prob(\bm{x}_{e,i})\right\}\left[\{1-\prob(\bm{x}_{e,i})\}^2 + \prob(\bm{x}_{e,i})^2\right]}{(\sum_{\{j:\ \bm{x}_{e,j} \in \hat\gp_{\data_{\ntrain},k_i}\}  }\prob(\bm{x}_{e,j})\left\{1 -  \prob(\bm{x}_{e,j})\right\})^{3/2}}\\
\leq&\ \frac{0.125}{(\hat n_{2,k}\cst_1(1-\cst_1))^{3/2}}\\
\leq&\ \frac{0.125}{(\underline m_n\cst_1(1-\cst_1))^{3/2}},
\end{align*}
where $\bm{I}$ is the identity matrix, $k_i$ is the group index of the $i$th observation from $\data_{\ntest}$, and $\hat n_{2,k} = \sum_{i = 1}^{\ntest} I{\{\bm{x}_{e,i} \in \hat\gp_{\data_{\ntrain},k}\}}$ is the number of observations in the $k$th group.
Let $Z^{(K_n)}$ denote a vector of $K_n$ i.i.d.\ standard normal random variables and $\convex$ denote the collection of all convex sets of $\mathbb R^{K_n}$.
Combining the above results, the fact that $K_n\leq \ntest/{\underline m_n}$, and Theorem~1.1 of \cite{bentkus2005lyapunov}, we have that there exists a  positive constant $C_0$, such that
\begin{align}
\underset{A\in \convex}{\text{sup}}\left| \left. P\left(\V\in A\right|D_{\ntrain},D_{x_e} \right) - P(Z^{(K_n)}\in A\mid K_n) \right| \nonumber
\leq &\ C_0 \cdot K_n^{1/4}\cdot(\ntest/\underline m_n^{3/2})\\
\leq &\ C_0\cdot \ntest^{5/4}\cdot \underline m_n^{-7/4}\label{eq_BKnbound}, 
\end{align}
almost surely, which by Condition~1 goes to 0 as $n\rightarrow\infty$. Since $\convex$ contains all balls in $\mathbb R^{K_n}$ centered at the origin, we have
\begin{align*}
&\underset{x^*\in \mathbb R^+}{\text{sup}} | P(\csumt \leq x^*\mid K_n) - P(\chi^2_{K_n}\leq x^*\mid K_n)| \\
   \leq &\underset{A\in \convex}{\text{sup}}\left| \left. P\left(\V\in A\right|K_n\right) - P(Z^{(K_n)}\in A\mid K_n) \right|\\
   =& \underset{A\in \convex}{\text{sup}}\left| E\biggl(\left. P\left(\V\in A\right|D_{\ntrain},D_{x_e}\right) - P(Z^{(K_n)}\in A\mid K_n) \mid K_n\biggr)\right|\\
\leq &\ E\biggl(\underset{A\in \convex}{\text{sup}}\left| \left. P\left(\V\in A\right|D_{\ntrain},D_{x_e} \right) - P(Z^{(K_n)}\in A\mid K_n) \right|\mid K_n\biggr)\\
\leq &\ C_0\cdot \ntest^{5/4}\cdot \underline m_n^{-7/4},	
\end{align*}
almost surely. 
Therefore, we obtain  \eqref{eq_sup_csumt} and thus  \eqref{proof1_part1}.

Next, we prove (\ref{proof1_part3}).
First, let $$f_{K_n}(x)=\frac{1}{2^{K_n/2}\Gamma(K_n/2)}x^{K_n/2-1}\exp{(-x/2)},$$
with $x\in \mathbb R^+$
be the density function  of $\chi^2_{k}$ conditional on $k=K_n$. 
We have
$$|P(\chi^2_{K_n}\leq x_1\mid K_n) - P(\chi^2_{K_n}\leq x_2\mid K_n)| = \biggl|\int^{x_1}_{x_2}f_{K_n}(x)dx\biggr|\leq |x_1 - x_2| \cdot \sup_{x\in \mathbb R^+}f_{K_n}(x),$$
almost surely for all $x_1, x_2\in \mathbb R^+$. Recall that we require $K_n\geq 2$.
It can be verified 
%by taking the derivative of $f_{K_n}(x)$ 
that for $K_n=2$, $$\sup_{x\in \mathbb R^+}f_{K_n}(x) = 0.5,$$ and for $K_n>2$,
$$\sup_{x\in \mathbb R^+}f_{K_n}(x) = \frac{1}{(2\Gamma(K_n/2))}\cdot \biggl(\frac{K_n-2}{2}\biggr)^{K_n/2-1}\exp(-(K_n-2)/2),$$
almost surely.
It can be shown by applying the  Stirling's formula to $\Gamma(K_n/2)$ that the above supremum converges to 0 as $K_n\rightarrow\infty$. 
%%%%%%%%%%%%%%%%%%%%%%%%%%%%%%%%%%%%%
% Stirling's formula for gamma function: gamma(x)~Cx^{x - 1/2}exp(x)
%%%%%%%%%%%%%%%%%%%%%%%%%%%%%%%%%%%%%
So there exists a constant $C>0$, such that 
$$|P(\chi^2_{K_n}\leq x_1\mid K_n) - P(\chi^2_{K_n}\leq x_2\mid K_n)| \leq C|x_1 - x_2|,$$
almost surely 
for all $x_1, x_2\in \mathbb R^+$.
Then, 
$$|P(\chi^2_{K_n}\leq \csum\mid\csum, K_n) - P(\chi^2_{K_n}\leq \csumt\mid\csumt, K_n)| \leq C|\csum-\csumt|,$$
almost surely.
So to prove (\ref{proof1_part3}), it suffices to show 
\begin{equation}
  |\csum-\csumt|\limp 0\label{proof1_part2}  
\end{equation}
 as $n\rightarrow\infty$.

By our definition of $\csum$, 
$$\csum =\sum_{k=1}^{K_n} \biggl(\frac{1}{\sig_k}\sum_{\{\bm{x}_{e,i} \in \hat\gp_{\data_{\ntrain},k}\} } \rv_i \biggr)^2,$$
where $\rv_i =  y_{e,i} - \hat\prob_{\data_{\ntrain}}(\bm{x}_{e,i})$, $\sig_k^2=\sum_{\{i:\ \bm{x}_{e,i} \in \hat\gp_{\data_{\ntrain},k}\} } \sig_i^2$, and $\sig_i^2 =  \hat\prob_{\data_{\ntrain}}(\bm{x}_{e,i}) \left\{1 -  \hat\prob_{\data_{\ntrain}}(\bm{x}_{e,i})\right\}$.
We have the decomposition:
\begin{equation*}
\begin{split}\frac{1}{\sig_k}\sum_{\{\bm{x}_{e,i} \in \hat\gp_{\data_{\ntrain},k}\} } \rv_i 
&=  \frac{1}{\sig_k}\sum_{\{\bm{x}_{e,i} \in \hat\gp_{\data_{\ntrain},k}\} } \rvt_i + \frac{1}{\sig_k}\sum_{\{i:\ \bm{x}_{e,i} \in \hat\gp_{\data_{\ntrain},k}\} } \left\{\prob(\bm{x}_{e,i}) - \hat \prob_{\data_{\ntrain}}(\bm{x}_{e,i})\right\}\\
&=T_{1,k} + T_{2,k}+ T_{3,k},
\end{split}
\end{equation*}
where we define
\begin{itemize}
\item $T_{1,k} = \frac{1}{\sigt_k}\sum_{\{i:\ \bm{x}_{e,i} \in \hat\gp_{\data_{\ntrain},k}\} }\rvt_i$,
\item $T_{2,k} = \frac{\sigt_k - \sig_k}{\sig_k}\cdot \frac{1}{\sigt_k}\sum_{\{i:\ \bm{x}_{e,i} \in \hat\gp_{\data_{\ntrain},k}\} }\rvt_i$,
\item $T_{3,k} =  \frac{1}{\sig_k}\sum_{\{i:\ \bm{x}_{e,i} \in \hat\gp_{\data_{\ntrain},k}\} } \left\{\prob(\bm{x}_{e,i}) - \hat \prob_{\data_{\ntrain}}(\bm{x}_{e,i})\right\}$,
\item ${\sigt_k}^2 =\sum_{\{i:\ \bm{x}_{e,i} \in \hat\gp_{\data_{\ntrain},k}\} } {\sigt_i}^2$.
\end{itemize}
Then, we have \[
\csum =\sum_{k=1}^{K_n}  T_{1,k}^2 + \sum_{k=1}^{K_n}  T_{2,k}^2 + \sum_{k=1}^{K_n}  T_{3,k}^2 +  2\sum_{k=1}^{K_n} T_{1,k}T_{2,k} + 2\sum_{k=1}^{K_n} T_{1,k}T_{3,k} + 2\sum_{k=1}^{K_n} T_{2,k}T_{3,k}.
\]
By the Cauchy-Schwarz inequality, 
\[\biggl|\sum_{k=1}^{K_n} T_{1,k}T_{2,k}\biggr|\leq \sqrt{\sum_{k=1}^{K_n} T_{1,k}^2\sum_{k=1}^{K_n}T_{2,k}^2},\]
almost surely.
Similar inequalities hold for $\sum_{k=1}^{K_n} T_{1,k}T_{3,k}$ and $\sum_{k=1}^{K_n} T_{2,k}T_{3,k}$. 
From the above inequalities and $\sum_{k=1}^{K_n} T_{1,k}^2= \csumt$, we have
$$|\csum - \csumt| \leq \biggl|\sum_{k=1}^{K_n}  T_{2,k}^2\biggr| + \biggl|\sum_{k=1}^{K_n}  T_{3,k}^2\biggr| +  2\sqrt{\csumt\sum_{k=1}^{K_n}T_{2,k}^2} + 2\sqrt{\csumt\sum_{k=1}^{K_n}T_{3,k}^2} + 2\sqrt{\sum_{k=1}^{K_n} T_{2,k}^2\sum_{k=1}^{K_n}T_{3,k}^2}$$
almost surely. By
$$E(|\csumt/K_n|) = E(E(\csumt/K_n \mid \data_{\ntrain},\data_{x_e})) = 1$$
and the Markov's inequality, we have 
$$\csumt/K_n = O_p(1).$$
Since $K_n$ is lower bounded away from $0$,
to prove \eqref{proof1_part2},
it remains to show that
\begin{align}
    K_n\sum_{k=1}^{K_n}  T_{2,k}^2\rightarrow_p 0,\label{eq_T2}\\
    K_n\sum_{k=1}^{K_n}  T_{3,k}^2\rightarrow_p 0\label{eq_T3},
\end{align}
as $n\rightarrow\infty$.

We first prove \eqref{eq_T2}. Recall that $\hat n_{2,k} = \sum_{i = 1}^{\ntest} I{\{\bm{x}_{e,i} \in \hat\gp_{\data_{\ntrain},k}\}}$.
By our definition,
$$ K_n\sum_{k=1}^{K_n} T_{2,k}^2 =   K_n\sum_{k=1}^{K_n} \biggl\{(\sigt_k - \sig_k)\cdot (\sqrt{\hat n_{2,k}}/\sig_k)\cdot \frac{1}{\sqrt{\hat n_{2,k}}\sigt_k}\sum_{\{i:\ \bm{x}_{e,i} \in \hat\gp_{\data_{\ntrain},k}\} }\rvt_i\biggr\}^2.$$
Since  $|\rvt_i|$'s ($i = 1,\dots, \ntest$) are uniformly bounded above by one and by Condition~2, $1/\sigt_k\leq 1/\sqrt{\hat n_{2,k}\cst_1(1-\cst_1)}$ almost surely, it suffices to show that
\begin{equation}\label{eqsqareroot}
K_n  \sum_{k=1}^{K_n}\left(\sig_k - \sigt_k\right)^2\rightarrow_p 0,  
\end{equation}
and there exists $\overline C>0$ such that
\begin{align}
	P(\max_{k = 1,\dots,K_n}(\hat n_{2,k}/\sig_k^2)>\overline C) \limp 0\label{part2},
\end{align}
as $n\rightarrow \infty$.

First, we consider \eqref{eqsqareroot}, which can be written as
$$K_n\sum_{k=1}^{K_n}\hat n_{2,k}\cdot\left(\sig_k/\sqrt{\hat n_{2,k}} - {\sigt_k}/\sqrt{\hat n_{2,k}}\right)^2 = K_n\sum_{k=1}^{K_n}\hat n_{2,k}\cdot\left( \frac{\sig_k^2/\hat n_{2,k} - {\sigt_k}^2/\hat n_{2,k}}{\sig_k/\sqrt{\hat n_{2,k}} + \sigt_k/\sqrt{\hat n_{2,k}}}  \right)^2.$$
By Condition~2, $\sigt_k/\sqrt{\hat n_{2,k}}\geq \sqrt{\cst_1(1-\cst_1)}>0$. Therefore, it suffices to show that 
$$K_n\sum_{k=1}^{K_n}\hat n_{2,k}\cdot((\sig_k^2 - {\sigt_k}^2)/\hat n_{2,k})^2\limp 0$$ as $n\rightarrow\infty$. 
% By Condition~1 and $K_n\leq n_2/n_{min}$, it suffices to show 
% $(\sqrt{h_1} - \sqrt{h_2})/n_2^{1/6}\rightarrow_p 0$
% in order to prove \eqref{eqsqareroot}.
% We can write $\sqrt{h_1}-\sqrt{h_2} = (h_1-h_2) / (\sqrt{h_1}+\sqrt{h_2})$. The denominator is nonnegative. 
% By Condition~2, $\sqrt{h_2}$  is lower bounded by a constant. So we  can focus on the convergence rate of  $h_1-h_2$.
%By our definition, $\sig_i^2 =  \hat\prob_{\data_{\ntrain}}(\bm{x}_{e,i}) \left\{1 -  \hat\prob_{\data_{\ntrain}}(\bm{x}_{e,i})\right\}$, $ \sigt_i^2 =  \prob(\bm{x}_{e,i})\left\{1 -  \prob(\bm{x}_{e,i})\right\} $.
Consider $f(z) =z(1-z)$ with $z \in (0,1)$. By applying the Lagrange mean value theorem  on the function $f(z_1) - f(z_2)$  with $z_1 = \hat \prob_{\data_{\ntrain}}(\bm{x}_{e,i})$ and
$z_2 =\prob(\bm{x}_{e,i})$, we have 
\begin{equation}\label{con_sig}
    |\sig_i^2 - {\sigt_i}^2|\leq|\hat\prob_{\data_{\ntrain}}(\bm{x}_{e,i}) - \prob(\bm{x}_{e,i})|,
\end{equation}
almost surely.
%So $\xi_1$ is between 0 and 1, we have $|1 - 2\xi|\leq 1$, $|f(z_1) - f(z_2)|\leq |z_1 - z_2|$.
So
\begin{align}
    K_n\sum_{k=1}^{K_n}\hat n_{2,k}\cdot((\sig_k^2 - {\sigt_k}^2)/\hat n_{2,k})^2 &\leq K_n\sum_{k=1}^{K_n}\hat n_{2,k}\cdot\biggl(\sum_{\{i:\ \bm{x}_{e,i} \in \hat\gp_{\data_{\ntrain},k}\} }|\sig_i^2 - {\sigt_i}^2|/\hat n_{2,k}\biggr)^2 \nonumber \\
     &\leq K_n^2\cdot \ntest\cdot ( \underset{\forall \bm{x}\in \Support}{\text{sup}}|\hat \prob_{\data_{\ntrain}}(\bm{x})- \prob(\bm{x})|)^2\label{Kpi},
\end{align}
almost surely.
%Then consider \eqref{eqsqareroot}.
%Let $g(h) = \sqrt{h}$, for $h \in (0,1)$. By Lagrange mean value theorem, 
%$g(h_1) - g(h_2) = \frac{1}{2\sqrt{\xi_2}}(h_1 - h_2)$ for a $\xi_2$ between 
%$h_1$ and $h_2$.  Let $h_1 = \frac{1}{\hat n_{2,k}}\sum_{\{i:\ \bm{x}_{e,i} \in \hat\gp_{\data_{\ntrain},k}\} }\sig_i^2$ and
%$h_2 = \frac{1}{\hat n_{2,k}}\sum_{\{i:\ \bm{x}_{e,i} \in \hat\gp_{\data_{\ntrain},k}\} }\sigt_i^2$.
%So  when $n_{a}\geq N_{a}$, by Condition 2, for $M_2 = 2\sqrt{\cst_1(1-\cst_1)}M_1$,
%\begin{equation}\label{numRa}
%\begin{split}
%&P\left(\frac{1}{\rate(\ntrain)} \cdot   \left| \sqrt{\frac{1}{\hat n_{2,k}}\sum_{\{i:\ \bm{x}_{e,i} \in \hat\gp_{\data_{\ntrain},k}\} }\sig_i^2} - \sqrt{\frac{1}{\hat n_{2,k}}\sum_{\{i:\ \bm{x}_{e,i} \in \hat\gp_{\data_{\ntrain},k}\} }\sigt_i^2} \right| >M_2  \right) \\
%\leq &P\left(\frac{1}{\rate(\ntrain)} \cdot \frac{1}{2\sqrt{\cst_1(1-\cst_1)}} \frac{1}{\hat n_{2,k}}\sum_{\{i:\ \bm{x}_{e,i} \in \hat\gp_{\data_{\ntrain},k}\} } {|\sig_i^2- \sigt_i^2|} >M_2  \right) \\
%= &P\left(\frac{1}{\rate(\ntrain)} \cdot \frac{1}{\hat n_{2,k}}\sum_{\{i:\ \bm{x}_{e,i} \in \hat\gp_{\data_{\ntrain},k}\} } {|\sig_i^2- \sigt_i^2|}>M_1\right)\\
%\leq &\epsilon.
%\end{split}
%\end{equation}
Since $K_n\leq \ntest/\underline m_n$ almost surely,  \eqref{Kpi} is upper bounded by
\begin{align*}
    &\frac{1}{\rate_{\ntrain}^2}\cdot \rate_{\ntrain}^2\cdot \biggl( \underset{\forall \bm{x}\in \Support}{\text{sup}}|\hat \prob_{\data_{\ntrain}}(\bm{x})- \prob(\bm{x})|\biggr)^2 \ntest^3/\underline m_n^2\\
    &=\frac{1}{\rate_{\ntrain}^2}\cdot \biggl( \underset{\forall \bm{x}\in \Support}{\text{sup}}|\hat \prob_{\data_{\ntrain}}(\bm{x})- \prob(\bm{x})|\biggr)^2 (\ntest^{2/3}/\underline m_n)^2\cdot (\ntest^{5/6}\cdot \rate_{\ntrain})^2,
\end{align*}
almost surely.
Recall from  Condition~6 that
\begin{equation}\label{eq_cond6}
\underset{\forall \bm{x}\in \Support}{\text{sup}}|\hat \prob_{\data_{\ntrain}}(\bm{x})- \prob(\bm{x})| = O_p(\rate_{\ntrain})\ \text{as }n\rightarrow\infty.
\end{equation}
According to Condition~1, and  the requirement from Theorem~3, $\ntest^{2/3}/\underline m_n\rightarrow 0$ and $\ntest^{5/6} \rate_{\ntrain}\rightarrow 0$ as $n\rightarrow\infty$. Thus, we obtain \eqref{eqsqareroot}.

Next, we show \eqref{part2}. 
% First we have 
% \begin{equation}\label{eq_cstineq}
%     \sig_k - \sigt_k = (\sig_k^2 - \sigt_k^2)/(\sig_k + \sigt_k)\leq(\sig_k^2 - \sigt_k^2)/\sqrt{\cst_1(1-\cst_1)}.
% \end{equation}
By \eqref{con_sig} and Condition~2, when $\underset{\forall \bm{x}\in \Support}{\text{sup}}|\hat \prob_{\data_{\ntrain}}(\bm{x})- \prob(\bm{x})| < \cst_1(1-\cst_1)$,  we have
\begin{align*}
    \hat n_{2,k}/\sig_k^2&\leq\hat n_{2,k}/((\sigt_k)^2 -  \hat n_{2,k}\cdot \underset{\forall \bm{x}\in \Support}{\text{sup}}|\hat \prob_{\data_{\ntrain}}(\bm{x})- \prob(\bm{x})|)\\
    &\leq 1/\bigl(\cst_1(1-\cst_1) - \underset{\forall \bm{x}\in \Support}{\text{sup}}|\hat \prob_{\data_{\ntrain}}(\bm{x})- \prob(\bm{x})|\bigr),
\end{align*}
almost surely.
Then, by the uniform convergence of $\hat \prob_{\data_{\ntrain}}(\bm{x})$ to $\prob(\bm{x})$ from Condition~6, we obtain \eqref{part2}.

% follows from the convergence of $\sig_i^2$ to $\sigt_i^2$ from \eqref{con_sig} and the fact that $\sigt_i^2$ is lower bounded from 0 and upper bounded by $1/2$ as implied by Condition~2. 

It remains to show \eqref{eq_T3}. We have
\begin{equation*}
\begin{split}
K_n\sum_{k=1}^{K_n}  \csum_{3,k}^2 &= K_n\sum_{k=1}^{K_n}  \frac{1}{\sig_k^2} \biggl(\sum_{\{i:\ \bm{x}_{e,i} \in \hat\gp_{\data_{\ntrain},k}\} } \left\{\prob(\bm{x}_{e,i}) - \hat \prob_{\data_{\ntrain}}(\bm{x}_{e,i})\right\} \biggr)^2\\
&= K_n\sum_{k=1}^{K_n} \frac{\hat n_{2,k}^2}{\sig_k^2  }  \cdot \rate_{\ntrain}^2  \cdot \biggl(\frac{1}{\rate_{\ntrain}} \cdot \frac{1}{\hat n_{2,k}}\sum_{\{i:\ \bm{x}_{e,i} \in \hat\gp_{\data_{\ntrain},k}\} } \left\{\prob(\bm{x}_{e,i}) - \hat \prob_{\data_{\ntrain}}(\bm{x}_{e,i})\right\} \biggr)^2 \\
&\leq  K_n^2  \cdot \ntest \cdot \rate_{\ntrain}^2  \cdot \biggl( \frac{1}{\rate_{\ntrain}}\cdot  \underset{\forall \bm{x}\in \Support}{\text{sup}} |\hat \prob_{\data_{\ntrain}}(\bm{x})- \prob(\bm{x})|\biggr)^2 \cdot \sum_{k=1}^{K_n} \frac{\hat n_{2,k}}{\sig_k^2  }\big/K_n,
\end{split}
\end{equation*}
almost surely.
By \eqref{part2}, $\sum_{k=1}^{K_n} \frac{\hat n_{2,k}}{\sig_k^2  }\big/K_n = O_p(1)$. 
By Condition~6,
$$\frac{1}{\rate_{\ntrain}}\cdot  \underset{\forall \bm{x}\in \Support}{\text{sup}} |\hat \prob_{\data_{\ntrain}}(\bm{x})- \prob(\bm{x})| = O_p(1).$$
We also have 
$$ K_n^2 \cdot\ntest\cdot \rate_{\ntrain}^2 \leq \ntest^3\cdot \rate_{\ntrain}^2/\underline m_n^2 = (\ntest^{5/6}\rate_{\ntrain})^2 \cdot (\ntest^{2/3}/\underline m_n)^2,$$
almost surely. According to Condition~1 and the requirement from Theorem~3, 
the right-hand side of the above inequality goes to 0 as $n\rightarrow\infty$. 

So we have obtained both \eqref{eq_T2} and \eqref{eq_T3}, and thus we have  \eqref{proof1_part2}. 
% Since $K_n/n_2^{1/3}\rightarrow 0$, $\hat n_{2,k}/\sig_k^2 = O_p(1)$, $\underset{\forall \bm{x}\in \Support}{\text{sup}} |\hat \prob_{\data_{\ntrain}}(\bm{x})- \prob(\bm{x})|/\rate_{\ntrain} = O_p(1)$,  $\ntest^{2/3} \rate_{\ntrain}\rightarrow 0$  as $n\rightarrow \infty$, we obtain $\sum_{k=1}^{K_n}  \csum_{3,k}^2\rightarrow_p 0 $.
We have shown both \eqref{proof1_part1} and \eqref{proof1_part3},  and they indicate that $$\bagoft=1 - P(\chi^2_{K_n} \leq \csum|\csum, K_n)\rightarrow_d U,$$
where $U$ denotes the standard uniform distribution. This completes the proof.

\subsection{Proof of Theorem~4 (Consistency of $\bagoft$ under $\textrm{H}_1$ for classification procedures)}    \label{subsec_proof_thm4}
% Recall that 
% we calculate the  BAGofT statistic by 
% the \textit{p}-value of $\csum$
% with respect to a chi-squared distribution with degrees of freedom $K_n$  where 
% \[\csum  =  \sum^{K_n}_{k=1}\left(\frac{\sum_{ \{i:\ \bm{x}_{e,i} \in \hat\gp_{\data_{\ntrain},k}\}  } \left\{ y_{e,i} - \hat\prob_{\data_{\ntrain}}(\bm{x}_{e,i}) \right\}  }{\sqrt{\sum_{\{i:\ \bm{x}_{e,i} \in \hat\gp_{\data_{\ntrain},k}\} }\left[\hat\prob_{\data_{\ntrain}}(\bm{x}_{e,i})\left\{1 -  \hat\prob_{\data_{\ntrain}}(\bm{x}_{e,i})\right\}  \right]  }}\right)^2.\]
% Since a chi-squared random variable divided by  its degrees of freedom is bounded in probability and our \textit{p}-value statistic is defined by $F_{\chi^2}(\csum, K_n)$,

We need to prove that under $H_1$,
$$
    P(\chi^2_{K_n} \leq \csum\mid\csum, K_n) = P(\chi^2_{K_n}/K_n \leq \csum/K_n|\csum, K_n)\limp 1,
$$
as $n\rightarrow\infty$.
% When $K_n\limp\infty$ as $n\rightarrow\infty$, by the law of large numbers, for each $\epsilon>0$, there exist positive constants $N_\epsilon$ and $M$, if $n\geq N_\epsilon$,
% $$P(\chi^2_{K_n}/K_n \leq M\mid K_n) > 1 - \epsilon.$$ When $K_n$ is almost surely upper bounded, $\chi^2_{K_n}/K_n$ is also bounded in probability. So
First, we will show that it suffices to 
prove that 
\begin{equation}\label{eq_th4eq}
    \csum/K_n\limp\infty,\ \text{as}\ n\rightarrow\infty.
\end{equation}
According to 
$$E(\chi^2_{k}/k) = 1$$
and the Markov's inequality, for each $\epsilon'>0$, we have
$$P(|\chi^2_{k}/k| >\epsilon') < 1/\epsilon'$$
 uniformly for all $k \in \mathbb N$.
Thus, for each $0<\epsilon<1$,
there exists a positive constant $M_\epsilon$ such that
$$P(\chi^2_{K_n}/K_n \leq M_\epsilon\mid K_n) > 1 - \epsilon,$$
almost surely.
Also, $\csum/K_n\geq M_\epsilon$ implies that
\begin{align}
    P(\chi^2_{K_n}/K_n \leq \csum/K_n\mid \csum, K_n)
    &\geq P(\chi^2_{K_n}/K_n \leq M_\epsilon\mid K_n), \nonumber%\label{eq_TK2}
\end{align}
almost surely.
Therefore, we have
\begin{align*}
    P(P(\chi^2_{K_n} \leq \csum\mid \csum, K_n) > 1-\epsilon)
    &\geq P(\{P(\chi^2_{K_n} \leq \csum\mid \csum, K_n) > 1-\epsilon\}\cap \{\csum/K_n\geq M_\epsilon\})\\
    &\geq P(\{P(\chi^2_{K_n}/K_n \leq M_\epsilon\mid K_n) > 1 - \epsilon\}\cap \{\csum/K_n\geq M_\epsilon\})\\
    &\geq P(P(\chi^2_{K_n}/K_n \leq M_\epsilon\mid K_n) > 1 - \epsilon)
    + P(\csum/K_n\geq M_\epsilon) - 1\\
    &= P(\csum/K_n\geq M_\epsilon).
\end{align*}
Thus, it suffices to show \eqref{eq_th4eq}. 

We rewrite $T$ as
\begin{equation}\label{eq_Tdecomp}
    \csum =\sum^{K_n}_{k=1}(\bd_k\cdot \csum_{a,1,k} + \csum_{a,2,k})^2,
\end{equation}
where 
\begin{equation*}
\begin{split}
\bd_k &= 
\frac{\sum_{\{i:\ \bm{x}_{e,i} \in \hat\gp_{\data_{\ntrain},k}\} } \{y_{e,i} - \prob(\bm{x}_{e,i})\}}{\sqrt{\sum_{\{i:\ \bm{x}_{e,i} \in \hat\gp_{\data_{\ntrain},k}\} }\left[\prob(\bm{x}_{e,i})\left\{1 -  \prob(\bm{x}_{e,i})\right\}  \right]  }},\\
\csum_{a,1,k} &= \frac{\sqrt{\sum_{\{i:\ \bm{x}_{e,i} \in \hat\gp_{\data_{\ntrain},k}\} }\left[\prob(\bm{x}_{e,i})\left\{1 -  \prob(\bm{x}_{e,i})\right\}  \right]  }}{\sqrt{\sum_{\{i:\ \bm{x}_{e,i} \in \hat\gp_{\data_{\ntrain},k}\} }\left[\hat\prob_{\data_{\ntrain}}(\bm{x}_{e,i})\left\{1 -  \hat\prob_{\data_{\ntrain}}(\bm{x}_{e,i})\right\}  \right]  }},\\
 \csum_{a,2,k} &= \frac{\sum_{ \{i:\ \bm{x}_{e,i} \in \hat\gp_{\data_{\ntrain},k}\}  } \left\{ \prob(\bm{x}_{e,i}) -\hat\prob_{\data_{\ntrain}}(\bm{x}_{e,i}) \right\}  }{\sqrt{\sum_{\{i:\ \bm{x}_{e,i} \in \hat\gp_{\data_{\ntrain},k}\} }\left[\hat\prob_{\data_{\ntrain}}(\bm{x}_{e,i})\left\{1 -  \hat\prob_{\data_{\ntrain}}(\bm{x}_{e,i})\right\}  \right]  }}.
\end{split}
\end{equation*}  
Since for $k = 1,\dots, K_n$, $(\bd_k\cdot \csum_{a,1,k} + \csum_{a,2,k})^2 \geq 0$,  it is enough to show that
 for the $k^*$th group specified in Condition~7,
 \begin{equation}\label{eq_T12conv}
   |\bd_{k^*}\cdot \csum_{a,1,k^*} + \csum_{a,2,k^*}|\big/\sqrt{K_n}\rightarrow_p\infty,
\end{equation} 
as $n\rightarrow\infty$.
In a similar way as the proof of Theorem~3, we can show that $\bd_{k^*}$ converges in distribution to the standard normal distribution as $n\rightarrow\infty$.  Together with Conditions~2 and~8 and the fact that $1/\sqrt{K_n}\leq1/\sqrt{2}$, we have that $|\bd_{k^*}\cdot \csum_{a,1,k^*}|/\sqrt{K_n}$ is bounded in probability. Thus, to obtain \eqref{eq_T12conv}, it suffices to show that 
 \begin{equation}\label{eq_T2conv}
   |\csum_{a,2,k^*}|\big/\sqrt{K_n}\rightarrow_p\infty
\end{equation} 
as $n\rightarrow\infty$.

Since the denominator of $\csum_{a,2,k^*}$ satisfies 
 \begin{equation}\label{eq_demonT2}
     \sqrt{\sum_{\{i:\ \bm{x}_{e,i} \in \hat\gp_{\data_{\ntrain},k^*}\} }\left[\hat\prob_{\data_{\ntrain}}(\bm{x}_{e,i})\left\{1 -  \hat\prob_{\data_{\ntrain}}(\bm{x}_{e,i})\right\}  \right]  }\leq \sqrt{\hat n_{2,k^*}}\big/2,
 \end{equation}
 together with inequalities 
$K_n\leq \ntest/\underline m_n$ and $\underline m_n\leq\hat n_{2,k^*}$, we have
\begin{align}
    |\csum_{a,2,k^*}|\big/\sqrt{K_n} &= \frac{| \csum_{a,2,k^*}|}{\sqrt{\underline m_n}\rate_{\ntrain}^{(a)}} \cdot  \frac{\sqrt{\underline m_n}\rate_{\ntrain}^{(a)}}{\sqrt{K_n}}\nonumber\\
    &\geq \frac{| \csum_{a,2,k^*}|}{\sqrt{\underline m_n}\rate_{\ntrain}^{(a)}} \cdot \frac{\underline m_n  \rate_{\ntrain}^{(a)}}{\sqrt{\ntest}}\nonumber\\
   &\geq\left(\frac{\biggl|\sum_{ \{i:\ \bm{x}_{e,i} \in \hat\gp_{\data_{\ntrain},k^*}\}  } \left\{ \prob(\bm{x}_{e,i}) -\hat\prob_{\data_{\ntrain}}(\bm{x}_{e,i}) \right\} \biggr|}{(\sqrt{\hat n_{2,k^*}}\sqrt{\underline m_n}\rate_{\ntrain}^{(a)}) }\right)\cdot \frac{2\underline m_n  \rate_{\ntrain}^{(a)}}{\sqrt{\ntest}}\nonumber\\
   &\geq\left(\frac{\biggl|\sum_{ \{i:\ \bm{x}_{e,i} \in \hat\gp_{\data_{\ntrain},k^*}\}  } \left\{ \prob(\bm{x}_{e,i}) -\hat\prob_{\data_{\ntrain}}(\bm{x}_{e,i}) \right\} \biggr|}{\hat n_{2,k^*}\rate_{\ntrain}^{(a)}} \right)\cdot \frac{2\underline m_n  \rate_{\ntrain}^{(a)}}{\sqrt{\ntest}}\label{eq_inequth2}, 
\end{align}
almost surely.
According to Condition~1 and 
$\ntest = \Omega((\rate_{\ntrain}^{(a)})^{-6})$, we have
$$ \underline m_n \rate_{\ntrain}^{(a)} /\sqrt{\ntest} = 
\rate_{\ntrain}^{(a)}\cdot\ntest^{1/6}\cdot \underline m_n/\ntest^{2/3} = \omega(1),$$
as $n\rightarrow \infty$.
Thus, to show (\ref{eq_T2conv}), it remains to show that 
\begin{equation}\label{eq_phatRate}
\biggl|\sum_{ \{i:\ \bm{x}_{e,i} \in \hat\gp_{\data_{\ntrain},k^*}\}  } \left\{ \prob(\bm{x}_{e,i}) -\hat\prob_{\data_{\ntrain}}(\bm{x}_{e,i}) \right\} \biggr|\bigg/(\hat n_{2,k^*}\rate_{\ntrain}^{(a)}) = \Omega(1)
\end{equation}
as $n\rightarrow\infty$.

Recall that  $\hat n_{2,k}^{\suset} = \sum_{i = 1}^{\ntest} I{\{\bm{x}_{e,i} \in \hat\gp_{\data_{\ntrain},k}\cap \suset\}}$. 
 We have
 \begin{align}
     &\biggl|\sum_{ \{i:\ \bm{x}_{e,i} \in \hat\gp_{\data_{\ntrain},k^*}\}  } \left\{ \prob(\bm{x}_{e,i}) -\hat\prob_{\data_{\ntrain}}(\bm{x}_{e,i}) \right\} \biggr|\bigg/(\hat n_{2,k^*}\rate_{\ntrain}^{(a)}) \nonumber \\
     &\geq \frac{\hat n_{2,k^*}^{\suset}}{\hat n_{2,k^*}}\cdot \frac{\underset{\bm{x}\in \suset}{\inf} | \hat\prob_{\data_{\ntrain}}(\bm{x})- \prob(\bm{x}) |}{\rate_{\ntrain}^{(a)} }  - \frac{ \hat n_{2,k^*}-\hat n_{2,k^*}^{\suset} }{ \hat n_{2,k^*}\rate_{\ntrain}^{(a)} },\nonumber\\
     &\geq  \frac{\hat n_{2,k^*}^{\suset}}{\hat n_{2,k^*}}\cdot\zeta - \frac{ \hat n_{2,k^*}-\hat n_{2,k^*}^{\suset} }{ \hat n_{2,k^*}\rate_{\ntrain}^{(a)} }, \label{eq_piinequ}
 \end{align}
 almost surely, where the last inequality is due to Condition~7. 
By taking (7) from the main text  into (\ref{eq_piinequ}), we obtain \eqref{eq_phatRate}, which completes the proof.

 \subsection{Proof of Theorem~1 (Convergence of $\bagoft$ for parametric models under $\textrm{H}_0$)}\label{subsec_proof_of_thm2}
By Condition~3, the convergence rate of the parametric classification model is $1/\sqrt{n}$. Therefore, we  can take $\rate_{\ntrain} = 1/\sqrt{\ntrain}$, and the result follows from Theorem~3.

 \subsection{Proof of Theorem~2 (Consistency of $\bagoft$ for parametric models under $\textrm{H}_1$)} \label{subsec_proof_thm2}

By the same reasoning as the proof of Theorem~4, it suffices for us to show \eqref{eq_T2conv}.
First, we can obtain 
$$|\csum_{a,2,k^*}|\big/\sqrt{K_n}\geq\biggl(\biggl|\sum_{ \{i:\ \bm{x}_{e,i} \in \hat\gp_{\data_{\ntrain},k^*}\}  } \left\{ \prob(\bm{x}_{e,i}) -\hat\prob_{\data_{\ntrain}}(\bm{x}_{e,i}) \right\} \biggr|\bigg/\hat n_{2,k^*} \biggr)\cdot 2\underline m_n /\sqrt{\ntest},$$
almost surely, in a similar way as we derive the inequality \eqref{eq_inequth2}. 
 By Condition~1, $\underline m_n/\sqrt{n_2}\rightarrow\infty$ as $n\rightarrow\infty$. We also have
 \begin{align*}
     &\biggl|\sum_{ \{i:\ \bm{x}_{e,i} \in \hat\gp_{\data_{\ntrain},k^*}\}  } \left\{ \prob(\bm{x}_{e,i}) -\hat\prob_{\data_{\ntrain}}(\bm{x}_{e,i}) \right\} \biggr|\bigg/\hat n_{2,k^*} \nonumber \\
     &\geq \frac{\hat n_{2,k^*}^{\suset}}{\hat n_{2,k^*}}\cdot \underset{\bm{x}\in \suset}{\inf} | \hat\prob_{\data_{\ntrain}}(\bm{x})- \prob(\bm{x}) |  - \frac{(\hat n_{2,k^*}-\hat n_{2,k^*}^{\suset})}{\hat n_{2,k^*}}\\
     &\geq \frac{\hat n_{2,k^*}^{\suset}}{\hat n_{2,k^*}}\cdot \cst   - \frac{(\hat n_{2,k^*}-\hat n_{2,k^*}^{\suset})}{\hat n_{2,k^*}},
 \end{align*}
with probability going to one as $n\rightarrow\infty$, where the last inequality is due to Condition~5. 
According to (6) from the main text, the right-hand side of the above inequality  is lower bounded away from 0 in probability.
Thus, we complete the proof. 
 
%  \section{Proof of Corollary~\ref{lpsc} (Obtaining both size control and consistency for learning procedures)}
% We have
% $$\ntest = \Omega((\rate_{\ntrain}^{(a)})^{-6}),$$
% as $n\rightarrow \infty$.
% By Condition~1,
% $$\rate_{\ntrain}^{(a)} = \Omega(\ntest^{-1/6}\cdot \ntest^{2/3}/\underline m_n) = \Omega(\sqrt{\ntest}/\underline m_n),$$
% as $n\rightarrow \infty$.
% %Moreover, the requirement  $\xi_n = \omega(\rate_{\ntrain}^{-4/5})$ guarantees the existence of $\ntest$ such that $\ntest = \Omega((\rate_{\ntrain}\xi_n)^{-6})$ and $\ntest = o(\rate_{\ntrain}^{-6/5})$.
% So by Theorem~4, we obtain the asymptotic consistency of the BAGofT. The asymptotic size control of the BAGofT is obtained by a direct application of the Theorem~3.

\section{Identifying deviation sets} \label{supp_identify}

Recall that Algorithm~1 is based on the sets generated from $K$-quantiles of the fitted Pearson residuals on the training set. 
In this section, we justify our approach by showing that under some reasonable conditions,  the sets generated from the $K$-quantiles of the fitted Pearson residuals satisfy Condition~7. We focus on assessing general classification procedures. A similar result holds for testing parametric models, and we omit it for brevity.

Some additional notations are introduced below.
For $\bm{x}\in\mathbb S$, let
$$
q_{\ntrain}(\bm{x}) = \frac{\hat\prob_{\data_{\ntrain}}(\bm{x}) - \prob(\bm{x})}{\sqrt{\hat\prob_{\data_{\ntrain}}(\bm{x})(1-\hat\prob_{\data_{\ntrain}}(\bm{x}))}}.
$$
Let 
\begin{align}
\bm{x} \mapsto \hat q_{\ntrain}(\bm{x}) \label{eq_Rhat}
\end{align}
denote the Random Forest (or other regression methods) fitted from the 
response $$\frac{\hat\prob_{\data_{\ntrain}}(\bm{x_{t,i}}) - y_{t,i}}{\sqrt{\hat\prob_{\data_{\ntrain}}(\bm{x_{t,i}})(1-\hat\prob_{\data_{\ntrain}}(\bm{x_{t,i}}))}},$$ and covariate $\bm{x_{t,i}}$, where each $(\bm{x_{t,i}}, y_{t,i})$ is from the training set $\data_{\ntrain}$.

Since for most applications, relatively simple sets may be sufficient to reveal the systematic defects from the PTA, we consider sets from a Glivenko-Cantelli class defined as follows. 
\begin{definition}[Glivenko-Cantelli class]\label{def_GC}
A collection of sets $$\mathcal G\subseteq \{G: G\text{ is a $P_{\bm{x}}$-measurable subset of }\mathbb S\}$$  
is called a Glivenko-Cantelli (GC) class
if 
$$
\sup_{G\in\mathcal G}\biggl|\frac{1}{n}\sum_{i=1}^nI\{\bm{x_i}\in G\} - P(\bm{x}\in G)\biggr|\limp 0,
$$
as $n\rightarrow\infty$, where $\bm{x}_i$ ($i=1,\dots,n$) are i.i.d.\ from $P_{\bm{x}}$.

\end{definition}
It is well known that a class with a finite Vapnik-Chervonenkis (VC) dimension is a GC class. For example, the collection of all the rectangular sets in $\mathbb R^p$ has a VC dimension of $2^p$, which guarantees the collection to be a GC class when $p$ is fixed. In practice, when $p$ is large, we may restrict our attention to a selected sparse subset of variables.
% In the following, assume $\mathcal G$ denote a chosen GC class.

% The required conditions are as follows.
\begin{condition}[Accurate bias estimation]\label{lem_pbe}
We have
\begin{equation}\label{eq_rConv}
      \operatorname*{ess~\sup}_{\bm{x}\in\mathbb S}|\hat q_{\ntrain}(\bm{x}) - q_{\ntrain}(\bm{x})|= o_p(\rate_{\ntrain}^{(a)})
\end{equation}
as $\ntrain \rightarrow\infty$.
\end{condition}

\begin{condition}[Existence of a slow convergence set]\label{lem_scs}
Under $H_1$, we have a GC class $\mathcal G$ such that with probability going to one, there exists $\suset'\in\mathcal G$ that may depend on $\data_{\ntrain}$, with $P(\bm{x}\in\suset')$ being lower bounded by a positive constant, and
\begin{align}
\operatorname*{ess~\inf}_{\bm{x}\in\suset'} ( \hat\prob_{\data_{\ntrain}}(\bm{x})- \prob(\bm{x}) )&\geq 0, \quad\text{or} \label{eq_phatpG0}\\
\operatorname*{ess~\sup}_{\bm{x}\in\suset'} ( \hat\prob_{\data_{\ntrain}}(\bm{x})- \prob(\bm{x}) )&\leq 0, \quad\text{and} \nonumber
 \end{align}
\begin{equation}\label{qeq_qn1_rn1a_bound}
    \underset{\bm{x}\in \suset'}{\inf} | q_{\ntrain}(\bm{x}) |/\rate_{\ntrain}^{(a)} \geq\zeta_1 \text{ almost surely},
\end{equation}
for a positive constant $\zeta_1$. 
% for a positive constant $\zeta_1$ and a positive sequence $\rate_{\ntrain}^{(a)}\rightarrow 0$ as $n\rightarrow\infty$. 
\end{condition}
\begin{condition}[No residual collision]\label{cond_noConc} For each $\ntrain\in \mathbb N$ and each pair $\bm{x^{(1)}}, \bm{x^{(2)}}$ from $\data_{\ntrain}$, we have
$$\hat q_{\ntrain}(\bm{x^{(1)}})\neq \hat q_{\ntrain}(\bm{x^{(2)}})\quad and\ 
q_{\ntrain}(\bm{x^{(1)}})\neq  q_{\ntrain}(\bm{x^{(2)}}),$$
 almost surely.
\end{condition}

The above Condition~\ref{lem_pbe} requires the convergence speed  of the method (Random Forest) we used to fit the Pearson residual on the training set to be faster than that of the PTA under $H_1$ (measured by $\rate_{\ntrain}^{(a)}$). The set $\suset'$ from Condition~\ref{lem_scs} is similar to $\suset$ from Condition~7,  except that it is  from a Glivenko-Cantelli class, which is needed to obtain some desirable properties of the quantiles of the Pearson residuals. With Condition~\ref{cond_noConc}, we exclude the cases where some sample quantiles do not exist for technical convenience. 

%The following theorem considers the case for assessing general procedures.  %We can get the same result for testing parametric models with a similar proof and $r_n^{(a)}$ replaced by constant $1$.
\begin{theorem}[Identifying a slow-convergence set] \label{thm_identi_slow_set}
Assume that Conditions~\ref{lem_pbe}-\ref{cond_noConc} hold. 
Then, with $K$ large enough, there exists $\suset\in \mathcal G$ that  satisfies  Condition~7.

\end{theorem}

\textit{Proof of Theorem~\ref{thm_identi_slow_set}}:

We prove in two steps. 
First, we show that for each $\bm{x}$ in 
$$
\{\bm{x_{t,1}},{\dots}, \bm{x_{t,\ntrain}}\}\cap \{ \bm{x}: \text{$\hat q_{\ntrain}(\bm{x})$ is larger or equal to the upper $K$-quantile of $\hat q_{\ntrain}(\bm{x_{t,i}})$}\},
$$
we have that $\hat q_{\ntrain}(\bm{x})/\rate_{\ntrain}^{(a)}$ is bounded below by a positive constant with probability going to one. Second, based on the above result, we complete the proof by generating a  set that satisfies the requirements of $\suset$ in Condition~7. Without the loss of generality, we assume that \eqref{eq_phatpG0} holds. The proof under the other case is similar.

Let $\floor{x}$ denote the largest integer less than or equal to $x$, and $p_l>0$ denote the lower bound of $P(\bm{x}\in \suset')$ (where $\suset'$ is defined in Condition~\ref{lem_scs}).
We arbitrarily choose a constant $\epsilon_1 \in (0, p_l)$. We let $p_0 = p_l - \epsilon_1$, and let
$\mathcal H$ denote the event that there are at least $\floor{\ntrain p_0}$ observations of $\bm{x_{t,i}}$ with $\bm{x_{t,i}}\in \suset'$.
Since $\suset'\in \mathcal G$ (Condition~\ref{lem_scs}), \begin{align}
   P(\mathcal H)
   \geq& P\biggl(\sum_{i=1}^{\ntrain} I_{\{\bm{x_{t,i}}\in \suset'\}}  \geq \ntrain p_0\biggr)\nonumber\\
    \geq& P\biggl(\sum_{i=1}^{\ntrain} I_{\{\bm{x_{t,i}}\in \suset'\}}  \geq \ntrain (P(\bm{x}\in \suset') - \epsilon_1) \biggr)\nonumber\\
   \geq &P\biggl( \biggl|\frac{1}{\ntrain}\sum_{i=1}^{\ntrain} I_{\{\bm{x_{t,i}}\in \suset'\}} - P(\bm{x}\in \suset')\biggr| \leq\epsilon_1\biggr)\nonumber\\
    \geq & P\biggl( \sup_{G\in \mathcal G }\biggl|\frac{1}{\ntrain}\sum_{i=1}^{\ntrain} I_{\{\bm{x_{t,i}}\in  G\}} - P(\bm{x}\in G)\biggr| \leq\epsilon_1\biggr)\rightarrow 1,\label{eq_Hp1}
\end{align}
as $\ntrain \rightarrow\infty$, where the last limit is due to Definition~\ref{def_GC}. For the remaining part of the proof, we conditional on $\mathcal H$.

Let $\bm{x^{(p_0)}}$ denote the covariate observation such that  $q_{\ntrain}(\bm{x^{(p_0)}})$ is 
the upper $1/p_0$-quantile of $q_{\ntrain}(\bm{x_{t,i}})$ with $i=1,\dots,\ntrain$. If $\ntrain p_0$ is not an integer, we require $q_{\ntrain}(\bm{x^{(p_0)}})$ to be the smallest one that is larger or equal to the $1/p_0$-quantile instead. As a result, there are exactly $\floor{\ntrain p_0}$ observations of $\bm{x_{t,i}}$ with $q_{\ntrain}(\bm{x_{t,i}})\geq q_{\ntrain}(\bm{x^{(p_0)}})$.
Let $\mathcal Q_{\ntrain}$ denote such a set, namely
\begin{equation}\label{eq_defQn}
   \mathcal Q_{\ntrain}= \{\bm{x}\in \{\bm{x_{t,1}},{\dots},\bm{x_{t,\ntrain}}\}: q_{\ntrain}(\bm{x})\geq q_{\ntrain}(\bm{x^{(p_0)}})\}. 
\end{equation}
Therefore, $ \mathcal Q_{\ntrain}$ includes  $\floor{\ntrain p_0}$  observations from $ \{\bm{x_{t,1}},{\dots},\bm{x_{t,\ntrain}}\}$ with the largest residuals $q_{\ntrain}(\bm{x})$. 
Next, we show by contradiction that on $\mathcal H$,
\begin{equation}\label{eq_QgeqRZeta}
   q_{\ntrain}(\bm{x_{t,i}}) \geq \rate_{\ntrain}^{(a)} \zeta_1,\quad\forall \bm{x_{t,i}}\in\mathcal Q_{\ntrain}.
\end{equation} 
If \eqref{eq_QgeqRZeta} does not hold, there exists $\bm{x_0}\in\mathcal Q_{\ntrain}$ such that
$$
q_{\ntrain}(\bm{x^{(p_0)}})\leq q_{\ntrain}(\bm{x_0})<\rate_{\ntrain}^{(a)} \zeta_1.
$$
By Condition~\ref{lem_scs}, 
\begin{equation}\label{eq_Qgeqrz2}
    q_{\ntrain}(\bm{x_{t,i}}) = |q_{\ntrain}(\bm{x_{t,i}})|\geq \rate_{\ntrain}^{(a)} \zeta_1,\quad \forall \bm{x_{t,i}}\in \suset'
\end{equation}
holds. Thus, for each $\bm{x_{t,i}}\in \suset'$,
$$
q_{\ntrain}(\bm{x_{t,i}}) > q_{\ntrain}(\bm{x_0})\geq q_{\ntrain}(\bm{x^{(p_0)}}).
$$
According to the above result, we have $\bm{x_0}\notin\suset'$ and
\begin{align}
& (\{\bm{x_{t,1}},{\dots}\bm{x_{t,\ntrain}}\} \cap \suset') \subset \mathcal Q_{\ntrain}.
% \quad\{\bm{x}\in\{\bm{x_{t,1}},{\dots}\bm{x_{t,\ntrain}}\}:\bm{x}\in \suset'\}\subset \mathcal Q_{\ntrain}. 
\label{eq_MinQ}
\end{align}
Since we assume $\mathcal H$ holds, by combining \eqref{eq_MinQ}, the fact that $
\bm{x_0}\in\mathcal Q_{\ntrain}$, and the definition of $\mathcal H$, we conclude that there are at least $1 + \floor{\ntrain p_0}$ elements in $\mathcal Q_{\ntrain}$. This contradicts the fact that $Q_{\ntrain}$ contains exactly $\floor{\ntrain p_0}$ elements. Therefore, we obtain \eqref{eq_QgeqRZeta}.

Next, we define $\bm{x^{(\hat p_0)}}$ in a similar way as $\bm{x^{(p_0)}}$ except that $q_{\ntrain}(\cdot)$ is replaced by $\hat q_{\ntrain}(\cdot)$. 
In order to complete the first step of the proof,  we establish inequalities between $\hat q_{\ntrain}(\bm{x^{(\hat p_0)}})$ and $\min_{\bm{x_{t,i}}\in\mathcal Q_{\ntrain}} \hat q_{\ntrain}(\bm{x_{t,i}})$. By combining the result from \eqref{eq_QgeqRZeta} and  Condition~\ref{lem_pbe}, for an arbitrary $\zeta_2 \in (0,\zeta_1)$, we have
\begin{equation}\label{eq_pQgeqRzeta}
    P\biggl(\min_{\bm{x_{t,i}}\in\mathcal Q_{\ntrain}} \hat q_{\ntrain}(\bm{x_{t,i}}) \geq \rate_{\ntrain}^{(a)} (\zeta_1-\zeta_2)\biggr)\rightarrow 1,
\end{equation}
as $\ntrain\rightarrow\infty$. By the definition of $\bm{x^{(\hat p_0)}}$, we have
\begin{equation}\label{eq_QhatgeqQhat}
   \hat q_{\ntrain}(\bm{x^{(\hat p_0)}})\geq \min_{\bm{x_{t,i}}\in\mathcal Q_{\ntrain}} \hat q_{\ntrain}(\bm{x_{t,i}}) 
\end{equation}
 holds almost surely. 
%If the inequality~\eqref{eq_QhatgeqQhat} does not hold, for each $\bm{x_{t,i}}\in\mathcal Q_{\ntrain}$,
%\begin{equation}\label{eq_qHatQn_geq_aHatp0}
%	\hat q_{\ntrain}(\bm{x_{t,i}})  > \hat q_{\ntrain}(\bm{x^{(\hat p_0)}}).
%\end{equation}
%Since $\mathcal Q_{\ntrain}$ has $\floor{{\ntrain}p_0}$ elements, by \eqref{eq_qHatQn_geq_aHatp0}, we have at least $1 + \floor{{\ntrain}p_0}$ observations of $\bm{x_{t,i}}$ with $\hat q_{\ntrain}(\bm{x_{t,i}})  \geq \hat q_{\ntrain}(\bm{x^{(\hat p_0)}})$. It
% contradicts with  the definition of $\hat q_{\ntrain}(\bm{x^{(\hat p_0)}})$. Thus, we obtain the Inequality~\eqref{eq_QhatgeqQhat}.
  Combining the results from \eqref{eq_pQgeqRzeta} and \eqref{eq_QhatgeqQhat}, we have 
\begin{equation}\label{eq_QgeqZeta23}
     P\biggl(\hat q_{\ntrain}(\bm{x^{(\hat p_0)}}) \geq \rate_{\ntrain}^{(a)} (\zeta_1-\zeta_2)\biggr)\rightarrow 1,
\end{equation}
as ${\ntrain}\rightarrow\infty$.

In the remaining step of our proof, we show that the set
\begin{equation}\label{eq_susetDef}
\suset =\{\bm{x}\in \mathbb S: \hat q_{\ntrain}(\bm{x})\geq \rate_{\ntrain}^{(a)} (\zeta_1-\zeta_2)\}
\end{equation}
satisfies the requirements in Condition~7. First, recall that 
$$
\operatorname*{ess~\inf}_{\bm{x}\in\suset} ( \hat\prob_{\data_{\ntrain}}(\bm{x})- \prob(\bm{x}) )\geq 0,
$$ which holds by our assumption without losing generality. Second, according to the definition of $\suset$ in \eqref{eq_susetDef} and  Condition~8, we have
$$
\underset{\bm{x}\in \suset}{\inf} | \hat\prob_{\data_{\ntrain}}(\bm{x})- \prob(\bm{x}) |/\rate_{\ntrain}^{(a)} \geq\zeta
$$
holds almost surely with 
 $\zeta = \sqrt{\cst_3(1-\cst_3)}\cdot (\zeta_1-\zeta_2)$.
 
 Third, we show that $P(\bm{x}\in\suset)$ is lower bounded by a positive constant.
By \eqref{qeq_qn1_rn1a_bound} from Condition~\ref{lem_scs}, we have
\begin{align}
	P(\bm{x}\in\suset )
	=&P(\hat q_{\ntrain}(\bm{x}) \geq \rate_{\ntrain}^{(a)} (\zeta_1-\zeta_2))\nonumber\\
	\geq&
P\biggl(\{\operatorname*{ess~\sup}_{\bm{x}\in\mathbb S}|\hat q_{\ntrain}(\bm{x}) - q_{\ntrain}(\bm{x})|<\rate_{\ntrain}^{(a)} \zeta_2\}\cap \{\bm{x}\in \suset'\}\biggr)\nonumber\\
\geq&  P\biggl(\operatorname*{ess~\sup}_{\bm{x}\in\mathbb S}|\hat q_{\ntrain}(\bm{x}) - q_{\ntrain}(\bm{x})|<\rate_{\ntrain}^{(a)} \zeta_2\biggr) + P(\bm{x}\in \suset') - 1.\label{eq_pXinMn_lowerbound}
\end{align}
By Conditions~\ref{lem_pbe}, 
%for each $\bm{x}\in\suset'$,
\begin{equation}\label{eq_qhat_minus_q_bound}
    P\biggl( \operatorname*{ess~\sup}_{\bm{x}\in\mathbb S}|\hat q_{\ntrain}(\bm{x}) - q_{\ntrain}(\bm{x})|<\rate_{\ntrain}^{(a)} \zeta_2\biggr)\rightarrow 1.
\end{equation}
Combining \eqref{eq_pXinMn_lowerbound} and \eqref{eq_qhat_minus_q_bound}, and by the fact that $P(\bm{x}\in\suset')$ is lower bounded away from 0 (Condition~\ref{lem_scs}), we have that $P(\bm{x}\in\suset )$ is lower bounded away from 0 when $\ntrain$ is sufficiently large.

 Lastly, for the requirement~(7) in Condition~7, 
 we consider the case with $K$ sufficiently large, such that $K>1/p_0$. Next, we index the
 set generated from the upper $K$-quantile of $\hat q_{\ntrain}(\bm{x_i})$, which is the group from Algorithm~1 with the largest Pearson residual, by $k^*$. For each $\bm{x}$ in this group,  by $K>1/p_0$, we have $\hat q_{\ntrain}(\bm{x})\geq \hat q_{\ntrain}(\bm{x^{(\hat p_0)}})$. 
% Therefore,
%  By \eqref{eq_QgeqZeta23}, \begin{equation}\label{eq_pGeqpra1}
%  P\biggl(\hat q_{\ntrain}(\bm{x}) \geq \rate_{\ntrain}^{(a)} (\zeta_1-\zeta_2)\mid \text{$\bm{x}$ is in the $k^*$ group}\biggr)\rightarrow 1,
% \end{equation}
% as $\ntrain \rightarrow\infty$.
 Combining this inequality and the definition of $\suset$ in \eqref{eq_susetDef}, we have
\begin{align*}
    %P\biggl(\frac{\hat n_{2,k^*}^{\suset}}{\hat n_{2,k^*}}=1\biggr)\rightarrow 1,\\
   P\biggl( \frac{\hat n_{2,k^*}-\hat n_{2,k^*}^{\suset}}{\hat n_{2,k^*}\rate_{\ntrain}^{(a)} }=0\biggr) =&\ P\biggl( \frac{\hat n_{2,k^*}-\hat n_{2,k^*}^{\suset}}{\hat n_{2,k^*}}=0\biggr) \\
   \geq &\ P\biggl(\hat q_{\ntrain}(\bm{x^{(\hat p_0)}})\geq \rate_{\ntrain}^{(a)} (\zeta_1-\zeta_2)\biggr)\rightarrow 1,
\end{align*}
as $\ntrain\rightarrow\infty$. Thus, we complete the proof.

\section{A discussion about the conditions on the parametric model experimental studies}\label{sec_diss_Cond_parametricExperiment}
%%%%%%%%%%%%%%%%%%%%%%%%%%%%%%%%%%%%%%%%%%%%%%%%%%%%%%%%%%%%%%%%%%%%%%%
%Idea for proving Condition 4.
% First, when beta = 0, delta = -log(2) (beta = 0 is the special beta^*)
% Second, as xbeta goes to infinity, pixbeta - log(1 + e^{xbeta}) goes to negative infinity
%%%%%%%%%%%%%%%%%%%%%%%%%%%%%%%%%%%%%%%%%%%%%%%%%%%%%%%%%%%%%%%%%%%%%%%
In the simulation studies, Condition~1 is met via the algorithm implementation (recall that $D_{x_e}$ is used to control the group sizes). For the remaining conditions, we first consider \textbf{Setting 2} with covariates from uniform distribution. Since the support $\mathbb S$ is compact, $\pi(\bm{x})$ from the logistic regression data-generating model is bounded away from 0 and 1. Thus, Condition~2 is satisfied. For testing Model \textbf{A}, by Corollary~1 of \cite{fahrmeir1985consistency}, since the smallest eigenvalue of $\bm{X}^\T \bm{X}$ ($\bm{X}$ is the $n\times p$ design matrix) goes to infinity in probability as $n\rightarrow\infty$, we have the $\sqrt{n}$-consistency of the estimated coefficients $\bm{\hat\beta}$. Together with the compactness of $\mathbb S$ and mean value theorem, we obtain Condition~3. For Condition~4 in testing Model \textbf{B}, let 
% $$
% \pi^{(m)}( \bm{x_{i}}) = \frac{1}{1 + e^{-\beta_0 - \beta_1 x_{i,1}-\beta_2 x_{i,2}}},
% $$
% which follows from the misspecified MTA and
% $$
% {\color{red}\delta_n(\bm{\beta})} =
% \begin{pmatrix}
% \sum_{i=1}^n (\pi(\bm{x_{i}}) - \pi^{(m)}( \bm{x_{i}}))\\
% \sum_{i=1}^n x_{i,1}(\pi(\bm{x_{i}}) - \pi^{(m)}( \bm{x_{i}}))\\
% \sum_{i=1}^n x_{i,2}(\pi(\bm{x_{i}}) - \pi^{(m)}( \bm{x_{i}}))
% \end{pmatrix},
% $$
$$
\delta_n(\bm{\beta})= \frac{1}{n}\sum_{i=1}^n\bigl(\pi(\bm{x_{i}})\cdot \bm{x_{i}}^\T\bm{\beta} - \log(1 + e^{\bm{x_{i}}^\T\bm{\beta}})\bigr), 
$$
which is the expected sample log-likelihood conditional on the covariates and plays an important role in the asymptotic theory of $\bm{\hat\beta}$ from the MTA \citep{fahrmexr1990maximum}. Let $\bm{\hat\beta_n} = \arg\max_{\bm{\beta}\in\mathbb R^3}\delta_n(\bm{\beta})$. It can be verified that $\bm{\hat\beta_n}$ exists, and $\|\bm{\hat\beta_n}\|_2$ is upper bounded with probability going to 1 as $n\rightarrow\infty$. Therefore, Condition~4 can be verified by Theorem~1 of \cite{fahrmexr1990maximum}.
%According to the properties of the logistic regression model with linearly independent covariates, the Hessian matrix $\frac{d^2}{d\bm{\beta^2}}\delta_n(\bm{\beta})$ is negative definite with probability going to 1 as $n\rightarrow\infty$. Together with Theorem~1 of \cite{fahrmexr1990maximum}, for the verification of Condition~4, it suffices to show that $\bm{\hat\beta_n}$ exists, and $\|\bm{\hat\beta_n}\|_2$ is upper bounded with probability going to 1 as $n\rightarrow\infty$.
%It can be shown with additional derivations that when $\|\bm{\beta}\|_2$ is large enough, $\|\delta_n(\bm{\beta})\|_2$ is upper bounded, and there exists $\bm{\beta^*}\in\mathbb R^3$ with $\|\delta_n(\bm{\beta^*})\|_2$ larger or equal to that upper bound with probability going to 1. Therefore, Condition~4 is verified. 
Since $\hat\pi( \bm{x})$ converges to a function that is different from $\pi( \bm{x})$, and both $\hat\pi( \bm{x})$  and $\pi( \bm{x})$ are continuous with respect to $\bm{\beta}$, we obtain Condition~5.

For \textbf{Settings 1\&3}, Condition~2 is not strictly guaranteed since with normal and chi-squared covariates, the probabilities are not bounded away from 0 and 1. Also, the remaining conditions are hard to verify. Nevertheless, our experiment results show that those assumptions are not critical to obtain the desirable performance of the BAGofT. 
% To ensure Conditions~2 and 3 under the setting of  Model \textit{A} and Conditions~2, 4, and 5 under the setting of  Model \textit{B}, one may assume the compactness of the parameter set and covariates support \citep{fahrmeir1985consistency, fahrmexr1990maximum}. Our simulation settings violate these assumptions since the ranges of some covariates are unbounded,  and we do not restrict the estimated parameters to a compact set. Nevertheless, our experiment results show that those assumptions are not critical to obtain the desirable performance of the BAGofT.

 \section{Testing parametric models:\ Q-Q plots under the null hypothesis} \label{supp_qq}

\begin{figure}[!ht]
  \caption{The  Q-Q plots of  the BAGofT bootstrap \textit{p}-values from Model \textit{A} versus $\textrm{Uniform}[0,1]$
   distribution in Settings~2 and~3. The \textit{x}-axis and \textit{y}-axis correspond to the  theoretical quantiles and observed sample quantiles, respectively. }\label{qplot_others}
  \centering
    \includegraphics[width=1\textwidth]{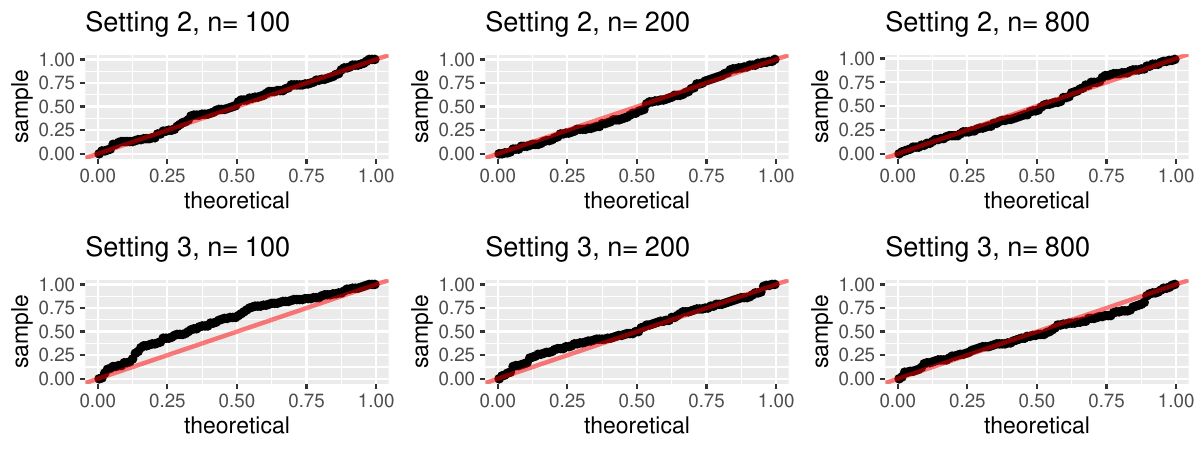}
\end{figure}

 Here, we present the Q-Q plots of Model \textit{A} in Settings~2 and~3 from Section~5.1 of the main text. The results are shown in Figure~\ref{qplot_others}. 

 %%%%%%%%%%%%%%%%%%%%%%%%%%%%%%%%%%%%%%%%%%%%%%%%%%%%%%%%%%%%%
 %Rcodes99_3dplots
 %%%%%%%%%%%%%%%%%%%%%%%%%%%%%%%%%%%%%%%%%%%%%%%%%%%%%%%%%%%%%
 \section{Graphical illustrations of the partitions in the BAGofT and HL test}
 \label{supp_graph}

\begin{figure}[!ht]
  \centering
  \includegraphics[width=1\textwidth]{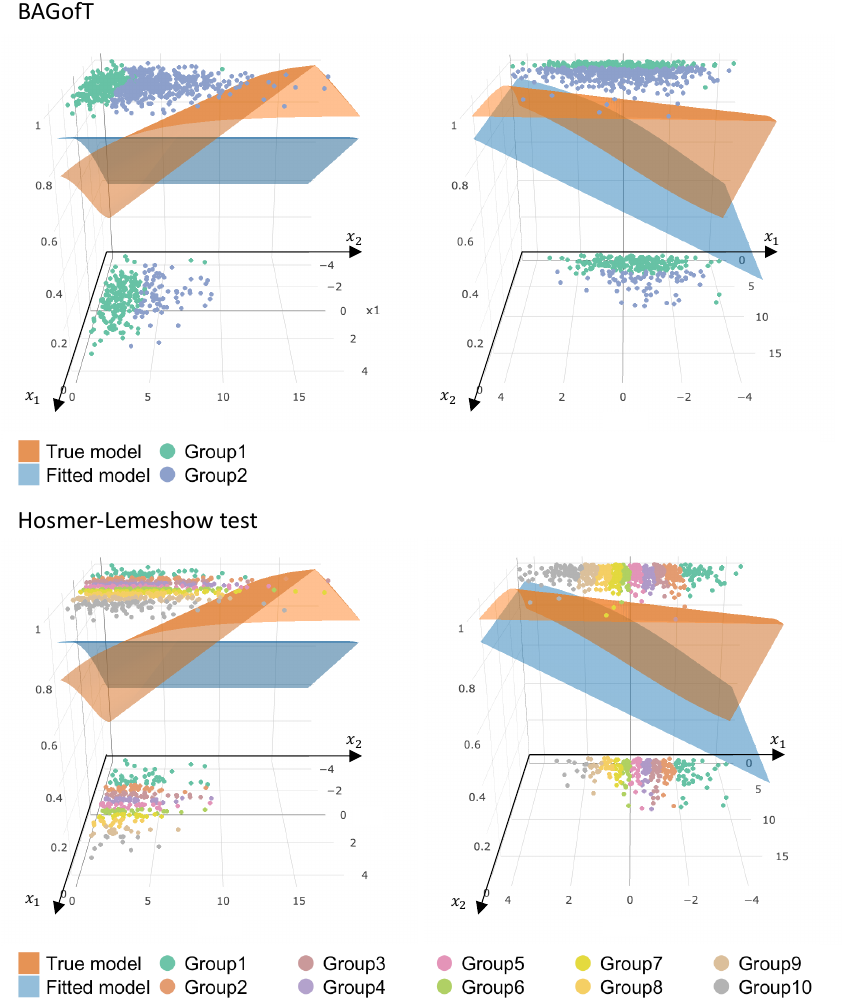}
  \caption{We visualize the data generating model by the orange surface that varies with both $x_1$ and $x_2$, as well as the MTA by the blue surface which does not vary with (missing) $x_2$. 
  The dots in different colors are observations in different groups.
  %The vertical axis is for the values of observed responses and data-generating model probabilities. 
%The $x_1$-axis and $x_2$-axis correspond to the values of two covariates. 
%The dots on the top and bottom planes represent the observations with $y=1$ and $y=0$, respectively. Different  groups in the HL  test and the BAGofT are marked by different colors.  
}\label{s5BAG}
\end{figure}

\begin{figure}[!ht]
  \centering
  \includegraphics[width=1\textwidth]{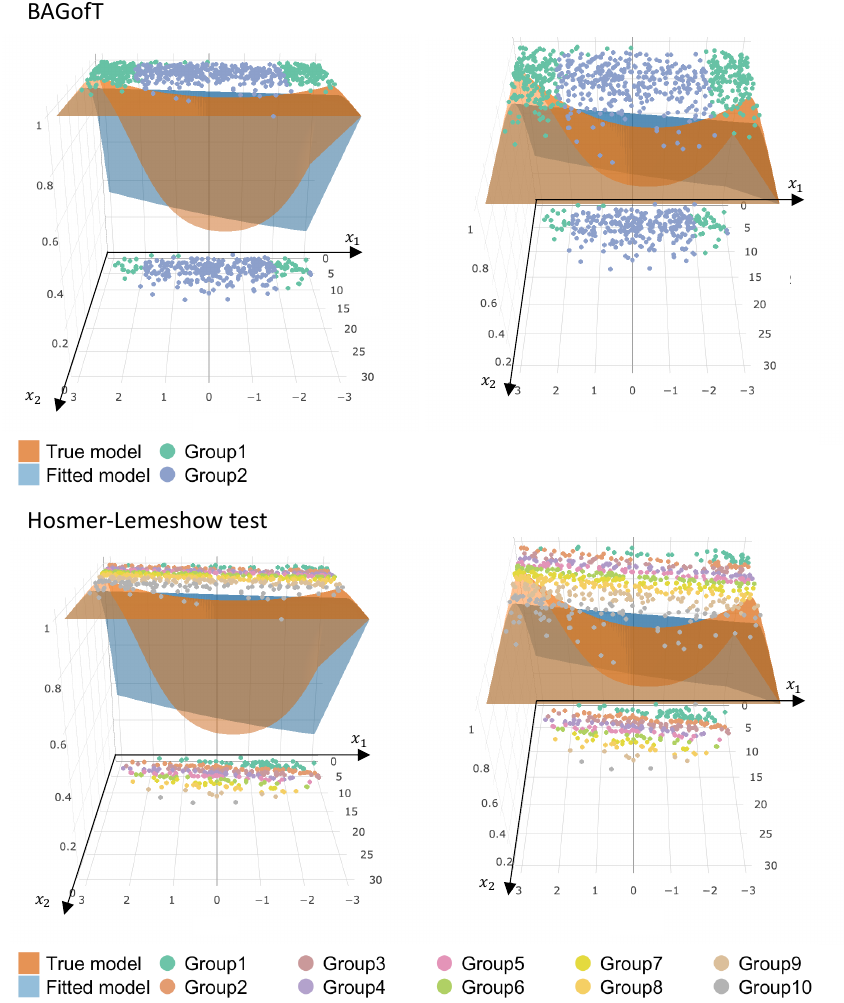}
  \caption{We visualize the data generating model by the orange surface  that is parabolically related with  $x_1$, and the MTA  is the blue surface linear in $x_1$ and $x_2$. Other settings are the same as Figure~\ref{s5BAG}.
}\label{S5HL}
\end{figure}

To illustrate the efficiency of our adaptive partition, we  compare the partition in the BAGofT with the one in the HL test. The following two settings are considered in our study.
\begin{description}
\item[Setting 1.] Generate the data from the Bernoulli distribution with
\[P(y = 1|x_1,x_2) = 1/(1 + \exp(-(0.267x_1 + 0.267x_2))),\]
where $x_1$ and $x_2$ are independently generated from $\N(0,2.25)$, and $\chi^2_4$,  respectively. The MTA is
$$P(y = 1|x_1) = 1/(1 + \exp(-(\beta_0 + \beta_1x_1))).$$

\item[Setting 2.] Generate the data from the Bernoulli distribution with
\[P(y = 1|x_1,x_2) = 1/(1 + \exp(-(-2 + 0.3x_1 + 0.3x_2 + 0.3x_1^2))),\]
where $x_1$ and $x_2$ are independently generated from $\textrm{Uniform}[-3,3]$, and $\chi^2_4$,  respectively. The MTA is 
$$P(y = 1|x_1) = 1/(1 + \exp(-(\beta_0 + \beta_1x_1 + \beta_2x_2))).$$
\end{description}
We visualize the data generating models, the sample points, and the fitted models together with
the partition result from the BAGofT and HL test in Figure~\ref{s5BAG} and Figure~\ref{S5HL}.
Note that an efficient GOF test that based on grouping will partition the part where the data-generating model (orange surface) is higher than the fitted model (blue surface) and the part that is not into different groups. 

In Setting 1, the fitted model misses the covariate $x_2$. Therefore, the fitted model surface in Figure~\ref{s5BAG} does not change with $x_2$. 
Since the fitted probability is only related to $x_1$, the partition boundaries in the 
HL  test are vertical
to $x_1$-axis. However, this partition cancels the difference between the fitted surface and the data-generating model surface, since half 
of the fitted model surface is above the data-generating model surface and the other half is below. For the BAGofT, it can be seen that  the partition lines
are parallel to the $x_1$ axis. 
Furthermore, this adaptive partition divides the part that the data-generating model surface is lower than the fitted model surface and the part that the data-generating model surface is higher than the fitted model surface into different groups, thus producing larger power for the GOF test. 
In Setting 2, since the fitted model misses a quadratic term, we also have a part of the data-generating model surface higher than the fitted model surface
and the other part lower than the fitted model surface in Figure~\ref{S5HL}. We can see that the partition of the BAGofT is again better than the partition of the HL test in this case. 

\clearpage

% \section{Simulation studies for the variable pre-selection}
% In the following Subection~\ref{supp_preselection}, we compare the performance of the BAGofT with variable pre-selection to the one without pre-selection in a high dimensional setting. In  Subection~\ref{supp_grp_comparison}, we compare the performance of the BAGofT with variable pre-selection with the GRP test.

%%%%%%%%%%%%%%%%%%%%%%%%%%%%%%%%%%%%%%%%%%%
% Rcodes110_HD_HDSEL Rcodes111_HD_HDSEL
%%%%%%%%%%%%%%%%%%%%%%%%%%%%%%%%%%%%%%%%%%%

 \section{%High dimensional setting 1: testing a low dimensional model
A comparison between the BAGofT with and without covariates pre-selection %extra
}\label{supp_preselection}

% In this setting, we study the performance of the BAGofT when we have a low dimensional model to assess but there is a large number of covariates available for the goodness-of-fit test. To our best knowledge, there is no existing method that we can use for comparison. So we only study the effect of variable pre-selection in the BAGofT as mentioned in Section~\ref{parAlg}. 
In this section, we show by simulations that a variable pre-selection by the distance correlation 
between the Pearson residual and covariates can significantly improve the performance of the BAGofT in high dimensional settings with many covariates.

We consider the high dimensional setting with 500 covariates and the sample size of 800.  We generate the data from the Bernoulli distribution with the following settings.
\begin{description}
\item[Setting 1.]
$$P(y=1\mid x_1,\dots,x_{500}) = 1/(1 + \exp(-(\beta_1x_1 +\cdots + \beta_5x_5 + \beta_6 x_1x_2))).$$
\item[Setting 2.]
$$P(y=1\mid x_1,\dots,x_{500}) = 1/(1 + \exp(-(\beta_1x_1 +\cdots + \beta_5x_5 + \beta_6 x_1^2))).$$
\end{description}
The covariates $x_1,\dots,x_{500}$ are independently  generated from the multivariate normal distribution with mean $\bm{0}$  and covariance matrix $(\Sigma)_{i,j} = 0.4^{|i-j|}$. The MTA is 
\begin{equation}\label{eq_hdcMTA}
    P(y=1\mid x_1,\dots,x_{5}) = 1/(1 + \exp(-(\beta_0 + \beta_1x_1 +\cdots + \beta_5x_5))).
\end{equation}

We first randomly generate the sample data and apply both the BAGofT that pre-selects 5 covariates out of the 500 available ones and the one without pre-selection. Note that we only care about the overall rejection rates rather the a single outcome. According to this, we take 1 data splitting only to save computational time. The above process is repeated with 100 independent replications and the rejection rates are summarized in Table~\ref{tab_fixcoef} and Table~\ref{tab_rancoef}.

In Table~\ref{tab_fixcoef}, the data are generated with $\beta_1=\cdots=\beta_5 = 1$, $\beta_6=0$, $1$ or $2$ in setting 1 and $\beta_6=0$, $0.5$ or $1$ in setting 2, respectively.  When $\beta_6=0$, both the pre-selected and not pre-selected BAGofT have approximately controlled sizes. When $\beta_6$ gets larger, the pre-selected BAGofT has larger power than the counterpart without pre-selection. Table~\ref{tab_rancoef} is from a more comprehensive  study where $\beta_1,\dots,\beta_5$ are randomly generated from $\N(0,1)$ and $\beta_6$ is randomly generated from $\N(0,\sigma_6^2)$, with $\sigma_6^2 = 0$ ($\beta_6 = 0$), $1$, or $2$ in setting 1 and $\sigma_6^2 = 0$, $0.5$ or $1$ in setting 2.  It can be seen from the table that the BAGofT with pre-selection still has better performance than the one without pre-selection.

\begin{table}[!ht]
\centering
\caption{Rejection rates at the significance level of $0.05$ from 100 replications with fixed coefficients. We assess \eqref{eq_hdcMTA}.  The available covariates for the BAGofT are $x_1,\dots,x_{500}$.}\label{tab_fixcoef}
\begin{tabular}{lllll}
    & $\beta_6$   values     & 0   & 1  & 2   \\\hline
Setting 1&Pre-selected & 0.04 & 0.61 & 1.00    \\
        &Not Pre-selected  & 0.06 & 0.08 & 0.19 \\
 &    $\beta_6$   values     & 0   & 0.5  & 1   \\\hline
Setting 2 &Pre-selected & 0.06 & 0.34 & 0.97 \\
&Not Pre-selected  & 0.08 & 0.08 & 0.25
\end{tabular}
\end{table}

\begin{table}[!ht]
\centering
\caption{Rejection rates  with randomly generated coefficients. Other settings are the same as Table~\ref{tab_fixcoef}.}
\begin{tabular}{lllll}\label{tab_rancoef}
    & $\sigma_6$   values     &  0 & 1 & 2  \\\hline
Setting 1&Pre-selected &   0.05	&0.30	& 0.56  \\
        &Not Pre-selected  & 0.07	 &0.09&	0.15\\
 &    $\sigma_6$   values     & 0   & 0.5  & 1   \\\hline
Setting 2 &Pre-selected & 0.01	&0.28	&0.51  \\
&Not Pre-selected  & 0.04 &	0.09 &	0.27
\end{tabular}
\end{table}

%%%%%%%%%%%%%%%%%%%%%%%%%%%%%%%%%%%%%%%%%%%
% Rcodes116_HD_HDSEL
%%%%%%%%%%%%%%%%%%%%%%%%%%%%%%%%%%%%%%%%%%%

\section{A comparison between the BAGofT and GRP test in testing high dimensional models}\label{supp_grp_comparison}

In this section, we consider assessing high dimensional parametric classification models.
The BAGofT is compared with the GRP test, which is the state-of-the-art to measure the GOF of high dimensional generalized linear models.

The simulation procedure is the same as Section~\ref{supp_preselection} with fixed coefficients only and the MTA is the lasso logistic regression fitted on the main effects of $x_1,\dots x_{500}$. 
The BAGofT applies variable pre-selection with size 5.   The results in Table~\ref{hdhmcomp} shows that the BAGofT outperforms the GRP test in both Setting 1 (missing an interaction term) and Setting 2 (missing a quadratic term). It seems that the GRP test may be too conservative in rejecting~$H_0$. 
\begin{table}[!ht]
\centering
\caption{Rejection rates of the BAGofT and GRP test for assessing the lasso logistic regression model fitted on  $x_1,\dots x_{500}$ at the significance level of $0.05$. }\label{hdhmcomp}
\begin{tabular}{lllll}
          & $\beta_6$   values     &  0 & 1 & 2  \\\hline
Setting 1&BAG & 0.05 & 0.52 & 1.00    \\
&GRP & 0.00    & 0.04 & 0.79 \\
          & $\beta_6$   values     &  0 & 0.5 & 1  \\\hline
Setting 2&BAG & 0.05 & 0.32 & 0.96 \\
&GRP & 0.00    & 0.00    & 0.61
\end{tabular}
\end{table}

\section{Assessing low dimensional classification learning procedures} 
\label{supp_lowdim}

%Rcodes147_procedures

In this Section, we focus on some low dimensional classification learning procedures and demonstrate the application of the BAGofT. The data are generated from the Bernoulli distribution with conditional probability 
\[P(y=1|x_1,x_2,x_3) = 1/(1 + \exp(  \sin(x_1) + 1.8  x_2  x_3 + x_4) ).\]
The covariates are independently generated, where 
 $x_1$ is from $\N(0,2.25)$,  $x_2$, $x_3$, and $x_4$ are from $\N(0,1)$.  The PTAs are  feed-forward neural network, Random Forest, and  logistic regression model.   For the logistic regression model, we consider the main effects of $x_1$-$x_4$ only. Therefore, it does not converge to the data generating model. 

\begin{figure}[!ht]
  \centering
  \includegraphics[width=1\textwidth]{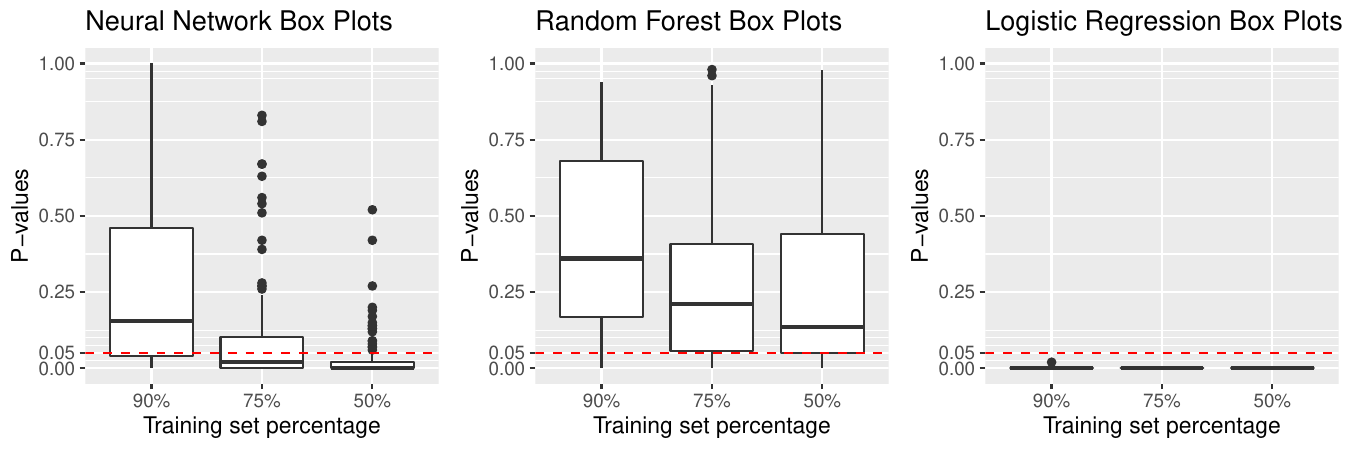}
  \caption{The BAGofT \textit{p}-value box plots for the neural network, Random Forest, and logistic regression at the significance level of 0.05. }\label{npComp}
\end{figure}

We first randomly generate a sample with size 500 and apply the BAGofT to  the PTA. The BAGofT takes 40 data splittings, and its adaptive partition is based on all available covariates $x_1$-$x_4$. The above process is performed with 100 replications and the \textit{p}-values are summarized in  Figure~\ref{npComp}.  The  neural network is fitted  by the package \textit{keras} \citep{keras} with two hidden layers that consists of 80 and 5 neurons, respectively.  The activation function is \textit{ReLu}~\citep{nair2010rectified}. The Random Forest is fitted by the package \textit{randomForest} \citep{randomForest}. We average over 500 trees, and each tree randomly takes 2 covariates.

It can be seen from  Figure~\ref{npComp} that the neural network is likely to be rejected except for the splitting ratio of $90\%$. Thus, it corresponds to \textbf{Pattern 3}. The majorities of Random Forest's \textit{p}-values are above 0.05. Therefore, it corresponds to \textbf{Pattern 1}. Apparently, the logistic regression model with main effects only fails to capture the nonlinearity from the data generating model (\textbf{Pattern 4}). % For the majority ($40\%$) of the times, the neural network belongs to \textbf{Pattern 3} and majority ($59\%$) of the times the Random Forest belongs to \textbf{Pattern 1}. %{\color{red}Our readers may question the nonconformity between the single realization results and the population-level results.   Since although we have a major Case from multiple replications, other cases also have a chance to occur in a single realization. We argue that this situation is similar to the confidence interval, where the interval may or may not cover the true parameter in a single realization.  The BAGofT helps measure the population level goodness-of-fit in the sense that the major case will dominate if we can repeat the experiment multiple times.}

 %Rcodes97_adaptive_K_std
  \section{Test statistic variance and number of splittings} \label{supp_splits}

\begin{figure}[!hb]
  \caption{The test statistic value versus the numbers of splittings in the simulations from Section~5.1 in the main text for Model \textit{A}. The test statistics are calculated by taking the mean of the values obtained from the multiple splittings. Each line stands for the results with different number of splittings from a  dataset generated from random coefficients. }\label{trace1}
  \centering
    \includegraphics[width=0.95\textwidth]{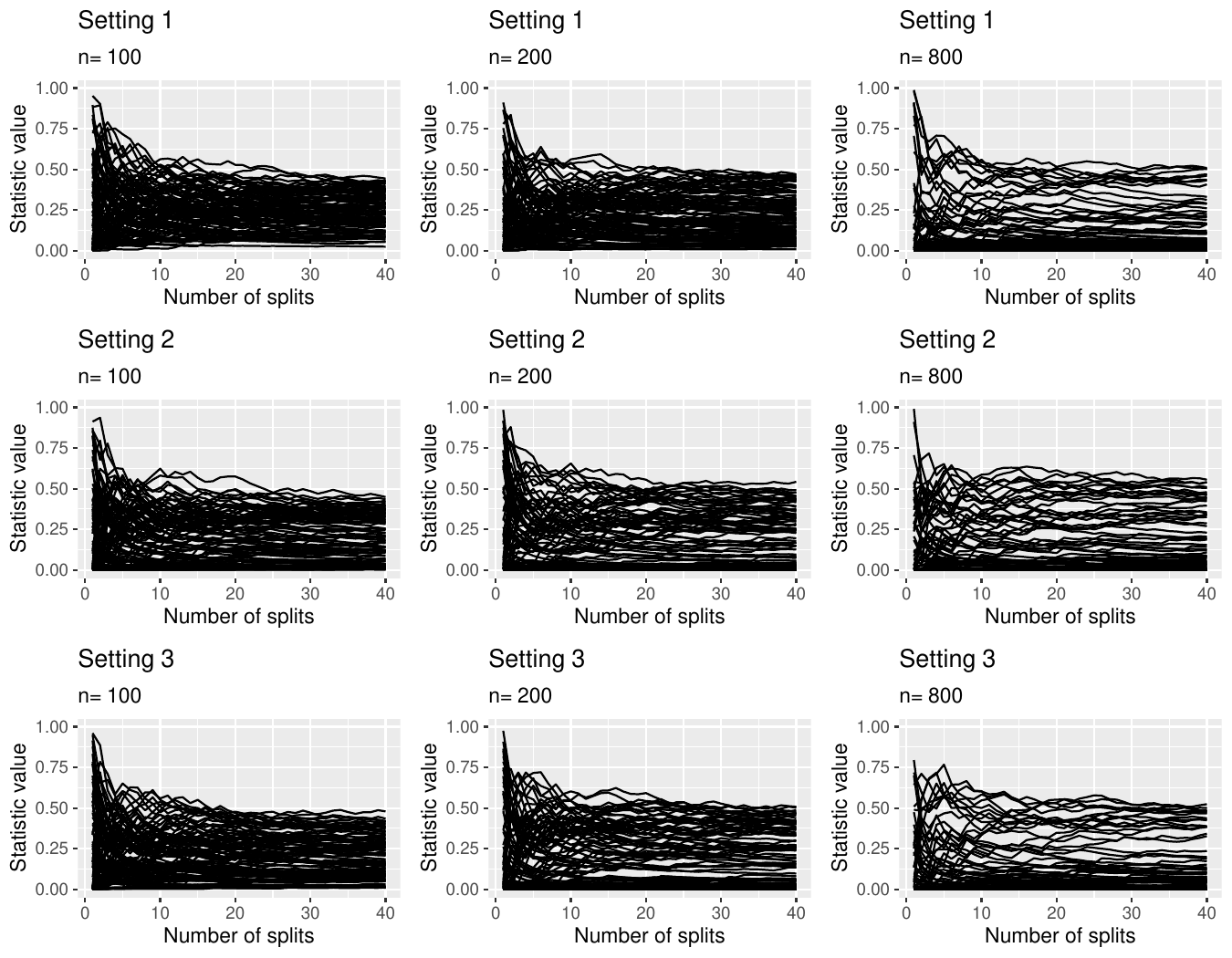}
\end{figure}

\begin{figure}[!ht]
  \caption{The test statistic value versus the  number of splittings in the simulations from  Section~5.1 in the main text for Model \textit{B}. Other settings are the same as Figure~\ref{trace1}.}\label{trace2}
  \centering
    \includegraphics[width=0.9\textwidth]{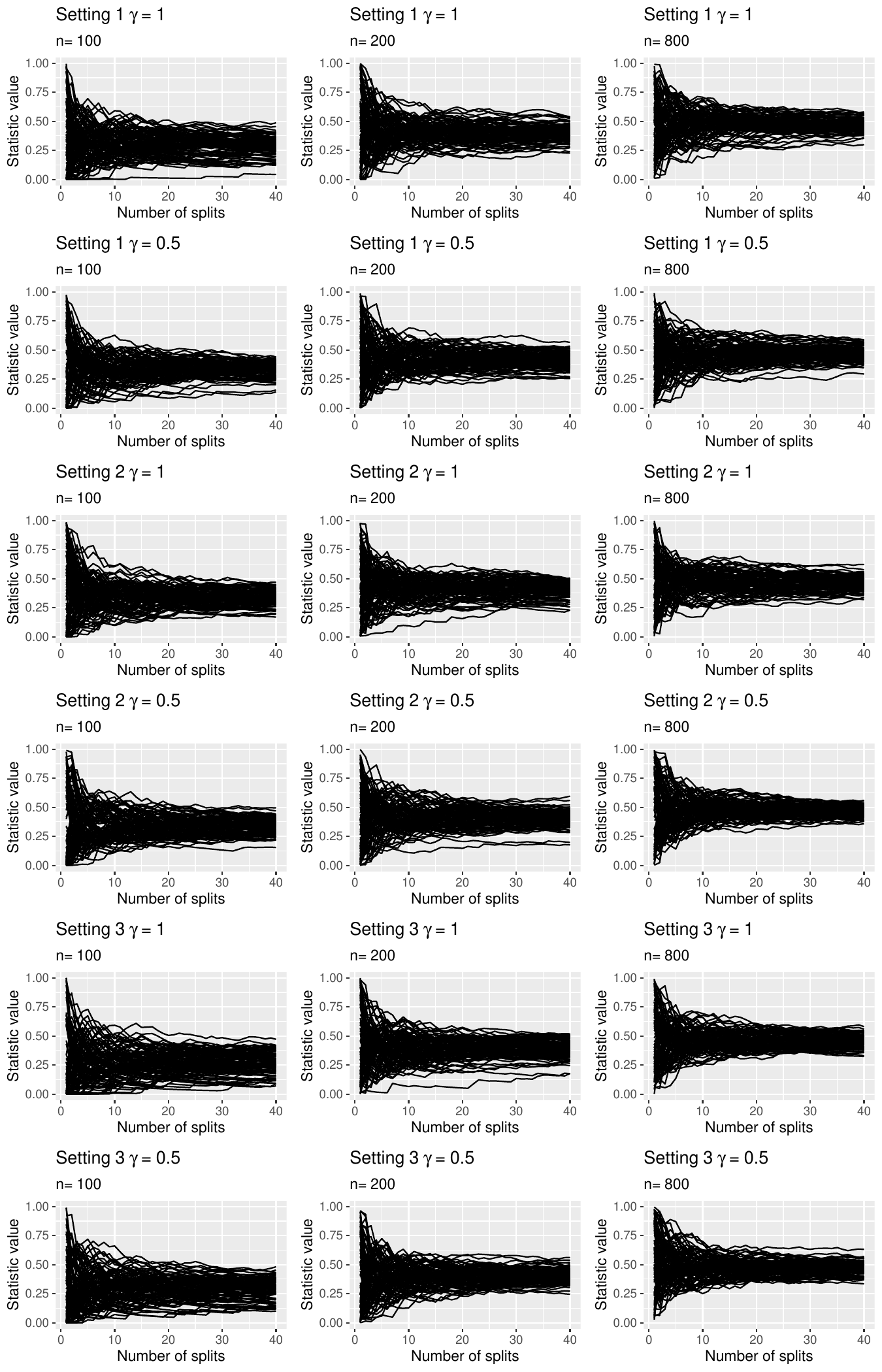}
\end{figure}

  To study the relationship between the test statistic variation and the number of splittings, we calculate the test statistic from the settings in Section~5.1 in the main text. The result from multiple splittings is combined by taking the sample mean. The results are shown in Figure~\ref{trace1} and Figure~\ref{trace2}. It can be seen that 10 to 20 splittings are sufficient to obtain a stable result for most cases.
  
\section{Variable importance of the covariates}
\label{supp_varimport}

\begin{figure}[!ht]
  \caption{Frequencies of the covariates with the largest (Random Forest) variable importance in Setting 1 (missing the main effect of $x_3$) from Section~5.1 in the main text  when the model is misspecified. }\label{vp1}
  \centering
    \includegraphics[width=0.8\textwidth]{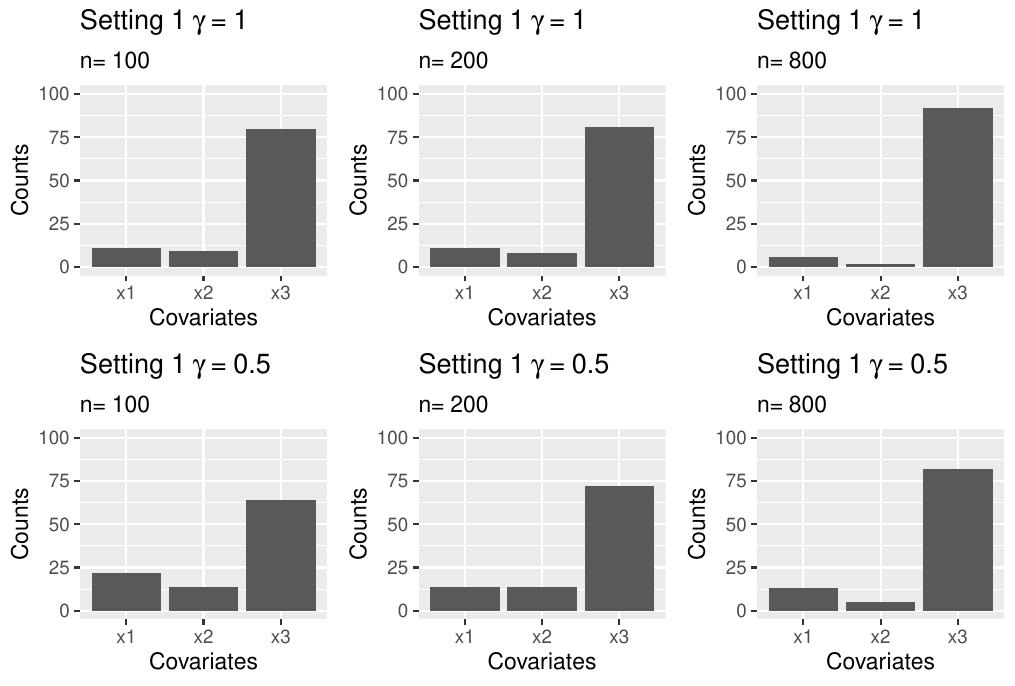}
\end{figure}

\begin{figure}[!ht]
  \caption{Frequencies of the covariates with the largest (Random Forest) variable importance in Setting 3 (missing the quadratic effect of $x_1$) from Section~5.1 in the main text  when the model is misspecified. }\label{vp2}
  \centering
    \includegraphics[width=0.8\textwidth]{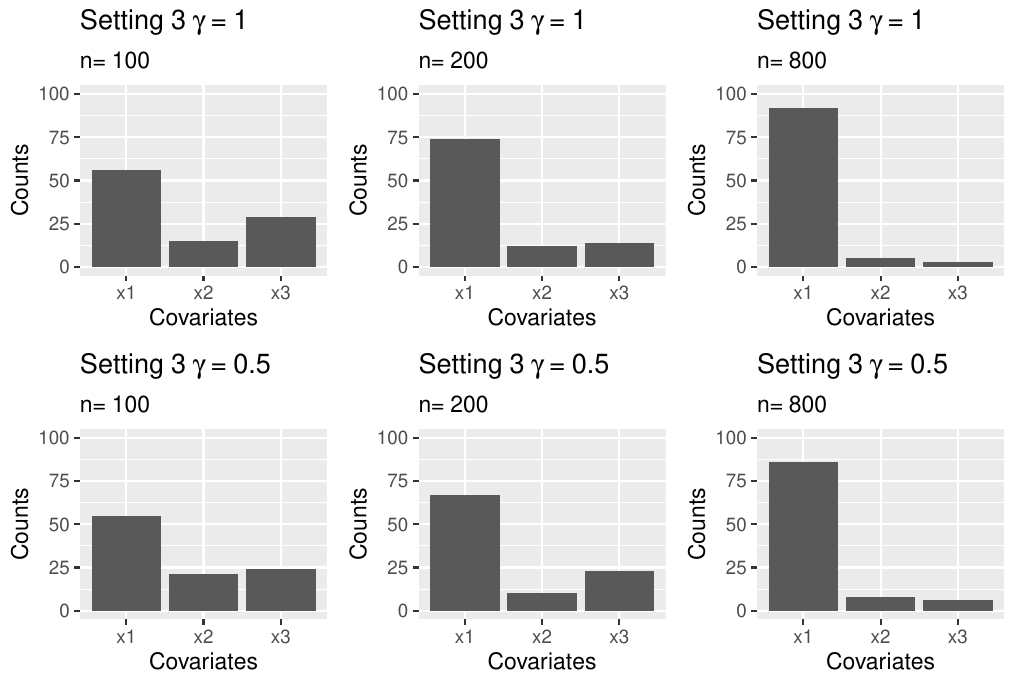}
\end{figure}

 We plot the frequencies of the covariates with the largest variable importance in Setting 1 and Setting 3 from Section~5.1 in the main text when the model is misspecified. The results in Figure~\ref{vp1} and Figure~\ref{vp2} show that the variable importance can be used to successfully identify the source of underfitting in majority of the times.

\section{PTA settings and  AUC results from COVID-19 CT scans data} \label{supp_covid}
The detailed settings of the PTAs in Section~6.3 in the main text are as follows.
The neural networks are fitted by the \textit{R} package \textit{keras} \citep{keras} with 1 hidden layer and the \textit{ReLu}~\citep{nair2010rectified} activation function. The XGBoost classifiers are fitted by the \textit{R} package \textit{xgboost} \citep{xgboost} with learning rate (eta) 0.04, maximum depth of a base learner (max\_depth) 7, subsample ratio of the training data when training each base learner (subsample) 0.6, subsample ratio of the variables when training each base learner (colsample) 0.1, and number of base learners (nrounds) 10 or 500. Due to the complexity of the neural networks and XGBoost with 500 based learners, their outputs are unstable. To improve the reproducibility of the results, for each training dataset, we independently fit those classifiers  with the same structure but with different random seeds 20 times, and output their averaged fitted probabilities.
For the PTAs in Section~6.3 from the main text, we also calculate the area under the receiver operating characteristic curve (AUC) in Table~\ref{tab_AUC}. The results here are consistent with those reported in the main paper.

\begin{table}[!ht]
\centering
\caption{
Prediction AUC from classification procedures fitted on the COVID-19 data \citep{he2020sample}. The notations are the same as Table~2 from the main text.}
\label{tab_AUC}
\begin{tabular}{rrrr}
 Splitting ratio& 90\% & 75\% & 50\% \\ 
  \hline
NNET-1 & 0.77 & 0.77 & 0.75 \\ 
NNET-7 & 0.79 & 0.78 & 0.77 \\ 
XG-10& 0.69 & 0.70 & 0.69\\
XG-500 & 0.79 & 0.78 & 0.76 
\end{tabular}
\end{table}

\end{appendices}

% \section*{Acknowledgement}
% {\color{blue} We greatly appreciate the insightful and constructive comments by the reviewers, AE and the Editor, which helped us improve the work substantially.}
\clearpage
\section*{Acknowledgement}
%\input{Acknowledgement}

% (For double-blind submission, hide this section. Otherwise, put this (including the arxiv version). )}

%The authors thank two anonymous reviewers and the Associate Editor for their constructive and very helpful comments.

This paper is based upon work supported by the Army Research Laboratory and the Army Research Office under grant number W911NF-20-1-0222, and the National Science Foundation under grant number ECCS-2038603. 

\bibliographystyle{asa}

\bibliography{bibliography}
\end{document}